\documentclass[smallextended,final]{svjour3}
\makeatletter
\usepackage[12pt]{extsizes}
\newcommand\subparagraph{%
	\@startsection{subparagraph}{5}
	{\parindent}
	{3.25ex \@plus 1ex \@minus .2ex}
	{-1em}
	{\normalfont\normalsize\bfseries}}
\makeatother
\let\subparagraph\relax
\usepackage{amsmath,amssymb,amsbsy,amsfonts,latexsym,amsopn,amstext,amsxtra,euscript,amscd,hyperref}
\usepackage{mathptmx}      % use Times fonts if available on your TeX system
\usepackage{graphicx}
\usepackage{algorithm}
\usepackage{algorithmic}
\usepackage[numbers]{natbib}
\usepackage{notoccite}
\usepackage{graphics,graphicx}
\usepackage{epsfig,epstopdf}
\usepackage{array,tikz}
\usepackage{caption}
\usepackage{slashbox}
\usepackage{wrapfig}
\usepackage{setspace}
\usepackage{titlesec}
\usepackage{tikz}
\usetikzlibrary{shapes.geometric, positioning}
\newcolumntype{C}{@{}c@{}}

\renewcommand{\arraystretch}{1.25}
\usepackage{float}
\usepackage[top=1.5cm, left=1.5cm,right=1.6cm,bottom=1.28cm]{geometry}
\newcounter{myeqno}
\usepackage{chngcntr}
\usepackage{tikz}
\usetikzlibrary{shapes.geometric,arrows}
\usetikzlibrary{shapes.geometric,arrows,fit,matrix,positioning,shapes.multipart}
\tikzstyle{startstop}=[rectangle, rounded corners, minimum width=3cm, minimum height=1cm, draw=black]
\tikzstyle{startstop1}=[rectangle, rounded corners, minimum width=8cm, minimum height=4cm, draw=black]
\tikzstyle{startstop2}=[square, rounded corners, minimum width=0.5cm, minimum height=1cm, draw=black]
\tikzstyle{startstop3}=[square, rounded corners, minimum width=5cm, minimum height=1cm, draw=black]
\tikzstyle{startstop4}=[square, rounded corners, minimum width=1cm, minimum height=5cm, draw=black]
\tikzstyle{round} = [ellipse, minimum width=3cm, minimum height=1cm, draw= black]
\tikzstyle{startstop2}=[rectangle, rounded corners, minimum width=2cm, minimum height=1cm, draw=black]
\tikzstyle{arrow} =[draw, -latex']
\renewcommand{\arraystretch}{1.5}
\usepackage{longtable,lscape}
\usepackage{rotating}
\usepackage{multirow,multicol}
\usepackage{enumerate}
\usepackage{subcaption}
\usepackage{makecell}

\usepackage{stfloats}

\usepackage{pdflscape}
\usepackage{xcoffins}
\NewCoffin\tablecoffin

\usepackage{color,soul}
\usepackage{xcolor}
\usepackage{colortbl}
\definecolor{mygrey}{RGB}{220,220,220}
\graphicspath{{images/}}
\usepackage{amsmath}
\usepackage{multirow}
\begin{document}
	\title{Advancements in Multimodal Differential Evolution: A Comprehensive Review and Future Perspectives}
    % \title{Multimodal Differential Evolution: A Detailed Review}
	\author{Dikshit Chauhan$^{1,2}$\and Shivani$^2$\and Donghwi Jung$^{3*}$ \and  Anupam Yadav$^{2*}$}
	\thanks{corresponding author}
	\institute{$^*$Corresponding author\\$^1$Department of Electrical and Computer Engineering, National University of Singapore, 119077, Singapore 
    \\$^2$Department of Mathematics and Computing, Dr. B.R. Ambedkar National Institute of Technology Jalandhar, Jalandhar-144008, Punjab, India.
    \\$^3$School of Civil, Environmental, and Architectural Engineering, Korea University, Seoul 02841, Republic of Korea\\
	\email{dikshitchauhan608@gmail.com (Dikshit Chauhan), sainishivani2310@gmail.com (Shivani), sunnyjung625@korea.ac.kr (Donghwi Jung), anupam@nitj.ac.in (Anupam Yadav)}}
	\maketitle
	\begin{abstract}
	% Multimodal optimization involves identifying multiple global and local optima of a function, allowing users to gain insights into different optimal solutions within the search space. Evolutionary Algorithms (EAs), owing to their population-based nature, are particularly effective in finding multiple solutions in a single simulation run, which gives them a clear advantage over classical optimization techniques that require multiple restarts without any guarantees of discovering diverse solutions. Among the various evolutionary techniques, Differential Evolution (DE) has emerged as one of the most powerful and versatile optimizers for continuous parameter spaces. DE has demonstrated considerable success in multimodal optimization by leveraging its population-based search capabilities to maintain and promote the formation of multiple stable subpopulations, each focusing on different optima. The evolution of DE, particularly in multimodal optimization, has progressed significantly over recent years, with numerous advancements in parameter adaptation, hybridization with other algorithms, and its application to a wide range of domains. So, we find that it is high time to provide a critical review of the latest literature published and also to point out some important future avenues of research. This paper is the first of its kind to present and aim to provide a comprehensive review of the latest developments in DE for multimodal optimization, including the adaptation of DE techniques for handling multiple optima and its diverse applications in real-world scenarios.

    Multi-modal optimization involves identifying multiple global and local optima of a function, offering valuable insights into diverse optimal solutions within the search space. Evolutionary algorithms (EAs) excel at finding multiple solutions in a single run, providing a distinct advantage over classical optimization techniques that often require multiple restarts without guarantee of obtaining diverse solutions. Among these EAs, differential evolution (DE) stands out as a powerful and versatile optimizer for continuous parameter spaces. DE has shown significant success in multi-modal optimization by utilizing its population-based search to promote the formation of multiple stable subpopulations, each targeting different optima. Recent advancements in DE for multi-modal optimization have focused on niching methods, parameter adaptation, hybridization with other algorithms including machine learning, and applications across various domains. Given these developments, it is an opportune moment to present a critical review of the latest literature and identify key future research directions. This paper offers a comprehensive overview of recent DE advancements in multimodal optimization, including methods for handling multiple optima, hybridization with EAs, and machine learning, and highlights a range of real-world applications. Additionally, the paper outlines a set of compelling open problems and future research issues from multiple perspectives.

	\end{abstract}
% \textbf{MM-MOOP: Multimodal Multi-objective Optimization Problem}\\
% \textbf{MMOP: Multimodal Optimization Problem}

\section{Introduction}

Optimization plays a pivotal role in solving complex challenges across diverse domains, including engineering, healthcare, and machine learning~\cite{bandaru2017data,del2019bio}. In many real-world scenarios, identifying not just one, but multiple global and local optima of an objective function is highly desirable. This is especially significant in cases where physical or cost constraints limit the feasibility of implementing a single best solution \cite{zhan2022survey}. By discovering multiple solutions, decision-makers gain the flexibility to seamlessly switch between alternatives, ensuring system performance remains robust while minimizing disruptions~\cite{das2010differential}. Additionally, identifying multiple solutions reveals hidden relationships within the problem landscape, enhancing our understanding of the system's functional behavior.

Multimodal optimization problems (MMOPs) arise in contexts where multiple optimal solutions exist \cite{nekouie2016new}. These are common in real-world applications, where solutions are represented as peaks in the objective landscape~\cite{li2013benchmark,ghorbanali2023comprehensive}. Unlike traditional optimization tasks that focus solely on finding a single global optimum, MMOPs require the discovery of both global and local optima through comprehensive search-space exploration~\cite{li2016seeking}. The importance of MMOPs spans various fields, such as industrial engineering, computer vision \cite{oskouie2014multimodal}, and bioinformatics~\cite{lu2022evolving}. For example, pedestrian detection~\cite{yang2021probabilistic}, electromagnetic machine design~\cite{yoo2015novel,yoo2017new}, multi-solution traveling salesman problems~\cite{huang2019niching}, and protein structure prediction~\cite{wong2010protein} exemplify MMOPs where the ability to identify multiple globally optimal solutions provides decision-makers with robust and diverse options.

Addressing MMOPs becomes particularly challenging when dealing with MMOPs, where evaluating objective functions involves significant computational cost. Such scenarios often require time-intensive simulations or costly physical experiments, as seen in warship decoy system design~\cite{ji2021dual,hong2017simulation}. In addition, certain multi-objective optimization problems (MOPs) exhibit multimodal characteristics, leading to what is termed multimodal multi-objective optimization problems (MM-MOOPs). These involve multiple Pareto-optimal sets (PSs) corresponding to the same points on the Pareto front (PF) \cite{liu2018multimodal}. Examples include the multiobjective knapsack problem \cite{jaszkiewicz2002performance}, architectural layout design~\cite{rahbar2022architectural}, and multiobjective scheduling~\cite{han2017evolutionary}. Unlike conventional MOPs, MM-MOOPs require discovering multiple PSs to offer robust, diverse solutions tailored to varying needs.

% A key challenge in solving MMOPs lies in the efficient identification and retention of multiple global optima in a single algorithmic run. These problems are often characterized by rugged, multimodal landscapes, with global optima scattered across different regions of the search space~\cite{wang2014mommop}. Evolutionary algorithms (EAs), such as genetic algorithms (GA) \cite{akopov2019parallel}, particle swarm optimization (PSO) \cite{houssein2021major,zhang2020modified}, and DE \cite{price2006differential,das2016recent}, are widely used for solving complex optimization problems due to their self-learning and self-adaptive characteristics~\cite{zhang2011evolutionary}. They explore and exploit the search space by maintaining a population of candidate solutions, making them suitable for problems that traditional mathematical methods struggle to solve~\cite{talbi2009metaheuristics}. However, conventional EAs are primarily designed to locate a single global optimum, limiting their direct applicability to MMOPs~\cite{qu2012differential}.

Addressing MMOPs effectively requires identifying and maintaining multiple global optima within a single execution of an algorithm. These problems exhibit complex, multimodal landscapes where optimal solutions are distributed across different regions of the search space~\cite{wang2014mommop}. Evolutionary algorithms (EAs), including genetic algorithms (GA) \cite{akopov2019parallel}, particle swarm optimization (PSO) \cite{houssein2021major,zhang2020modified}, and differential evolution (DE) \cite{price2006differential,das2016recent}, are widely employed for tackling complex optimization tasks due to their adaptability and learning capabilities~\cite{zhang2011evolutionary}. By maintaining a diverse population of candidate solutions, these algorithms effectively explore and exploit the search space, making them advantageous over traditional mathematical methods, which often struggle with such problems~\cite{talbi2009metaheuristics}. Nevertheless, standard EAs are inherently designed to converge toward a single global optimum, which limits their effectiveness in handling MMOPs~\cite{qu2012differential}.

To address this limitation, niching techniques have been integrated into EAs, enabling the population to be divided into subpopulations (or niches) that independently evolve toward different optima~\cite{li2016seeking,li2009niching}. Inspired by ecological niches, this approach encourages diversity, allowing EAs to search multiple regions of the solution space simultaneously~\cite{yang2016multimodal,wu2016differential}. Traditional niching methods, such as crowding~\cite{de1975analysis} and fitness sharing~\cite{holland1992adaptation}, rely on user-defined parameters, which may hinder their performance~\cite{stoean2010multimodal}. To overcome this, parameter-independent niching techniques~\cite{qu2012differential,li2009niching,gao2013cluster} have been developed, focusing on both exploration and exploitation to enhance performance on MMOPs. While niching-based EAs are effective in identifying multiple regions of interest, they often struggle to maintain and recover optima with high accuracy, especially in rugged search spaces~\cite{lacroix2016region}. 

Hybrid EAs, also called memetic algorithms, have been proposed to improve exploitation capabilities by integrating local search techniques, such as Gaussian-based refinements~\cite{yang2016multimodal}. These hybrid approaches refine solutions near optima while maintaining global diversity, making them more effective for complex MMOPs~\cite{huang2019niching,ren2013scatter}. Among EAs, DE has gained significant attention for its ability to maintain diversity and its simplicity of implementation~\cite{das2016recent,pant2020differential}. Researchers have further enhanced DE for MMOPs by incorporating strategies such as multimodal mutation~\cite{biswas2014improved,biswas2014inducing,zhoa2020local}, multi-objective methods~\cite{bandaru2013parameterless,basak2012multimodal}, and archive-based techniques~\cite{zhoa2020local,lin2019differential,wang2022adaptive,wang2022multimodal}. 

Multimodal mutation strategies in DE enhance exploration by considering both the fitness and spatial distance between individuals when selecting parents, ensuring offspring are distributed across diverse regions of the solution space~\cite{biswas2014improved,biswas2014inducing,liang2014differential}. Archive-based techniques preserve population diversity by storing potential solutions and mitigating premature convergence~\cite{zhoa2020local,wang2022adaptive}, though they often involve complex rules and operate primarily at the population level, limiting exploitation capabilities~\cite{wang2022multimodal}. This approach helps maintain high search diversity, enabling DE to effectively locate scattered optima across different regions. Multi-objective approaches to MMOPs draw on the similarity between MMOPs and MM-MOOPs, where both aim to find multiple optimal solutions. This involves transforming MMOPs into bi-objective problems~\cite{bandaru2013parameterless,deb2012multimodal}. Typically, one objective reflects the original MMOP, while a second, complementary objective is defined to encourage diversity and ensure comprehensive coverage of the PF~\cite{cheng2017evolutionary}. The main challenge lies in designing this secondary objective to maintain conflict with the primary objective, ensuring the resulting PF encompasses all global optima~\cite{bandaru2013parameterless}.
\begin{figure}
    \centering
    \includegraphics[width=0.8\linewidth]{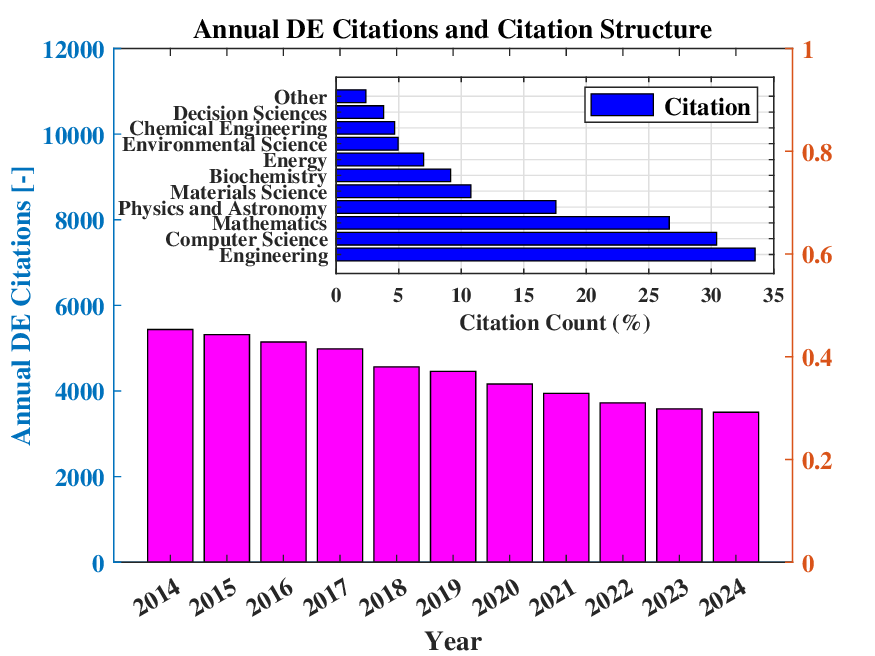}
    \caption{DE annual citations.}\label{fig: DE citations}
\end{figure}
 As discussed above, many DE-based algorithms have been developed specifically to handle the unique challenges posed by MMOPs in recent years. These approaches incorporate various strategies such as niching, clustering, and dynamic population adaptations to enhance DE's performance in multimodal multi-objective settings. Despite the success of these methods, there remain several open questions regarding the most effective ways to balance exploration and exploitation in high-dimensional search spaces and to improve the scalability and robustness of DE algorithms. This survey paper provides a comprehensive review of DE algorithms developed for MMOPs from 2017 to 2024, categorizing them based on learning mechanisms and applications. It highlights approaches that address critical challenges, including diversity maintenance, convergence, and scalability. Additionally, it explores hybrid methods combining DE with other evolutionary techniques and machine learning models, offering insights into future research directions for adaptive, scalable, and robust DE algorithms tailored to real-world applications. 

\begin{figure}
\begin{center}
\resizebox{1\linewidth}{!}{\begin{tikzpicture}[scale=0.7]

    % Define hexagon style
    \tikzstyle{hexagon} = [draw, shape=regular polygon, regular polygon sides=6, 
    minimum size=2cm, align=center, inner sep=2pt, text width=1.7cm, font=\footnotesize]

    % Colors for alternation
    \def\colorA{red!40}
    \def\colorB{pink!60}

    % Center Hexagon (First Layer)
    \node[hexagon, fill=red!40] (A) at (0,0) {\textbf{\Large{DE}}};

    % First Layer (Around Center)
    \foreach \i/\text in {
        0/{Basic DE}, 
        60/{SHADE}, 
        120/{L-SHADE}, 
        180/{Niching-based DE}, 
        240/{Fitness Sharing}, 
        300/{Crowding}
    } {
        \node[hexagon, fill=\colorB] (B\i) at ([shift={(\i:3.2)}]A) {{\text}};
    }
    
    % Second Layer
    \foreach \i/\text in {
       30/{Hill-Valley Method}, 
        90/{Clustering-based DE}, 
       150/{Cluster-based Partitioning}, 
        210/{Minimum Spanning Tree}, 
        270/{Mutation and Parameter Adaptation}, 
        330/{Adaptive Mutation Strategies}
    } {
        \node[hexagon, fill=\colorA] (C\i) at ([shift={(\i:6.5)}]A) {{\text}};
    }

    % Third Layer
    \foreach \i/\text in {
        60/{Hybrid Approaches}, 
        120/{Hybrid with Memetic Algorithms}, 
        180/{Hybrid with PSO}, 
        240/{Machine Learning-based DE}, 
        300/{Reinforcement Learning}, 
        0/{Surrogate Models}
    } {
        \node[hexagon, fill=\colorB] (D\i) at ([shift={(\i:7.5)}]A) {{\text}};
    }

    % Fourth Layer
    \foreach \i/\text in {
        30/{Dynamic \& Noisy Environments}, 
        90/{Multiple Strategies-based DE}, 
        150/{Theoretical Analysis}, 
        210/{Healthcare \& Biomedical}, 
        270/{Feature Selection Tasks}, 
        330/{Machine Learning Applications}
    } {
        \node[hexagon, fill=\colorA] (E\i) at ([shift={(\i:10)}]A) {{\text}};
    }

    % Fifth Layer
    \foreach \i/\text in {
        60/{Experimental Analysis}, 
        120/{Open Issues}, 
        180/{DE Perspective}, 
        240/{Application Perspective}, 
        300/{Explainable DE}, 
        0/{DE\\\& \\ LLM}
    } {
        \node[hexagon, fill=\colorB] (F\i) at ([shift={(\i:11.5)}]A) {{\text}};
    }

    % Sixth Layer
    \foreach \i/\text in {
       % 30/{Dynamic & Noisy Environments}, 
        90/{Better Evaluation Metrics}, 
       % 150/{Theoretical Analysis}, 
      %  210/{Surrogate-Assisted DE}, 
        270/{Quantum-Inspired DE} 
     %    0/{DE and LLM}
     %   330/{Explainable DE}
    } {
        \node[hexagon, fill=\colorA] (G\i) at ([shift={(\i:13)}]A) {{\text}};
    }

    % % Seventh Layer
    %  \node[hexagon, fill=\colorB] (H) at ([shift={(0:12.5)}]A) {\textbf{DE and LLM}};

\end{tikzpicture}} \caption{Thematic overview of DE .}\label{fig: enter-overview}
\end{center}
\end{figure}
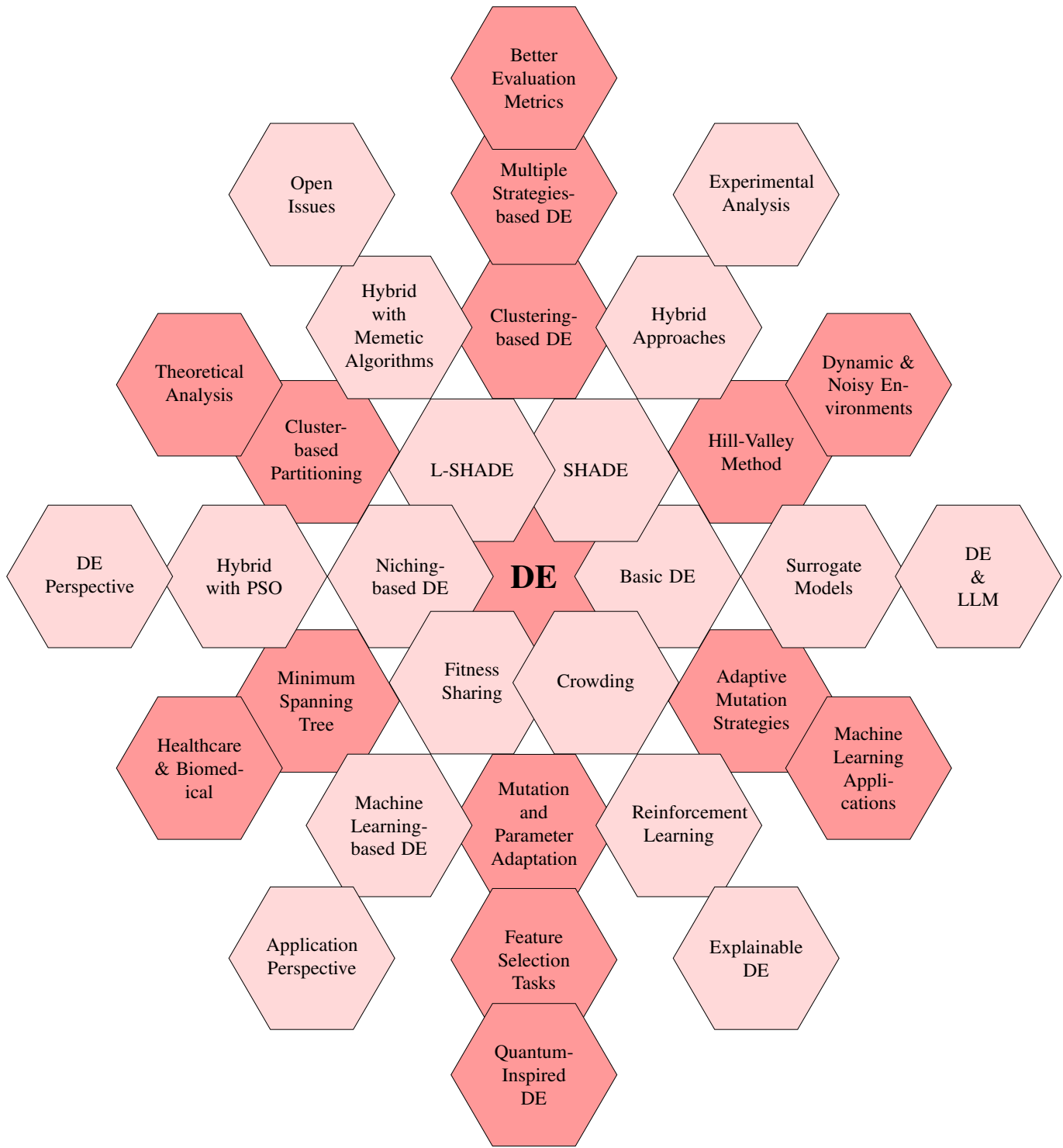

{The literature on MMOPs includes only a few review papers. Das et al. \cite{das2011real} published a comprehensive review in 2011, focusing on niching methods integrated into EAs for solving MMOPs. Barrera and Coello \cite{barrera2009review} reviewed hybrid approaches to particle swarm optimization (PSO) for MMOPs in 2010. More recently, Tanabe and Ishibuchi~\cite{tanabe2019review} examined advancements in multi-objective MMOPs, highlighting significant progress in this domain. In 2017, Li et al.~\cite{li2016seeking} provided an updated review of niching methods using EAs, summarizing developments since earlier works.

While several reviews on DE exist in the literature, these primarily address its general advancements and applications across various optimization scenarios. However, no dedicated review focuses on the progress and innovations in DE specifically for MMOPs. Recognizing this gap, this paper aims to bridge it by highlighting advancements and contributions of DE in tackling MMOPs. The growing relevance of DE in this domain underscores the need for such a focused review.

The citation structure of DE research is depicted in Fig.~\ref{fig: DE citations}, which provides an overview of the increasing body of work. The figure presents annual citation trends and distribution of DE across different disciplines. The magenta bar chart represents the annual citation counts from 2014 to 2024, showing consistent citation levels of approximately 4,000$-$6,000 per year, indicating sustained research interest. The horizontal blue bars illustrate the percentage distribution of DE citations across various fields. Engineering and Computer Science dominate, contributing over 30$\%$ and 25$\%$ of citations, respectively, followed by Mathematics, Physics, and Astronomy. Other areas, such as Biochemistry, Environmental Science, and Decision Sciences, contribute moderately, reflecting DE's interdisciplinary applications while emphasizing its prominence in technical and computational domains.} Fig.~\ref{fig: enter-overview} presents a thematic illustration of this work.

The structure of the paper is as follows: Section~\ref{sec: DE} explains the major and popular advancements in the original DE algorithm. Section~\ref{sec: niching DE} covers DE algorithms employing niching methods. Section~\ref{sec: clustering DE} focuses on clustering-based DE approaches. Mutation and parameter adaptation-based DE algorithms are presented in Section~\ref{sec: mutation DE}. Section~\ref{sec: hybrid DE} explores the hybridization of DE with other algorithms. Machine learning-integrated and multi-level DE approaches are discussed in Sections~\ref{sec: machine DE} and~\ref{sec: multi-level DE}, while Section~\ref{sec: multi objective DE} explains multi-objective DE algorithms, and the wide range of DE applications are discussed in Section~\ref{sec: applications DE}. The experimental results and analysis of various algorithms are discussed in Section~\ref{Sec: experimental}. Finally, future research directions are outlined in Section~\ref{sec: open questions}, and the paper concludes in Section~\ref{sec: conclusion}.

\section{Major Advancements in DE}\label{sec: DE}
\subsection{Basic DE}\label{subsec: de}

This section presents a concise overview of Differential Evolution (DE)~\cite{storn1997differential}, highlighting its core components. The standard DE variant, commonly denoted as DE/rand/1/bin, maintains a population \(\mathbf{P}\) consisting of \(N_p\) individuals. Each individual, also known as a target or parent vector, is represented as a solution vector \(\mathbf{x}_i = \{x_{i,1}, x_{i,2}, \ldots, x_{i, D}\}\), where \(D\) indicates the dimensionality of the search space. The evolutionary process progresses through multiple generations \(\{G = 1, 2, \ldots, G_{\text{max}}\}\), where \(G_{\text{max}}\) defines the maximum number of iterations. Each individual \(\mathbf{x}_i\) is randomly initialized within predefined bounds, given by \(\mathbf{x}_{\text{lb}} = \{x_{1,\text{lb}}, x_{2,\text{lb}}, \ldots, x_{N_p,\text{lb}}\}\) and \(\mathbf{x}_{\text{ub}} = \{x_{1,\text{ub}}, x_{2,\text{ub}}, \ldots, x_{N_p,\text{ub}}\}\), leading to an initial population \(\mathbf{x}_i^G = \{x_{i,1}^G, x_{i,2}^G, \ldots, x_{i,D}^G\}\) at generation \(G\).

Once initialized, DE refines the population through three primary operators: \textbf{mutation}, \textbf{crossover}, and \textbf{selection}. Mutation and crossover generate trial vectors, while selection determines whether the trial vector replaces the target vector in the subsequent generation. The detailed steps of DE are as follows:

\begin{enumerate}[(i)]
    \item \textbf{Mutation:} For each target vector \(\mathbf{x}_i\), a mutant vector \(\mathbf{v}_i = \{v_{i,1}, v_{i,2}, \dots, v_{i,D}\}\) is produced using the following rule:
    \begin{equation}
        DE/rand/1: \quad \mathbf{v}_{i} = \mathbf{x}_{r1} + F \cdot (\mathbf{x}_{r2} - \mathbf{x}_{r3}),
    \end{equation}
    where \(r_1, r_2, r_3\) are distinct, randomly chosen indices from the range \([1, N_p]\) and different from \(i\). The scaling factor \(F\) controls the magnitude of perturbation and typically falls within the range \([0, 2]\). A higher \(F\) encourages exploration, while a lower \(F\) promotes exploitation. The DE mutation strategy follows the convention \texttt{DE/x/y/z}, where \texttt{x} indicates the perturbed vector, \texttt{y} represents the number of difference vectors, and \texttt{z} specifies the crossover type (\texttt{exp} for exponential, \texttt{bin} for binomial).

    \item \textbf{Crossover:} The crossover step blends the mutant vector \(\mathbf{v}_i\) with the target vector \(\mathbf{x}_i\) to create a trial vector \(\mathbf{u}_i = \{u_{i,1}, u_{i,2}, \dots, u_{i,D}\}\). The binomial crossover, commonly used in DE, operates as follows:
    \begin{equation}\label{eq: binomial crossover}
        \mathbf{u}_{i,j} = 
        \begin{cases} 
            \mathbf{v}_{i,j}, & \text{if } \text{rand}_{i,j}(0,1) \leq CR \text{ or } j = j_{\text{rand}}, \\
            \mathbf{x}_{i,j}, & \text{otherwise}.
        \end{cases}
    \end{equation}
    Here, \(\text{rand}_{i,j}(0,1)\) is a uniformly distributed random value, and \(j_{\text{rand}}\) is a randomly chosen index ensuring at least one component of \(\mathbf{v}_i\) is inherited. The crossover rate \(CR \in [0, 1]\) determines the proportion of genes exchanged with the mutant vector.

    \item \textbf{Selection:} The selection phase applies a greedy criterion to decide whether the trial vector \(\mathbf{u}_i\) or the target vector \(\mathbf{x}_i\) proceeds to the next generation. For minimization problems, this decision follows:
    \begin{equation}\label{eq: de selection}
        \mathbf{x}_i = 
        \begin{cases} 
            \mathbf{u}_i, & \text{if } f(\mathbf{u}_i) \leq f(\mathbf{x}_i), \\
            \mathbf{x}_i, & \text{otherwise}.
        \end{cases}
    \end{equation}
    where \(f(\cdot)\) denotes the objective function. If the trial vector yields a superior or equal function value compared to the target vector, it replaces the target in the population. This mechanism enables DE to traverse flat fitness landscapes, reducing premature convergence~\cite{awad2018improved}.
\end{enumerate}

Since crossover exchanges information between mutant and target vectors, some decision variables may retain identical values across individuals. If multiple population members share the same values, the difference vector for those variables becomes zero. However, DE’s parent-offspring competition significantly reduces the likelihood of identical values persisting, particularly in early evolutionary stages.

	\subsection{SHADE}

The success history-based adaptive differential evolution (SHADE) \cite{tanabe2013success} builds upon JADE \cite{zhang2009jade}, refining its adaptive parameter control mechanism. In JADE, individuals evolve using the \textit{DE/current-to-pbest/1} mutation strategy:

\begin{equation}
    \mathbf{v}_i=\mathbf{x}_i+F_i\cdot (\mathbf{x}_{pbest}-\mathbf{x}_i)+F_i\cdot (\mathbf{x}_{r1}-\mathbf{x}_{r2}),
\end{equation}

where \( pbest \) is randomly chosen from the top \( 100p\% \) of the current population. Each solution vector \( \mathbf{x}_i \) is assigned adaptive control parameters \( F_i \) and \( CR_i \), which are sampled probabilistically from \( \mu_F \) and \( \mu_{CR} \) via Cauchy and normal distributions. The values of \( CR_i \) and \( F_i \) that lead to successful trial vectors are stored in \( S_{CR} \) and \( S_F \). These stored values are utilized to update \( \mu_{CR} \) and \( \mu_F \) at the end of each generation, using the arithmetic mean of \( S_{CR} \) and the Lehmer mean of \( S_F \).

In the \textit{DE/current-to-best/1} mutation strategy, the base vector is formed as a weighted combination of the target vector and the best individual in the population. The new donor vector is derived by modifying the position of \( \mathbf{x}_i \) along the direction of \( \mathbf{x}_\text{best} \), effectively guiding the search towards promising regions.

SHADE further improves upon JADE by incorporating a historical memory mechanism that records past successful parameter values. It maintains parameter archives \( M_F \) and \( M_{CR} \), each of size \( H \). In every generation, the control parameters \( CR_i \) and \( F_i \) for each individual \( \mathbf{x}_i \) are sampled from stored values \( M_{CR,r_i} \) and \( M_{F,r_i} \), where \( r_i \) is randomly chosen from the range \( [1, H] \). The parameters that contribute to improved offspring are recorded in \( S_{CR} \) and \( S_F \). At the end of the iteration, \( M_{CR,j} \) and \( M_{F,j} \) are updated using a weighted arithmetic mean and a weighted Lehmer mean, respectively, ensuring an adaptive and self-tuning mechanism for better convergence.

	\subsection{L-SHADE}\label{subsec: lshade}
	% L-SHADE~\cite{tanabe2014improving}, a winner of CEC 2014 single-objective optimization problems, extends SHADE with linear population size reduction (LPSR), continuously decreasing the population size according to a linear function. Initially, the population size is $N_p^{init}$, and it reduces to $N_p^{min}$ by the end of the run. After each generation $G$, the next generation's population size $N_p$ is calculated as:\begin{equation}\label{eq: linear reduction pop}
	% newN_p=round\left[N_p^{init}+\left(\frac{N_p^{min}-N_p^{init}}{MaxFE}\right)\cdot FE\right],
	% \end{equation} where $FE$ and $MaxFE$ are the current and maximum number of evaluations. If $newN_p$ is smaller than $N_p$, the $(N_p - newN_p)$ individuals in the population are removed from the bottom.
L-SHADE~\cite{tanabe2014improving}, the winner algorithm in the CEC 2014 competition for single-objective optimization, enhances SHADE by incorporating a linear population size reduction (LPSR) mechanism. This approach progressively decreases the population size following a linear function throughout the evolutionary process. Initially, the population consists of $N_p^{init}$ individuals, which gradually reduces to $N_p^{min}$ by the final stage of the optimization run. The population size at generation $G+1$ is determined using the equation:

\begin{equation}\label{eq: linear reduction pop}
	newN_p=round\left[N_p^{init}+\left(\frac{N_p^{min}-N_p^{init}}{MaxFE}\right)\cdot FE\right],
\end{equation}

where $FE$ represents the current function evaluations, and $MaxFE$ denotes the predefined maximum number of function evaluations allowed. If the computed $newN_p$ is smaller than the current population size $N_p$, then the algorithm eliminates $(N_p - newN_p)$ individuals, typically removing the least competitive solutions from the bottom of the ranking.

	\subsection{LSHADE-EpSin}\label{subsec: lshade-epsin}
To further improve the performance of L-SHADE, LSHADE-EpSin introduces an ensemble-based approach for adapting the scaling factor, leveraging an efficient sinusoidal adjustment mechanism~\cite{awad2016ensemble}. This strategy integrates two sinusoidal components: a non-adaptive decreasing function and an adaptive, history-informed increasing function. Additionally, a Gaussian Walk-based local search is incorporated in later generations to enhance the algorithm’s exploitation capabilities.

In LSHADE-EpSin, the mutation strategy follows the \textit{DE/current-to-pbest/1} scheme, augmented with an external archive. This mutation is defined as:

\begin{equation}
	\mathbf{v}_i=\mathbf{x}_i+F_i\cdot (\mathbf{x}_{pbest}-\mathbf{x}_i)+F_i\cdot (\mathbf{x}_{r1}-\mathbf{x}_{r2}),
\end{equation}

where $r_1 \neq r_2 \neq i$, with $\mathbf{x}_{r1}$ being randomly selected from the population. Meanwhile, $\mathbf{x}_{r2}$ is sampled from the combined set of the population $\mathbf{P}$ and an external archive $\mathbf{A}$, which retains previously replaced inferior solutions, ensuring additional diversity in search dynamics (i.e., $\mathbf{x}_{r2} \in \mathbf{P} \cup \mathbf{A}$).

\textbf{Parameter Adaptation:} LSHADE-EpSin employs an ensemble of adaptive mechanisms to regulate the scaling factor $F$. Two distinct sinusoidal-based strategies govern this adaptation:

\begin{equation}\label{eq: ensemble lshade-epsin}
	\begin{cases}
		F_i=0.5\cdot\left(\sin(\pi(2\cdot f_q\cdot G+1))\cdot \frac{G_{max}-G}{G_{max}}+1\right),\\
		F_i=0.5\cdot\left(\sin(\pi(2\cdot f_{q_1}\cdot G+1))\cdot \frac{G}{G_{max}}+1\right),
	\end{cases}
\end{equation}

where $f_q$ and $f_{q_1}$ represent sinusoidal function frequencies. The frequency $f_q$ remains constant, while $f_{q_1}$ adapts dynamically each generation following a Cauchy distribution:

\begin{equation}
	f_{q_1} = randc(\mu f_{r_i},0.1),
\end{equation}

where $\mu f_{r_i}$ is the Lehmer mean, randomly drawn from an external memory $M_{f_q}$ that maintains a record of successful mean frequencies from prior generations, stored in $S_{f_q}$. The index $r_i$ is selected randomly from $[1,H]$ at the end of each generation.

Both sinusoidal strategies contribute to $F_i$ adaptation during the initial half of the evolutionary process. In the latter half, $F_i$ is updated using a Cauchy distribution:

\begin{equation}
	F_i = randc(\mu F_{r_i},0.1).
\end{equation}

Moreover, the crossover rate $CR_i$ undergoes continuous adaptation throughout evolution, following a normal distribution:

\begin{equation}
	CR_i = randn(\mu CR_{r_i},0.1).
\end{equation}

	\subsection{LSHADE-cnEpSin}\label{subsec: lshade_cnepsin}

LSHADE-EpSin was developed as an enhancement of L-SHADE, incorporating an adaptive ensemble of sinusoidal functions for dynamically adjusting the scaling factor $F$. The algorithm randomly selects one of two sinusoidal strategies during the initial half of the generations: a non-adaptive sinusoidal decreasing function or an adaptive sinusoidal increasing function. A performance-based adaptation mechanism is further employed to dynamically determine the selection between these two strategies. 

Building on this framework, LSHADE-cnEpSin~\cite{awad2017ensemble} introduces an improved mechanism for selecting the scaling factor $F$, integrating covariance matrix learning with Euclidean neighborhoods to optimize the crossover process.

\textbf{Effective Selection Mechanism:}  
The selection between the two sinusoidal strategies is guided by their historical performance. During a designated learning period $L_p$, the algorithm tracks the number of successful and unsuccessful trial vectors produced by each sinusoidal configuration, denoted as $nS_j$ and $nF_j$, respectively. Initially, both strategies are assigned equal probability $p_j$ and are chosen randomly over the first $L_p$ generations. Afterward, their probabilities are adjusted using the following formulation:

\begin{align}\label{eq: effective selection}
	\begin{cases}
	S_j=\frac{\sum_{i=G-L_p}^{G-1}nS_{i,j}}{\sum_{i=G-L_p}^{G-1}nS_{i,j}+\sum_{i=G-L_p}^{G-1}nF_{i,j}}+\epsilon,\\
	p_j=\frac{S_j}{\sum_{j=1}^J S_j},
	\end{cases}
\end{align}

where $S_j$ represents the success rate of trial vectors produced by each sinusoidal strategy.

\textbf{Covariance Matrix Learning-Based Crossover:}  
LSHADE-cnEpSin employs a novel crossover operator that utilizes covariance matrix learning with Euclidean neighborhood selection (CMLwithEN). With probability $p_c$, this operator refines recombination by leveraging a neighborhood formed around the best individual $\mathbf{x}_{best}$. Individuals are ranked based on fitness, and the Euclidean distance to $\mathbf{x}_{best}$ is computed. A subset of the top-performing individuals, constituting $N_p\times p_s$ (where $p_s = 0.5$), defines the neighborhood. As the population size shrinks, the neighborhood dynamically scales down accordingly. From this selected neighborhood, the covariance matrix $C$ is estimated as:

\begin{equation}
	C = O_b D_g O_b^T,
\end{equation}

where $O_b$ and $O_b^T$ are orthogonal matrices, and $D_g$ is a diagonal matrix containing the eigenvalues. The target and trial vectors undergo a transformation using the orthogonal matrix $O_b^T$:

\begin{equation}
	\mathbf{x}_i^{'}=O_b^T\mathbf{x}_i,~~\mathbf{v}_i^{'}=O_b^T\mathbf{v}_i.
\end{equation}

The binomial crossover (Eq.~\eqref{eq: binomial crossover}) is then applied to the transformed vectors to generate the trial vector $\mathbf{u}_i^{'}$. Finally, the trial vector is projected back into the original coordinate system:

\begin{equation}
	\mathbf{u}_i=O_b\mathbf{u}_i^{'}.
\end{equation}

A key comparison among DE variants is presented in Table \ref{tab: de_variants_comparison} in terms of enhancements.

\begin{table}[h]
    \centering
    \caption{Comparison of DE, SHADE, L-SHADE, LSHADE-EpSin, and LSHADE-cnEpSin in terms of key enhancements.}\label{tab: de_variants_comparison}
  \resizebox{1\linewidth}{!}{  \begin{tabular}{|c|c|c|c|c|c|}
        \hline
        \textbf{Algorithm} & \textbf{Parameter adaptation} & \textbf{Population size reduction} & \textbf{External archive} & \textbf{Sinusoidal adaptation} & \textbf{Covariance learning} \\ 
        \hline
        DE & No & No & No & No & No \\ 
        \hline
        SHADE & Yes (Success-based) & No & No & No & No \\ 
        \hline
        L-SHADE & Yes (Success-based) & Yes (Linear Reduction) & No & No & No \\ 
        \hline
        LSHADE-EpSin & Yes (Success-based) & Yes (Linear Reduction) & Yes & Yes (Sinusoidal $F$) & No \\ 
        \hline
        LSHADE-cnEpSin & Yes (Success-based) & Yes (Linear Reduction) & Yes & Yes (Sinusoidal $F$) & Yes (Covariance with EN) \\ 
        \hline
    \end{tabular}}
\end{table}

\section{Niching-based DE}\label{sec: niching DE}
The niching strategies in DE are essential for addressing MMOPs, particularly in maintaining diversity and ensuring the identification of multiple optima. This section explores key niching techniques, including speciation, fitness sharing, crowding, and the hill-valley method, which are designed to preserve population diversity and guide the evolutionary process toward discovering several optima across the search space. Each of these methods offers a distinct approach to handling multimodal landscapes by segregating the population based on specific criteria or interactions.

\subsection{Speciation}
Speciation has proven effective for multimodal optimization~\cite{li2002species,li2004adaptively,petrowski1996clearing}. A speciation-based niching method classifies an EA population based on similarity, measured by Euclidean distance:  
\[
\text{dist}(\mathbf{x}_{i}, \mathbf{x}_{j}) = \sqrt{\sum_{k=1}^{D} \left( x_{i, k} - x_{j, k} \right)^2},
\]
where \(\mathbf{x}_{i}\) and \(\mathbf{x}_{j}\) represent two \(D\)-dimensional individuals. Speciation depends on a radius parameter \(r_s\), defining the Euclidean distance from a species center (seed) to its boundary. The species seed, the fittest individual within a species, includes all individuals within distance \(r_s\). The algorithm for identifying species seeds~\cite{li2002species,petrowski1996clearing} iterates through a fitness-sorted list \(L_{\text{sorted}}\) of individuals. Starting with an empty seed set \(S = \emptyset\), each individual is checked against existing seeds in \(S\). If it lies beyond \(r_s\) of all current seeds, it becomes a new seed and is added to \(S\). This algorithm identifies species at each iteration, allowing DE to operate independently within each species. The steps for determining species seeds are summarized in Algorithm~\ref{algo: speciation}.
\begin{algorithm}[h!]
\caption{Identification of Species Seeds~\cite{li2005efficient}}\label{algo: speciation}
\begin{algorithmic}[1]
\REQUIRE $L_{\text{sorted}}$: A list of all individuals ordered by fitness in decreasing order
\ENSURE $S$: A collection of individuals that serve as species seeds
\STATE $S \gets \emptyset$ \COMMENT{Initialize species seed set}
\WHILE{not at the end of $L_{\text{sorted}}$}
    \STATE Choose the best unprocessed individual $p \in L_{\text{sorted}}$
    \STATE $\text{found} \gets \FALSE$
    \FORALL{$s \in S$}
        \IF{$d(s, p) \leq r_s$}
            \STATE $\text{found} \gets \TRUE$
            \STATE \textbf{exit the loop}
        \ENDIF
    \ENDFOR
    \IF{not $\text{found}$}
        \STATE $S \gets S \cup \{p\}$ \COMMENT{Add individual $p$ to species seeds}
    \ENDIF
\ENDWHILE
\end{algorithmic}
\end{algorithm}

A bare-bones niching DE (BNDE)~\cite{gong2017learning} is proposed for locating multimodal optima. BNDE leverages Gaussian bare-bones DE (GBDE)~\cite{wang2013gaussian} as its baseline, eliminating the need for fine-tuning traditional control parameters in evolutionary algorithms (EAs). A neighborhood niching strategy is introduced, employing Gaussian mutation with a local mean (\(\mu = \mathbf{x}_{nbest}\)) and standard deviation (\(\sigma\), calculated as the difference between \(\mathbf{x}_{nbest}\) and \(\mathbf{x}_i\)), to effectively capture niches that align with the contours of landscape peaks. To maintain diversity and enhance global exploration, BNDE incorporates a diversity-preserving operator. This operator calculates the neighborhood radius: \begin{equation}
 r_k = ||C_k, \mathbf{x}_{nrand}||,  
\end{equation} where \(C_k\) is the neighborhood center, to reinitialize converged or overlapping neighborhoods, ensuring sustained exploration and robust optimization.

 Liu et al.~\cite{liu2021double} introduced the double-layer-clustering speciation DE (DLCSDE) strategy. DLCSDE employs a dual-layer search approach. In the first layer, a species-based clustering division divides the population into several subpopulations, with each subpopulation containing \(M\) individuals, where only \(M-1\) participate in generating offspring using the basic DE algorithm, while one is designated as the species seed. Over iterations, subpopulations gravitate toward nearby peaks, with each subpopulation having a unique species seed selected. In the second layer, these species seeds collectively form a new subpopulation that conducts a global DE-based search, aiming to uncover peaks not addressed in the initial clustering. The second layer employs the global DE algorithm within the new subpopulation, supplemented by a self-adaptive parameter strategy inspired by the JADE~\cite{zhang2009jade} parameter adaptation mechanism.

% Addressing the core challenge of the speciation niching technique involves striking a delicate balance between local solution exploitation and global exploration. Hui and Suganthan~\cite{hui2016ensemble} proposed a novel speciation variant, ensemble, and arithmetic recombination-based speciation DE (EARSDE), to augment exploration by employing arithmetic recombination alongside speciation, thereby enhancing the exploitation of individual peaks through neighborhood mutation within ensemble strategies. This approach eliminates the need for radius parameterization, instead utilizing arithmetic recombination within a state-of-the-art ensemble DE framework. Subsequently, speciation relies on regional information among neighbors within a local neighborhood, obviating the use of fixed distance-based parameters.

The core challenge of speciation niching techniques is to effectively balance local exploitation with global exploration. Hui and Suganthan~\cite{hui2016ensemble} introduced a novel speciation variant, EARSDE (arithmetic recombination-based speciation DE), to enhance exploration by integrating arithmetic recombination with speciation. This strategy improves the exploitation of local peaks through neighborhood-based mutation within an ensemble DE framework. By eliminating the need for radius-based parameters, EARSDE leverages arithmetic recombination and focuses speciation on regional information derived from neighboring solutions. This shift removes the dependency on fixed distance-based parameters while still maintaining effective neighborhood-based search.

FBK-DE~\cite{lin2019differential}, a DE variant for MMOPs, uses Formulation, Balance, and Keypoint strategies for species division and evolution. Nearest-better clustering (NBC)~\cite{preuss2010niching} (See Algorithm 1 of the \textit{Supplementary file}) divides the population into species with a minimum size constraint, called NBC-Minsize (Algorithm~\ref{algo: NBC-minsize}), controlled by the parameter \(\texttt{minsize}\) (\(\texttt{minsize}(g)=g/2+5\)), which starts small for early-stage diversification and grows as species converge near optima. This scaling allows smaller species to merge into those near global optima. A species balance strategy ensures that no single species dominates in size. To evolve each species, the first term in \textit{DE/rand/1} and \textit{DE/rand/2} is replaced by a \textit{keypoint} \(\mathbf{x}_{kp}\), enabling targeted evolution with \textit{DE/keypoint/1} and \textit{DE/keypoint/2} operators, alongside traditional DE operators.

\begin{algorithm}
    \caption{NBC-Minsize~\cite{lin2019differential}}\label{algo: NBC-minsize}
    \begin{algorithmic}[1]
    \STATE Calculate $\texttt{minsize} = g / 2 + 5$ and construct the spanning tree $T$
    \STATE Calculate the mean distance $\mu_{\text{dist}}$ and calculate the $follow$ vector
    \STATE Sort the edges in $T$ from the longest to the shortest  
    \FOR{each edge $\in T$}
        \IF {$\text{dist}(edge) > \phi \cdot \mu_{\text{dist}}$}
            \STATE Set $edge_f$ to the follower individual of edge;
            \STATE Set $edge_r$ to the root of the subtree containing $edge_f$
            \IF{$\text{follow}(edge_f) \geq \texttt{minsize}$ \textbf{and} $\text{follow}(edge_r) - \text{follow}(edge_f) \geq \texttt{minsize}$}
                \STATE Cut off $edge$ and set $edge_l$ to the leader individual of $edge$
                \FOR{each $\mathbf{x}$ on the path from $edge_l$ to $edge_r$}
                    \STATE $\text{follow}(\mathbf{x}) = \text{follow}(\mathbf{x}) - \text{follow}(edge_f)$
                \ENDFOR
            \ENDIF
        \ENDIF
    \ENDFOR
    \end{algorithmic}
\end{algorithm}

A novel DE with species conservation, termed MMODE/SC, is introduced~\cite{ji2023multimodal} for MM-MOOPs to locate diverse PSs efficiently. The algorithm integrates species conservation for preserving PSs in identified regions and a variant of \textit{DE/rand/1} to explore new regions. Species conservation incorporates three key operators: (1) \textit{Species Division}, which partitions the population into species in decision space to retain distinct PSs. Solutions are initially classified into species centers (\textit{choose}) and centers (\textit{center}). The species radius, \( r_s \), is adaptively calculated as:\begin{equation}
   r_s = \frac{\sum_{\mathbf{x} \in \mathbf{P}} \sum_{\mathbf{y} \in \mathbf{Pns(\mathbf{x})}} \text{dist}(\mathbf{x}, \mathbf{y})}{N_p \cdot Nns},
\end{equation} where \( Pns \) represents the set of \( Nns \) nearest neighbors of each solution. Species are created in two cases: unassigned solutions either form a new species based on a probability threshold or join the nearest existing species. When the number of solutions exceeds \( N_p \), non-dominated sorting and harmonic average distance (HAD) (defined in Section 1 of the \textit{Supplementary file}) are used to select a solution from the last non-dominated set. (2) \textit{Seed Determination} selects the top solutions from each species as seeds, promoting diversity to cover multiple PSs. A minimum species size is enforced to ensure convergence. (3) \textit{Seed Conservation} guarantees that all species seeds are retained, preventing the loss of regions with potential PSs and maintaining population diversity.

NetCDE$_\text{MMOPs}$~\cite{chen2023network} introduces a network community-based DE for MMOPs by representing the population as a network where individuals serve as nodes, inverse distances between them as edges, and recent historical data as node attributes. NetCDE$_\text{MMOPs}$ incorporates three key strategies: (1) Partitions the population into niches using community detection without sensitive parameters, transforming the population into an attributed network $G=(\mathbf{P}, \mathbf{E},\mathbf{W}, \mathbf{H}, \mathbf{A})$\footnote{Where, $\mathbf{P}=\{\mathbf{x}_1,\mathbf{x}_2,\ldots,\mathbf{x}_n\}$ is the set of the whole population, $\mathbf{E}=\{(\mathbf{x}_i,\mathbf{x}_j)|\mathbf{x}_i\in N, \mathbf{x}_j \in N, i\neq j\}$, $\mathbf{W}=\{w_{ij}=(\mathbf{x}_i,\mathbf{x}_j)\}$ is the adjacent matrix, where $w_{ij}$ is defined as the inverse distance between node $\mathbf{x}_i$, and $\mathbf{x}_j$; $\mathbf{H}=\{h_1,h_2,\ldots,h_n\}$ is the set of historical position information of all nodes, and $\mathbf{A}=\{a_1,a_2,\ldots,a_n\}$ the set of historical fitness information of all nodes.} and divide this population into several niches through community detection. (2) Community elites-based updating (CEU) allocates resources to community elites, selected as the top $N_p$ individuals by fitness due to their strong exploitation ability. This selective focus on elite exploration conserves $FEs$ and enhances population convergence. (3) Poor Individual Remolding (PIR) guides lower-ranking individuals toward promising areas. Parameter $s$ (0 to 1) marks the switch from exploration to exploitation when $s\times MaxFE$, $FE$ are reached. Parameter $r$ defines non-poor proportions, categorizing the top $r\times N_p$ sorted individuals as non-poor while others are considered poor.

Huang et al.~\cite{huang2017niching} introduced the niching community-based SDE (NCSDE) algorithm to enhance multimodal optimization. It augments the species-based DE (SDE)~\cite{li2005efficient} with the niching community and one-to-one greedy selection strategies, yielding several advantages: (i) Elimination of the need for prior knowledge to determine niching parameters. (ii) Improved precision in locating multiple optima through confinement within small niching communities. (iii) Preserving diversity via one-to-one greedy selection prevents genetic variability loss. NCSDE, akin to SDE, selects the fittest individual as the species seed but employs a fixed small-sized group based on Euclidean distance, gathering similar individuals into niching communities. The conventional practice of preserving the fittest NP individuals can dilute niche information by neglecting diversity. Our proposed niching strategy establishes fixed-sized niching communities, ensuring offspring always equals NP while employing one-to-one greedy selection to maintain diversity.

A self-organizing map (SOM)-based DE with dynamic search (SOMDE-DS) is proposed~\cite{yuan2022self} to address MMOPs and enhance DE performance. The SOM-based niching technique leverages similar information among individuals to partition the population effectively. A variable neighborhood search (VNS) strategy expands the search space to identify more potential optimal regions, while a dynamic selection (DS) strategy balances exploration and exploitation. 

The WSN-based adaptive DE (WSNADE) algorithm~\cite{huang2024wireless} for solving MMOPs by drawing inspiration from wireless sensor networks (WSNs), incorporating an adaptive niching technique (WANT) and two strategies: protection-based dual-scale mutation and multi-level reset. In WANT, each individual \(\mathbf{x}_i\) identifies related individuals whose monitoring areas intersect directly or indirectly with its own, and grades them based on the strength of the intersection. Directly intersecting individuals form the first level (\(L_1\)), while indirectly intersecting individuals form subsequent levels (\(L_2\), etc.). Unrelated individuals are labeled \(L_{\text{no}}\). To form a niche \(N_i\) for \(\mathbf{x}_i\), the process begins with only \(\mathbf{x}_i\) in the niche, then calculates the monitoring radius \(R_i\) as half the mean distance to the \(nSize\) nearest individuals:
\begin{equation}
 R_i = \frac{\sum_{j=1}^{nSize} \text{dist}(\mathbf{x}_i, \mathbf{x}_j)}{2 \times nSize},  
\end{equation}
where \(\mathbf{x}_j\) is the \(j\)th nearest individual and \(\text{dist}(\mathbf{x}_i, \mathbf{x}_j)\) is the Euclidean distance. This radius is shared with all individuals to identify related ones. If the niche size is less than three, \(R_i\) is expanded by a factor \(\phi\) to ensure the minimum required size. Within the niche, individuals evolve using mutation strategies suited to their fitness. Better-fit individuals use a small-scale mutation (\textit{DE/current/1} with \(F_{\text{min}} = 0.5\)) to refine solutions, while worse-fit individuals use large-scale mutations (\textit{DE/current/1 or DE/rand/1} with \(F_{\text{max}} = 0.9\)) to explore more regions. The probability \(p_i\) of using the small-scale strategy is based on the fitness rank of \(\mathbf{x}_i\) within the niche, calculated as \(p_i = \frac{\text{rank}_i}{|N_i|}\), where \(\text{rank}_i\) is the fitness rank of \(\mathbf{x}_i\) and \(|N_i|\) is the niche size. The stagnation and exclusion mechanisms are used in a multi-level reset strategy. A novel niche center identification (NCI) technique~\cite{liang2024niche}, Algorithm 2 of the \textit{Supplementary file}, is proposed and integrated with JADE~\cite{zhang2009jade} to form NCIDE. Addressing the challenge of identifying suitable niche centers, requiring high fitness (fitness aspect) and diverse distribution across search regions (distance aspect), NCI selects niche centers based on fitness and distance criteria, grouping non-center individuals into niches with their nearest centers. 

Additionally, a niche-level archival-adaptive parameter scheme (NAAPS) is introduced, incorporating an information matching mechanism (IMM) that leverages historical data to adjust parameters at the niche level, reducing sensitivity dynamically. An archive mechanism further enhances exploration by preserving identified optima and reinitializing stagnant individuals. Wang et al.~\cite{wang2022adaptive} introduced the adaptive estimation distribution DE (AED-DDE), a parameter-free distributed niching method. AED-DDE combines three crucial mechanisms: adaptive estimation distribution (AED), a master-slave multiniche distributed model, and probabilistic local search (PLS). In the AED mechanism, each individual denoted as \(\mathbf{x}_i\), forms an initial niche \(N_i\) by selecting its three nearest neighbors, adhering to the requirement of at least four individuals for DE (\(M_i\), the initial niche size of \(\mathbf{x}_i\), is set as 3). The distribution of the niche is estimated based on a univariate Gaussian distribution. The distribution's suitability is determined by verifying if the fourth nearest neighbor, \(\mathbf{x}_i^{4th}\), lies within the range \(\mu_i \pm 3\sigma_i\). If \(\mathbf{x}_i^{4th}\) falls outside this range, indicating significant dissimilarity from \(\mathbf{x}_i\), it is excluded from the niche, maintaining a size of 3 to ensure a productive evolutionary process. Conversely, if \(\mathbf{x}_i^{4th}\) is close to \(\mathbf{x}_i\), it is added to the niche, expanding its size by 1. This adaptive process continues until the niche size stabilizes. 

Following the adaptive formation of niches for each individual via AED, offspring are generated and selected within the master-slave multiniche distributed framework. The master allocates each niche to its corresponding slave, facilitating the concurrent co-evolution of different niches. Each niche, represented by a slave node, generates a trial vector \(\mathbf{u}_i\) using the standard DE mutation and crossover operators. After generating \(\mathbf{u}_i\), each slave returns it to the master for the selection phase. The master combines the newly generated trial vectors and organizes a competition, pitting each \(\mathbf{u}_i\) against its nearest parent \(\mathbf{x}\) from the entire population. To enhance solution accuracy, a PLS is employed, generating a new individual, \(\mathbf{x}_{new}\), in proximity to \(\mathbf{x}\). The standard deviation \(\sigma\) is initially large and gradually reduces during evolution to balance diversity and convergence. A local search probability is defined based on the individual ranks according to their fitness values, encouraging the acceleration of convergence toward global optima, as better individuals are more likely to be closer to them.

\subsection{Fitness Sharing}
Fitness sharing is a widely used niching method, first introduced by Holland~\cite{holland1992adaptation} and later refined to divide populations into subpopulations based on individual similarity~\cite{goldberg1987genetic}. Inspired by natural resource sharing, it adjusts an individual's fitness based on the density of nearby individuals within the same niche. This reduces fitness for individuals in crowded regions, discouraging overcrowding in a single niche and promoting diversity. Consequently, it rewards individuals exploring distinct regions of the search space, ensuring a balanced population across multiple niches~\cite{della2007niches,miller1996genetic,li2007multimodal}. The following discussion explores its application in DE. To improve the convergence, DHNDE~\cite{wang2022multimodal} includes an enhanced neighborhood speciation-based DE (INSDE) that identifies and removes highly similar individuals, conserving computational resources (See Algorithm 3 of the \textit{Supplementary file}). An optimal solution archive also prevents the loss of top solutions during the evolutionary procedure. Zhao et al.~\cite{zhoa2020local} introduced local binary pattern-based adaptive DE (LBPDE) to efficiently solve MMOPs by leveraging local binary patterns (LBP)\footnote{LBP~\cite{ojala2002multiresolution}, a well-established texture analysis technique in image processing and computer vision. LBP operates by comparing a central pixel's intensity to its neighboring pixels within a specified neighborhood, generating a binary pattern that encodes local texture details.} to extract information from neighbors, forming multiple niches that correspond to regions of interest (or peaks). The authors introduced a niching and global interaction (NGI) mutation strategy, which combines both local niche information and global population information. It relies on a set, denoted as $S$, which comprises individuals with equal or superior fitness values compared to others (i.e., $\mathbf{S}=\{\mathbf{x}_j|f(\mathbf{x}_j)\geq f(\mathbf{x}_i)\}$). When $|\mathbf{S}|\geq 1$, the offspring uses $\mathbf{x}_{nbest}$ to guide $\mathbf{x}_i$, enabling rapid convergence towards potential optima. To avoid entrapment in local optima, the NGI mutation strategy introduces global perturbation, randomly selecting $g_{r1}$ and $g_{r2}$ from the entire population. 

In cases where $\mathbf{x}_i$ is the best in the LBP-based niche (i.e., $|S|=0$), implying proximity to an optimum, two individuals are randomly chosen from the $M$-nearest neighbors to provide local exploitation information for $\mathbf{x}_i$. The APS mechanism dynamically adjusts parameters $(F,~CR)$ based on an individual's fitness and its LBP-based niche. Leveraging LBP-based niche information, the algorithm uses neighbors' information to determine the most promising evolution direction in MMOPs. Specifically, a larger $|S|$ indicates a need for greater learning from better individuals, hence a larger $F_i$. Conversely, when $|S|$ is smaller, indicating a promising fitness level for the individual, less learning is required, suggesting a smaller $F_i$. Therefore, the value of $|S|/|M|$ guides the parameter $F_i$ of the current individual. For a balanced trade-off between diversity and convergence, $\mathbf{x}_i$'s parameter $CR_i$ is adaptively controlled based on the distribution of individuals within its LBP niche. The NGI mutation strategy efficiently explores both niching and global areas, while an adaptive parameter strategy (APS) fine-tunes individual parameters guided by their LBP information, facilitating movement toward promising directions. A penalty-based DE, named PMODE~\cite{wei2021penalty}, is developed for MMOPs, incorporating a dynamic penalty strategy (other penalty strategies~\cite{pourjafari2012solving,hirsch2009solving} are discussed in Section 2 of the \textit{Supplementary file}) with two main components: a penalty function and a dynamic penalty radius. The penalty function penalizes solutions that are within the proximity of elite solutions, defined as follows:\begin{equation}
\textbf{Max}~PF(\mathbf{x}) = f(\mathbf{x}) \cdot \prod_{i=1}^{|\mathbf{S}|} G_r(d_n), \quad d_n = ||\mathbf{x} - \mathbf{x_n^*}||
\end{equation}
The function \( G_r(d_n) \) is governed by a dynamic radius \( R_g \), calculated as \( R_g = r_g \cdot R \), where \( r_g = |\alpha \cdot \sin(\text{freq} \cdot \pi \cdot \alpha^{1/2})| + \tau \alpha \). Based on specific conditions, \( G_r(d_n) \) takes the following forms: If \( f(\mathbf{x}) \geq 0 \) and \( d_n \leq R_g \), then \( G_r(d_n) = \tan^{-1}(\epsilon \cdot d_n) \). If \( f(\mathbf{x}) < 0 \) and \( d_n \leq R_g \), then \( G_r(d_n) = 1/\tan^{-1}(\epsilon \cdot d_n) \). Otherwise, \( G_r(d_n) = 1 \). In these equations, $|\mathbf{S}|$ represents the number of elite solutions, $\mathbf{x_n^*}$ is the \( n \)th elite solution in \( \mathbf{S} \), and \( \epsilon \) is a small constant. An elite selection mechanism, governed by a parameter \( \phi \), identifies elite solutions and applies penalties to their neighboring regions. PMODE employs JADE as its search engine.

\subsection{Crowding}
The crowding method uses a competition mechanism between offspring and their parents to maintain diversity and prevent premature convergence. Unlike fitness sharing, it adjusts selection pressure by comparing an offspring to a small random sample from the population and replacing the most similar individual~\cite{de1975analysis}. Mahfoud~\cite{mahfoud1992crowding,mahfoud1993simple,mahfoud1995niching} identified issues with the crowding method~\cite{de1975analysis} and developed deterministic crowding (DC), which eliminates the crowding factor and reduces replacement errors. DC can effectively maintain multiple peaks without needing prior knowledge of peak count or niche radius. Algorithm~\ref{algo: DC} outlines the DC procedure.
\begin{algorithm}[h!]
\caption{Pseudo-code of DC}\label{algo: DC}
\begin{algorithmic}[1]
\STATE Select two parents, $p_1$ and $p_2$ randomly, without replacement
\STATE Generate two offspring, $c_1$ and $c_2$
\IF{$d(p_1, c_1) + d(p_2, c_2) \leq d(p_1, c_2) + d(p_2, c_1)$}
    \IF{$f(c_1) > f(p_1)$}
        \STATE Replace $p_1$ with $c_1$
    \ENDIF
    \IF{$f(c_2) > f(p_2)$}
        \STATE Replace $p_2$ with $c_2$
    \ENDIF
\ELSE
    \IF{$f(c_2) > f(p_1)$}
        \STATE Replace $p_1$ with $c_2$
    \ENDIF
    \IF{$f(c_1) > f(p_2)$}
        \STATE Replace $p_2$ with $c_1$
    \ENDIF
\ENDIF
\end{algorithmic}
\end{algorithm}

\begin{algorithm}[h!]
\caption{Pseudo-code for CrowdingDE~\cite{thomsen2004multimodal}}\label{algo: crowdingDE}
\begin{algorithmic}[1]
\WHILE{termination criteria are not met}
    \FOR{each individual $i$ in the population}
        \STATE Use standard DE to produce an offspring $\mathbf{x}_{i}$
        \STATE Calculate the Euclidean distances of $\mathbf{x}_{i}$ to all other individuals in $\mathbf{P}$
        \STATE Sort all individuals based on their Euclidean distances to $\mathbf{x}_{i}$
        \STATE Let $\mathbf{x}_{\text{closest}}$ be the individual with the smallest Euclidean distance to $\mathbf{x}_{i}$
        \IF{$f(\mathbf{x}_{i}) > f(\mathbf{x}_{\text{closest}})$}
            \STATE Replace $\mathbf{x}_{\text{closest}}$ with $\mathbf{x}_{i}$
        \ENDIF
    \ENDFOR
    \STATE Proceed to the next iteration if the population size is reached
\ENDWHILE
\end{algorithmic}
\end{algorithm}

Thomsen~\cite{thomsen2004multimodal} demonstrated that CrowdingDE (general procedure given in Algorithm~\ref{algo: crowdingDE}) outperforms a fitness-sharing DE variant on MMOPs. While CrowdingDE is simple to implement, it incurs a higher computational cost due to the need to compare each offspring with all individuals in the population for similarity, resulting in a complexity of \( O(N^2) \). A two-archive DE~\cite{wang2022multimodal}, termed DHNDE, is introduced using a dynamically hybrid niching method that combines crowding and speciation techniques. A secondary archive stores inferior offspring, which enhances diversity when incorporated with crowding-based DE (CDE$/\mathbf{A}_{io}$). In the initial step, three individuals are randomly selected from \( \mathbf{P} \) to generate offspring \( \mathbf{u}_i\) for each target parent \( \mathbf{x}_i\) using the \textit{DE/rand/1} mutation to maintain diversity. Each offspring \( \mathbf{u}_i\) is then compared to its closest parent \( \mathbf{y} \) based on Euclidean distance. If \( \mathbf{u}_i \) performs better, it replaces \( \mathbf{y} \); otherwise, it is added to the archive \( \mathbf{A}_{io} \). 

Unlike traditional CDE methods~\cite{thomsen2004multimodal}, CDE/\( \mathbf{A}_{io} \) retains discarded offspring in \( \mathbf{A}_{io} \), with a maximum size \( \text{max}_{io} \), as given in Algorithm~\ref{algo: CDE_Aio}. When \( |\mathbf{A}_{io}| \geq \text{max}_{io} \), a specialized procedure is triggered to enhance convergence.
\begin{algorithm}
\caption{CDE$\backslash \mathbf{A}_{io}(\mathbf{P}, ~\mathbf{A}_{io},~ FE)$~\cite{thomsen2004multimodal}}\label{algo: CDE_Aio}
\begin{algorithmic}[1]
\FOR{$i = 1$ to $N_p$}
    \STATE Randomly choose 3 individuals from $\mathbf{P}$;
    \STATE Generate an offspring $\mathbf{u}_i$ by the mutation and crossover of DE;
    \STATE Evaluate $\mathbf{u}_i$;
    \STATE $FE = FE + 1$;
\ENDFOR
\FOR{$i = 1$ to $N_p$}
    \STATE Find the most similar individual $\mathbf{y}$ in $\mathbf{P}$ to $\mathbf{u}_i$;
   \IF{$f(\mathbf{u}_i) > f(\mathbf{y})$}
        \STATE $\mathbf{y} \gets \mathbf{u}_i$;
    \ELSE
        \STATE $\mathbf{A}_{io} = \mathbf{A}_{io} \cup \mathbf{u}_i$;
    \ENDIF
\ENDFOR
\end{algorithmic}
\end{algorithm}
Awad et al.~\cite{awad2018ensemble} introduce an enhanced DE algorithm (ESDE-NR), termed LSHADE-EpSin~\cite{awad2016ensemble}, to enhance DE's performance. It achieved a balance between exploration and exploitation by blending two sinusoidal formulas and a Cauchy distribution. Additionally, a restart method is implemented in later generations when the population size is reduced by 20 individuals. This technique enhances solution quality by reinitializing the 10 worst individuals while allowing the remaining 10 to employ the restart strategy. Furthermore, the algorithm introduces an innovative approach to adapt the population size using a niching-based reduction scheme. This mechanism utilizes two separate niches before implementing population reduction and effectively reduces the population size while maintaining solution diversity. Huang et al.~\cite{huang2018hypercube} introduced a hypercube-based crowding DE with neighbourhood mutation to address multimodal optimization challenges. Notably, this approach utilizes hypercube-based neighborhoods instead of simpler Euclidean-distance-based ones. 

Additionally, the authors employ a self-adaptive technique to regulate the hypercube's radius vector, ensuring a consistently reasonable neighbourhood size. Consequently, the algorithm conducts precise searches in densely populated sub-regions while adopting a more randomized approach in sparsely populated areas. An outlier-aware DE algorithm (OADE)~\cite{zhao2023outlier} is introduced, employing an outlier detection mechanism based on the \(K\) nearest neighbors of an individual, \(\mathbf{x}_i\), to form a niche set \(\mathbf{Q}_i\). A threshold is set using the average distance across the population. The algorithm's dimension and guidance-balanced mutation (DGM) strategy includes two new terms, \(\mathbf{x}_{nbest}\) (the neighborhood best), and \((\mathbf{x}_{n\_r1}-\mathbf{x}_{n\_r2})\), where \(\mathbf{x}_{n\_r1}\) and \(\mathbf{x}_{n\_r2}\) are distinct individuals within \(\mathbf{Q}_i\). For problems with dimensions \(d \leq 3\), \(\mathbf{x}_{nbest}\) is inactive to refine solution accuracy better. Upon detecting an outlier, an outlier-based selection (OBS) strategy promotes diversity by combining fitness and distribution data, enhancing multimodal peak detection. In high-dimensional spaces, an inactive outlier-based re-initialization (IOR) strategy is applied to help individuals escape local optima.

\subsection{Hill-Valley Method}
The Hill-Valley technique, proposed by Ursem~\cite{ursem1999multinational}, uses population topological information to form species. It detects valleys between two individuals \( \mathbf{x}_A \) and \( \mathbf{x}_B \) by the condition in Definition~\ref{def: hill valley}.\begin{definition}\label{def: hill valley}
 There is a valley between \( \mathbf{x}_A \) and \( \mathbf{x}_B \) if there exists a reference point \( \mathbf{x}_C \in \Omega \) such that \( f(\mathbf{x}_C) < \min\{f(\mathbf{x}_A), f(\mathbf{x}_B)\} \), where \( \Omega = \{\mathbf{x}_C \mid \mathbf{x}_C = \mathbf{x}_A + \lambda (\mathbf{x}_B - \mathbf{x}_A), \lambda \in (0, 1)\} \) is the set of reference points.   
\end{definition} 

If such a valley exists, \( \mathbf{x}_A \) and \( \mathbf{x}_B \) are considered to be on different peaks. However, identifying the correct reference point is challenging and computationally expensive since the fitness of \( \mathbf{x}_C \) must be evaluated. To address this, Yao et al.~\cite{yao2006clustering} proposed a dichotomy method to sample reference points, and Li et al.~\cite{li2014history} used historical solutions as reference points. While the latter approach reduces computational cost, it may lead to errors early in the algorithm when only a few historical solutions are available. Damanahi et al.~\cite{damanahi2016novel} introduced a novel method for effectively addressing high-dimensional multimodal problems by employing a roaming method to create parallel sub-populations, each assigned random improved proposing strategies and utilizing the variable parameter setting DE method. They incorporate the Hill-Valley method %\footnote{In Hill-Valley, for two individuals $\mathbf{x}_A$ and $\mathbf{x}_B$, if $\exists~ \mathbf{x}_C \in C$ such that $f(\mathbf{x}_C) < \min\{f(\mathbf{x}_A), f(\mathbf{x}_B)\}$, then $\mathbf{x}_A$ and $\mathbf{x}_B$ are on different peaks, where $C$ is a set of reference points and $C = \{\mathbf{x}_C|\mathbf{x}_C = \mathbf{x}_A + \lambda \mathbf{x}_B,~\lambda \in (0, 1)\}$. However, it is difficult to find the right reference point. Meanwhile, this method is expensive since the fitness of $\mathbf{x}_C$ is calculated.}
to determine if two points belong to the same species. In their subsequent work, Damanahi et al.~\cite{damanahi2016improved} introduced an accurate solution for high-dimensional multimodal problems using DE with parallel subpopulations generated by the roaming algorithm. This method offers flexibility in subpopulation count and ensures stability by classifying subpopulations as stable or unstable based on their evolutionary progress. Un-evolved subpopulations are iteratively evolved until stability is achieved, yielding local optima which are archived externally. This approach also features the Hill-Valley method, which reduces function evaluations, mitigates overhead from surplus individuals, and improves accuracy in locating optima across various dimensions. Building on this, a new DE approach (ESPDE)~\cite{li2023history} for MMOPs is introduced based on an improved Hill-Valley technique, which leverages historical information to classify individuals on the same peak as a species. In this technique, for two individuals \( \mathbf{x}_A \) and \( \mathbf{x}_B \), if there exists \( \mathbf{x}_C \in C \) such that \( f(\mathbf{x}_C) < \min\{f(\mathbf{x}_A), f(\mathbf{x}_B)\} \), then \( \mathbf{x}_A \) and \( \mathbf{x}_B \) are deemed to be on different peaks, where \( C \) is a set of reference points defined as:
\begin{equation}
 C = \{\mathbf{x}_C \mid \min\{\mathbf{x}_{A,j}, \mathbf{x}_{B,j}\} < \mathbf{x}_{C,j} < \max\{\mathbf{x}_{A,j}, \mathbf{x}_{B,j}\}, \; j=1,2,\ldots,N_p\}.   
\end{equation}

Unlike the original Hill-Valley technique~\cite{ursem1999multinational}, this improved version is parameter-insensitive and avoids extra computational costs by utilizing historical data. Specifically, it identifies whether two individuals reside on the same peak by examining eliminated individuals.
For each species, an evolutionary state recognition method assesses the evolutionary phase, with distinct strategies tailored for exploration (\(||\sigma||_{-\infty} > 0.1\)), exploitation (\(0.0001 < ||\sigma||_{-\infty} < 0.1\)), and convergence (\(||\sigma||_{-\infty} < 0.0001\)) phases. Additionally, a prediction mechanism forecasts each species' potential, facilitating efficient resource allocation. Similarly, A history archive-assisted niching DE with the variable neighborhood (HANDE/VN) is proposed~\cite{liao2023history} for MMOPs. A variable neighborhood strategy dynamically adjusts neighborhood size as $m = randint(m_1, m_2)$, where $m_1$ and $m_2$ vary by iteration, balancing exploration and exploitation. The history archive $\rho$ stores individuals who lose in selection; if $\rho$ has fewer than $2 *N_p$ entries, the losing individual is added, with fitness-based comparisons ensuring only the better one remains. A mutation operation leveraging the archive enhances information exchange between the archive and the population, improving search accuracy for multiple global solutions. Additionally, a history archive-based mechanism manages individuals trapped in local optima. Additionally, Li et al.~\cite{li2021adaptive} proposed an adaptive subpopulation-based niching DE (SHVDE) to address challenges in finding reference points for the Hill-Valley niching. They modified the reference set \( C \) to make the method computationally more efficient and introduced an optimal solution recognition technique based on the distribution of individuals. 

SHV selects individuals from the eliminated ones based on crowding. When the subpopulation size falls below 5, the authors generate individuals using $\mathbf{x}_{temp}=Gauss(\mathbf{x}_{best},0.1)$. This technique calculates the standard deviation of individuals in a subpopulation and, when the standard deviation is below a threshold, removes the subpopulation and adds the best individual to the optimal solution set. This combination of approaches, ranging from dynamic neighborhood adjustment and adaptive subpopulation size to history archive mechanisms, effectively enhances DE's performance for MMOPs by promoting stability, resource efficiency, and accurate global solution searches. A summary of selected niching-based DE variants is presented in Table~\ref{tab: niching DE}, highlighting their key results, specific applications or the MMOPs they addressed, and possible future research directions (specified in the original papers). This table provides a comprehensive overview of how these variants leverage niching techniques to tackle challenges such as maintaining diversity, improving convergence, and identifying multiple optimal solutions in various problem domains.
\begin{table}[]
    \centering
        \caption{Summary of niching-based DE.}
    \label{tab: niching DE}
\resizebox{1\linewidth}{!}{
    \begin{tabular}{c|l|c|p{6.5cm}|p{3.5cm}|p{4cm}|p{4.5cm}}\hline
S. No. & Variants & Year & Description & Applications & Results & Future Work \\\hline
1 & BNDE~\cite{gong2017learning} & 2017 & Combines bare-bones DE with Gaussian mutation for local search and diversity-preserving operator. & 20 benchmark suites, neural network ensembles. & Consistently superior across MMOPs. & Improve local search, and explore ML applications. \\\hline
2 & DLCSDE~\cite{liu2021double} & 2021 & Uses two-layer clustering for global and local optima, with a self-adaptive strategy. & 29 MMOPs. & Matches or outperforms 17 algorithms. & Extend to other niching methods, higher dimensions. \\\hline
3 & EARSDE~\cite{hui2016ensemble} & 2016 & Enhances exploration with arithmetic recombination, speciation, and ensemble strategies. & Solves MMOPs with distributed solutions. & Competitive on 29 MMOPs. & Apply to real-world MMOPs for efficiency. \\\hline
4 & FBK-DE~\cite{lin2019differential} & 2019 & Uses NBC for species division and keypoint-based mutation for balance between exploration and exploitation. & MMOPs requiring balance in decision-making. & Competitive with 15 algorithms. & Improve for problems with many global peaks. \\\hline
5 & MMODE/SC~\cite{ji2023multimodal} & 2023 & Combines species division, seed determination, and conservation for diversity. & MM-MOOPs. & Competitive performance on CEC 2019 and real-world problems. & Enhance global/local PS search, and improve performance. \\\hline
6 & NetCDE$_\text{MMOPs}$~\cite{chen2023network} & 2023 & Uses community detection for niche division and CEU/PIR strategies for resource allocation. & MMOPs requiring high diversity. & Outperforms many recent algorithms. & Improve efficiency for complex MMOPs. \\\hline
7 & NCSDE~\cite{huang2017niching} & 2017 & Enhances DE with niching community and greedy selection strategy. & MMOPs with multiple optima. & Outperforms other niching algorithms. & Investigate adaptive niching strategy. \\\hline
8 & SOMDE-DS~\cite{yuan2022self} & 2022 & Combines SOM for division, VNS for space expansion, and DS for balance. & CEC 2013 MMOPs. & Outperforms several MMOP algorithms. & Study parameter interactions for improvements. \\\hline
9 & WSNADE~\cite{huang2024wireless} & 2024 & Uses WSN-based adaptive niching and dual-scale mutation for improved exploration. & MMOPs and real-world MCFLD problems. & Competitive on 20 CEC 2015 MMOPs. & Improve in complex environments, extend to new applications. \\\hline
10 & NCIDE~\cite{liang2024niche} & 2024 & Identifies niche centers using fitness and distance, with a NAAPS for parameter sensitivity. & Multimodal nonlinear equation systems. & Outperforms several state-of-the-art algorithms. & Refine niche identification, and extend to complex MMOPs. \\\hline
11 & AED-DDE~\cite{wang2022adaptive} & 2022 & Uses AED-based niching and multiniche distributed model for global optima discovery. & CEC 2015 MMOPs. & Outperforms CEC 2015 winner. & Test on more complex MMOPs and real-world applications. \\\hline
12 & DHNDE~\cite{wang2022multimodal} & 2022 & Integrates crowding and speciation with secondary archive and INSDE for convergence. & MMOPs with many optima. & Outperforms 17 methods on CEC-2013 MMOPs. & Improve parameter adaptability and balance. \\\hline
13 & ESDE-NR~\cite{awad2018ensemble} & 2018 & Uses sinusoidal formulas and Cauchy distribution for exploration-exploitation balance. & Real-parameter optimization (CEC 2014). & Outperforms CMA-ES variants in CEC 2016. & Refine strategies for complex problems. \\\hline
14 & OADE~\cite{zhao2023outlier} & 2023 & Introduces DGM, OBS, and IOR strategies for peak detection and accuracy in high-dimensional problems. & MMOPs and high-dimensional problems. & Outperforms competitive algorithms. & Extend to other algorithms and real-world applications. \\\hline
15 & LBPADE~\cite{zhoa2020local} & 2020 & Uses LBP for multiple niches and NGI mutation for efficiency. & Engineering design and image processing. & Competitive with state-of-the-art MMOP algorithms. & Utilize parallel computing for efficiency. \\\hline
    \end{tabular}}
\end{table}

\section{Clustering-based DE}\label{sec: clustering DE}
Clustering involves organizing data points into clusters such that points within the same cluster exhibit high similarity, while those in different clusters show high dissimilarity. This similarity or dissimilarity is often quantified using distance metrics like Euclidean or Mahalanobis distance. The clustering principles align closely with those of niching in optimization, as both aim to segregate entities based on specific attributes. In niching, data points can be viewed as individuals, and clusters are analogous to niches representing regions of interest in the search space. This alignment enables clustering methods to be effectively adapted for niching tasks, where identifying and maintaining multiple optima is crucial~\cite{li2016seeking}.

In this context, clustering-based DE methods have emerged as powerful tools for multimodal optimization. These methods leverage clustering techniques to divide the population into subgroups, enabling efficient exploration and exploitation of different regions of the search space. To provide a structured overview, clustering-based DE approaches can be broadly categorized into two groups: partition-based clustering and spanning tree-based clustering. Partition-based clustering divides the population into fixed regions based on predefined criteria while spanning tree-based clustering identifies groups using graph-based connectivity. The following subsections delve deeper into the methodologies and applications of these clustering-based DE algorithms.

\subsection{Cluster-based Partitioning Strategies}%{Cluster Pool Approach}
\begin{figure}[ht]
    \centering
    % Subfigure (a)
    \begin{subfigure}[t]{0.45\textwidth}
        \centering
        \begin{tikzpicture}[scale=1.2, every node/.style={font=\small}]
            % Nodes
            \node[draw,circle, minimum size=2pt] (candidate) at (0, 2) {};
            \node[above] at (candidate) {$\mathbf{x}_k$};
            \node[draw, circle] (individual) at (2, 0){};\node[below] at (individual) {$\mathbf{x}_i$};
            \node[draw, circle] (competing) at (2, 2){};\node[right] at (competing) {Competing candidate};
            \node[draw, circle](blank)  at (0,0){};
            % Arrows
            \draw[arrow, thick,line width=2pt, red, dashed] (individual)--(candidate) node[midway, above left,black] {$r(i, k)$};
            \draw[arrow, thick,line width=2pt] (competing) -- (individual)node[midway, above right] {$a(i, k')$};
            \draw[arrow, thick,line width=2pt] (blank) -- (individual);
            % Labels
            \node[above right] at (competing) {};
        \end{tikzpicture}
        \caption{Sending responsibilities.}
    \end{subfigure}
    \hfill
    % Subfigure (b)
    \begin{subfigure}[t]{0.45\textwidth}
        \centering
        \begin{tikzpicture}[scale=1.2, every node/.style={font=\small}]
            % Nodes
            \node[draw,circle, minimum size=2pt] (candidate) at (0, 2) {};
            \node[above] at (candidate) {$\mathbf{x}_k$};
            \node[draw, circle] (individual) at (2, 0){};\node[below] at (individual) {$\mathbf{x}_i$};
            \node[draw, circle] (supporting) at (2, 2){};\node[below] at (supporting) {Supporting Individual $i'$};
            \node[draw, circle](blank)  at (0,0){};
            % \node[draw, circle] (candidate) at (0, 2) {Candidate exemplar $k$};
            % \node[draw, circle] (individual) at (2, 0) {Individual $i$};
            % \node[draw, circle] (supporting) at (2, 2) {Supporting Individual $i'$};

            % Arrows
            \draw[arrow, thick,line width=2pt, red, dashed] (candidate) -- (individual) node[midway, below left,black] {$a(i, k)$};
            \draw[arrow, thick,line width=2pt] (supporting) -- (candidate) node[midway, above] {$r(i', k)$};
            \draw[arrow, thick,line width=2pt] (blank) -- (candidate);
        \end{tikzpicture}
        \caption{Sending availabilities.}
    \end{subfigure}

    \caption{Message-passing in APC. (a) Sending responsibilities. (b) Sending availabilities.}
    \label{fig: message_passing}
\end{figure}
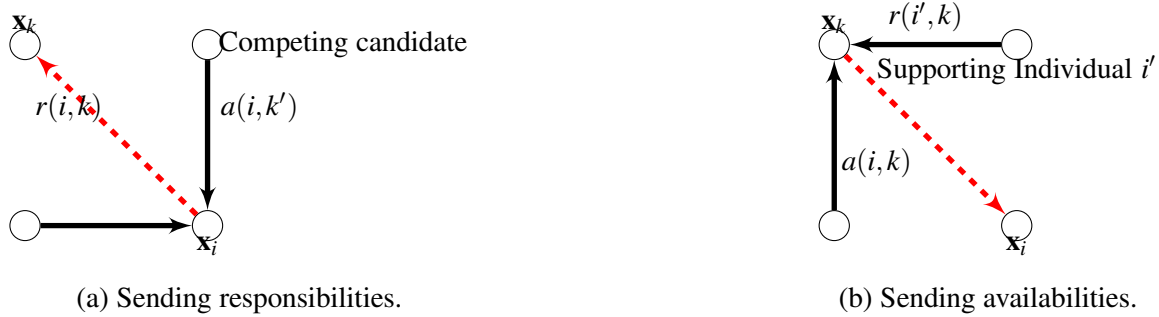

Sheng et al.~\cite{sheng2020differential} introduced a DE algorithm with adaptive niching and $K$-means operation (DE$\_$ANS$\_$AKO) tailored for hard partitional data clustering, where data sets are divided into separate, non-overlapping clusters, i.e., given a data set $\mathbf{X} =\{x_1,x_2, \cdots,x_{N_p}\}$ with $d$-dimensional features, the goal is to create clusters $\mathbf{H}=\{H_1,H_2,\cdots,H_k\}$ such that: (1) $H_j=\Phi,~1\leq j\leq k$; (2) $\cup_{j=1}^k H_j = \mathbf{X}$; and (3) $H_i\cap H_j=\Phi, ~i\neq j,~ 1 \leq i, j \leq k$. To address this challenge, they introduced an adaptive niching scheme (ANS) that dynamically adjusts niche sizes in the population. ANS adapts the niche size, $N_i$, as $N_i=cauchy(N_{i,mean},0.5)$, where $N_{i,mean}$ is computed using a weighted sum of performance measures $S_i$, as follows:  
\begin{equation} 
N_{i,mean}=(1-p)N_{i,mean}+p\frac{S_i}{S_{ave}}\frac{\sum N_i S_i}{\sum S_i}, 
\end{equation}  
here, $S_i$ gauges the success rate of offspring replacing their parents in the $i^{th}$ niche, and $S_{ave}$ represents the average value of all $S_i$. ANS prevents premature convergence during evolutionary search, aiding in discovering optimal or near-optimal solutions. Additionally, an adaptive $K$-means operation (AKO) is introduced to enhance search efficiency. Within each niche, the fitness of the selected parent individual is compared to the niche's average fitness. If the selected parent has superior fitness, a K-means operation is performed on its offspring, and the number of iterations, $G_{LS}$, is determined by a formula involving fitness values and scaling constants, as follows:  
\begin{equation} 
G_{LS}=T\frac{f-f_{ave}}{best_f-worst_f}, 
\end{equation}  
where $f$ denotes the fitness of the selected parent individual, $T$ is a scaling constant, and $f_{ave},~best_f,$ and $worst_f$ are the average, best, and worst fitness, respectively. This AKO process fine-tunes clustering within niches to improve solution quality. Duan et al.~\cite{duan2018adaptive,huang2020concurrent} introduced the adaptive niching population-based DE (ANPDE) algorithm, which dynamically adjusts population size during evolution. They built upon the niching \textit{dADE/nrand/1} algorithm~\cite{epitropakis2013dynamic}, known for controlled parameter adaptation, exploration of uncharted search areas, and maintaining stable, high-performing solutions. However, \textit{dADE/nrand/1} uses a fixed population size, limiting its adaptability. ANPDE overcomes this limitation by dividing the population into subpopulations and dynamically adjusting both the number of individuals and subpopulations as evolution progresses. Three performance-enhancing techniques are proposed:

(i) A heuristic clustering method generates non-overlapping subpopulations by calculating nearest distances between clusters, ensuring that the intracluster distance ($d_{intra}$) is greater than or equal to the intercluster distance ($d_{inter}$). Here, $d_{intra}$ represents the sum of distances within clusters, while $d_{inter}$ is the sum between clusters. (ii) The algorithm includes a population adaptation strategy, and (iii) an auxiliary movement strategy for the best individuals, promoting population diversity while reducing computational costs. Wang et al.~\cite{wang2020automatic} introduced an innovative automatic niching technique based on affinity propagation clustering (APC) and designed a novel niching DE algorithm (ANDE) for addressing MMOPs. In ANDE, APC serves as a parameter-free automatic niching method, eliminating the need to predefine the number of clusters or cluster sizes (A detail of APC is provided in Section 3 of the \textit{Supplementary file}). In APC, two message-passing processes\footnote{In the APC algorithm, two message-passing processes, responsibility and availability, are defined. These processes are iterative loops used to determine how suitable it is for an individual to act as an exemplar for another individual and how appropriate it is for an individual to select another individual as its exemplar.}, ``responsibility" and ``availability," are defined. Responsibility ($r(i, k)$) is sent from individual $\mathbf{x}_i$ to its candidate exemplar $\mathbf{x}_k$, as shown in Fig.~\ref{fig: message_passing}(a). It represents how suitable $\mathbf{x}_k$ is to be an exemplar for $\mathbf{x}_i$. The responsibility is calculated as the similarity between $\mathbf{x}_i$ and $\mathbf{x}_k$, minus the maximum of availabilities and similarities between $\mathbf{x}_i$ and other competing candidates. Availability ($a(i, k)$) is sent from individual $\mathbf{x}_k$ to $\mathbf{x}_i$, as illustrated in Fig.~\ref{fig: message_passing}(b). It indicates how appropriate it is for $\mathbf{x}_i$ to choose $\mathbf{x}_k$ as its exemplar. Availability is determined by the self-responsibility $r(k, k)$ of $\mathbf{x}_k$ and the sum of positive responsibilities that $\mathbf{x}_k$ receives from other supporting individuals. To prevent strong positive responsibilities from overly influencing the process, $a(i, k)$ is capped at zero. After the APC algorithm automatically partitions the population into appropriate clusters or niches to identify different peak regions, DE evolutionary operators are applied within each niche. Additionally, after each generation's evolution, a contour prediction approach (CPA) is used to estimate the contour landscape of each niche. The CPA predicts potential optima's rough positions within niches by leveraging distribution information from select individuals to expedite convergence. In a $2-D$ context, solutions represent positions, and fitness values ($f$) represent elevations, forming pairs like $(x_{i,1}, x_{i,2})~f$. Niches, each with a niche seed, form networks comprising individuals closest to the seed. 

To enhance solution accuracy, the two-level local search (TLLS) (Algorithm~\ref{alg: TLLS}) follows CPA. TLLS comprises two local search levels: niching-level and individual-level. Initially, niching-level local search focuses on niche seeds to discover promising solutions. Given MMOPs' objective of locating multiple optima, the initial focus is on the niching-level local search, as different niches target distinct peaks. On the niching level, the local search targets the niche seed to discover more promising solutions. Since each niche has its seed, better niche seeds are closer to global optima and are more likely to undergo local search. The probability of executing local search at the niche level depends on the fitness values of niche seeds. Consequently, after executing the niching-level local search operator, an individual-level local search also occurs. Specifically, if the current niche $i$ meets the probability $p_i$ for local search, some individuals with superior fitness values will also undergo local search.
\begin{algorithm}
\caption{TLLS~\cite{wang2020automatic}}\label{alg: TLLS}
\begin{algorithmic}[1]
% \STATE \textbf{Begin}
\STATE Generate the sample standard deviation $\sigma$ \(=10^{-1-\frac{10/D+3}{MaxFE}} \).
\STATE Calculate the niche-level local search probability $p_i=rank_i/N$ ($rank_i$ is the rank of niche seed).
\FOR{each niche $S_i$}
    \IF{$\text{rand} < p_i$}
        \FOR{each individual $x_{ik}$ in $S_i$}
            \STATE Calculate the individual-level local search probability $P_{ik}=rank_{k}/N_i$ ;
            \IF{$\text{rand} < p_{ik}$}
                \STATE Sample 2 points around individual $x_{ik}$ based on the Gaussian distribution;
                \STATE Evaluate these 2 points and denote the better one as $x'_{ik}$;
                \STATE If $x'_{ik}$ is better than $x_{ik}$, replace $x_{ik}$ with $x'_{ik}$;
            \ENDIF
        \ENDFOR
    \ENDIF
\ENDFOR
% \STATE \textbf{End}
\end{algorithmic}
\end{algorithm}
Yang et al.~\cite{yang2018cluster} employed a cluster-based niching DE algorithm, amalgamating the cluster pool, niche method, and DE approach. The cluster pool generates niche subpopulations, while the DE algorithm handles population evolution. Various mutation strategies and the plane cut crossover operator were incorporated to enhance convergence speed and structural diversity. A self-adaptive DE~\cite{bovskovic2017clustering} for MMOPs integrates self-adaptive parameter control, clustering, and crowding. Small subpopulations centered on the best individuals dynamically adjust population size and replace less fit individuals. Crowding prevents poor individuals from disrupting convergence, while population size expands with the identified optima and subpopulations.  

A surrogate-assisted DE (DSADE)~\cite{ji2023surrogate} with region decomposition (ARD) for MMOPs adaptively divides the decision space into subregions, assigning an independent population to each. Initial individuals are sorted by objective function values, and seeds for each sub-population are selected based on low objective values and a large distance criterion $\gamma$ $((\gamma=max_{j=2,\ldots, PS}(min_{i<j}dist(\mathbf{x}_i,\mathbf{x}_j))+0.1))$. A multilayer perceptron-based global surrogate model (MLPGS) evolves each sub-population, aiming to approximate optimal solutions by optimizing the surrogate function. Additionally, a self-adaptive gradient descent-based local search (SaGDLS) refines solutions by defining a new vector $\mathbf{x}_{new}=\mathbf{x}+\Delta \mathbf{\eta}$, where $\Delta \mathbf{\eta}$ is derived from the function minimum and gradient, enhancing convergence to multiple optima. A dual-strategy DE (DSDE) with affinity propagation clustering (APC) is proposed~\cite{wang2017dual} for MMOPs. A dynamic cluster sizing method\footnote{In every generation, an integer is chosen randomly from a fixed interval as the cluster size $M$, while the number of niches is $N/M$. If $N\%M = 0$, the last niche has $M + N\%M$ individuals. The interval for $M$ is set as [4, 20] because DE must have at least four individuals.}~\cite{yang2016multimodal} randomly selects a cluster size $M$ within a set interval, partitioning the population into subpopulations. Each subpopulation is further divided into superior and inferior individuals based on fitness, with \textit{DE/lbest/1} applied to superior individuals and \textit{DE/current-to-rand/1} to inferior ones. APC is used to adaptively select diverse individuals from various optimal regions, where clusters form through message-passing, exchanging responsibility $r(i, k)$ and availability $a(i, k)$. Archive stores stagnated and converged individuals, preserving promising solutions and allowing reinitialization to explore new areas. 

\subsection{Minimum Spanning Tree}
Wang et al.~\cite{wang2019distributed} introduced a novel niching technique utilizing minimum spanning trees (MST) and applied it to DE, creating MSTDE to address MMOPs. In each generation, an MST is constructed based on the distance information among individuals. Subsequently, they prune the $M$ largest weighted edges of the MST, resulting in the formation of subtrees or subpopulations where DE operators are executed. Furthermore, the authors introduced a dynamic pruning ratio strategy to determine $M$, aiming to reduce its sensitivity and enhance niching performance. This strategy effectively strikes a balance between diversity and convergence. Additionally, they harnessed the availability of virtual machines (VMs) to implement a distributed model within MSTDE. This approach allows different subpopulations to run concurrently on distributed VMs, optimizing computational efficiency.

A novel DE variant, called MHDE~\cite{bu2024multi}, is proposed for MMOPs, consisting of three main steps: population clustering, intra-cluster evolution, and extra-cluster evolution. In the \textit{population clustering} phase, a balanced locality-sensitive hashing (BLSH) method with adaptive clustering quantity is introduced. BLSH, an enhancement of the original LSH~\cite{gionis1999similarity} (details provided in Section 4 of the \textit{Supplementary file}), mitigates its imbalances by sequentially selecting individuals to ensure a balanced distribution across clusters~\cite{zhang2016toward}. First, individuals are mapped to hash values, sorted, and evenly distributed into $N_C$ clusters, each containing $cs = N_p/N_C$ individuals. Clustering is repeated periodically (after $G$ iterations) to avoid overlapping search regions and to regenerate the population (excluding the best individual). To adaptively determine $N_C$, an entropy-based method~\cite{gray2011entropy} is applied. In the \textit{intra-cluster evolution}, four mutation strategies are employed: \textit{DE/current-to-best/1, DE/rand-to-best/1, DE/current-to-rand/1,} and \textit{DE/rand/1}. Before applying these strategies, an individual improvement indicator is computed. Furthermore, an improved JADE-like adaptive parameter mechanism, based on this indicator, is used to adjust parameters $F$ and $CR$ dynamically (details provided in Section 5 of the \textit{Supplementary file}). In the \textit{extra-cluster evolution}, an extra-cluster subpopulation (EcPop) is constructed to enhance convergence and enable better information exchange between clusters. The mutation strategies \textit{DE/current-to-best/1} and \textit{DE/rand-to-best/1} are used within this subpopulation. The combination of these components ensures effective exploration, exploitation, and convergence for MMOPs.

A minimum spanning tree niching-based DE (TNDE)~\cite{li2023minimum} with a knowledge-driven update (KDU) strategy is proposed for MMOPs. The term ``knowledge" in KDU means historical evolution, fitness distribution, and individual distribution information. In TNDE, the evolutionary phase is divided into the exploration and exploitation phases, determined by whether the maximum fitness value's stagnant time ($ST_{mf}$) exceeds a threshold ($ST$). In the exploration stage ($ST_{mf} < ST$), individuals converge toward optimal regions, while in the exploitation stage ($ST_{mf} \geq ST$), they converge toward the optima within these regions, with stagnant individuals updated by KDU. TNDE also features a minimum spanning tree niching (MSTN) strategy to adaptively divide the population, dynamically adjusting niche numbers. During local stage-based mutation (LSM), the \textit{DE/rand/1} strategy adapts by replacing the first term $\mathbf{x}_{r1}$ based on the phase, i.e., for exploration $\mathbf{x}_{r1} \rightarrow \mathbf{x}_b$ or for exploitation $\mathbf{x}_{r1} \rightarrow \mathbf{x}_l$. Additionally, a directional guidance selection (DGS) inspired by crowding distance is used: when an offspring $\mathbf{u}_i$ outperforms the most similar parent $\mathbf{x}_p$, a beneficial vector is calculated, yielding an improved individual $\mathbf{u}_i' = \mathbf{u}_i + N(1, 0.1)(\mathbf{u}_i - \mathbf{x}_p)$.

A summary of selected clustering-based DE variants is presented in Table~\ref{tab: clustering DE}, detailing their results, specific applications, or the MMOPs they addressed, and proposed future research directions. This table emphasizes how clustering techniques are utilized within these DE variants to enhance solution diversity, effectively explore decision spaces, and handle complex multimodal landscapes. The insights provided help showcase their effectiveness in both theoretical benchmarks and real-world applications.
\begin{table}[]
    \centering
    \caption{Summary of clustering-based DE.}
    \label{tab: clustering DE}
    \resizebox{1\linewidth}{!}{
    \begin{tabular}{c|l|c|p{6.5cm}|p{3cm}|p{5.5cm}|p{6cm}}\hline
        S. No. & Variants & Year & Description & Applications & Results & Future Work \\
        \hline
        1 & DE\_ANS\_AKO~\cite{sheng2020differential} & 2020 & Combines adaptive niching and k-means to enhance search efficiency and prevent premature convergence. & Partitional data clustering. & Outperforms other methods on synthetic and real datasets. & Can be extended to automatic clustering and applied to time series, graph, and spatial data. \\
        \hline
        2 & ANPDE~\cite{duan2018adaptive,huang2020concurrent} & 2018 & Integrates heuristic clustering, parameter adaptation, and auxiliary movement for multimodal optimization. & Training multiple neural networks. & Outperforms traditional methods in training NNs. & Extend to include hyperparameters and model diversity for incremental learning. \\
        \hline
        3 & ANDE~\cite{wang2020automatic} & 2020 & Uses APC with DE for peak identification, incorporating CPA and TLLS strategies for convergence. & Solving MMOPs. & Outperforms other multimodal DE algorithms in solution accuracy and convergence. & Potential to address high-dimensional MMOPs and dynamic environments. \\
        \hline
        4 & DSADE~\cite{ji2023surrogate} & 2023 & Incorporates adaptive region decomposition, a multilayer perceptron surrogate, and self-adaptive gradient descent. & Expensive MMOPs. & Promising results on 20 test functions. & Future work focuses on improving surrogate models and evolutionary strategies for high-dimensional problems. \\
        \hline
        5 & DSDE~\cite{wang2017dual} & 2017 & Dual-strategy mutation, APC, and an archive mechanism for stagnation. & Multi-global optima problems. & Outperforms state-of-the-art algorithms in terms of global optima and accuracy. & Extension to complex real-world problems and use of parallel computing. \\
        \hline
        6 & MSTDE~\cite{wang2019distributed} & 2019 & Builds MST for niching, balancing diversity and convergence with distributed processing. & MMOPs requiring multi-optima search. & Outperforms existing algorithms with improved niching and diversity. & Extension to dynamic environments with minimal prior knowledge. \\
        \hline
        7 & MHDE~\cite{bu2024multi} & 2024 & Combines hash clustering with adaptive mutation for fast population clustering and improved performance. & Global optimization and WSNs. & Effective in node deployment in WSNs. & Future work on real-time optimization tasks and parallel computing. \\
        \hline
        8 & TNDE~\cite{li2023minimum} & 2023 & Uses MSTN for dynamic niching and KDU strategy to refine solutions. & CEC 2013 MMOPs. & Outperforms 16 algorithms in high-dimensional MMOPs. & Future focuses on high-dimensional and dynamic MMOPs like routing and scheduling. \\
        \hline
    \end{tabular}}
\end{table}

\section{Mutation and Parameter Adaptation-based DE}\label{sec: mutation DE}
In this section, we discuss various adaptive mutation strategies and parameter adaptation mechanisms designed to enhance the performance of DE in solving MMOPs. These techniques aim to balance exploration and exploitation by adjusting mutation operations and control parameters, fostering diversity, and guiding the search toward multiple optima. We explore several approaches, including distributed mutation frameworks, niche-level adaptability, and self-adaptive parameter tuning, which have shown promising results in optimizing complex multimodal landscapes.

\subsection{Adaptive Mutation Strategies}

An adaptive DE with archive (ADEA)~\cite{agrawal2022solving} is proposed for locating diverse and accurate optimal solutions in MMOPs. ADEA incorporates the following components: a distributed mutation framework where each acts as an exemplar, exploring its local search space using an adjustable range mechanism designed to perform two tasks: (a) identify the nearest best neighbor ($\mathbf{x}_{nbest}$) of the current individual, and (b) adjust the upper and lower bounds for virtual random individuals used in the mutation process. An adaptive mutation strategy, \textit{DE/nbest/2}, given by Eq.~\eqref{eq: virtual mutation}, where $\mathbf{vr}_{rj}$ ($j=1,\ldots,4$) are the virtual random individuals, guides promising solutions toward better positions. An elite archive is also introduced to handle stagnated individuals, preventing them from being trapped in local optima.
\begin{equation}\label{eq: virtual mutation}
    \mathbf{v}_i = \mathbf{x}_{nbest} + F \cdot (\mathbf{vr}_{r1} - \mathbf{vr}_{r2}) + F \cdot (\mathbf{vr}_{r3} - \mathbf{vr}_{r4}).
\end{equation}

A distributed individuals-based DE (DIDE)~\cite{chen2019distributed} introduced a distributed approach for MMOPs by treating each individual as an independent unit with a virtual population (DIMP framework) to maintain diversity and track multiple peaks. Each individual's virtual population, $\mathbf{vx}_i=\{\mathbf{vx}_1,\mathbf{vx}_2,\ldots,\mathbf{vx}_n\}$, is generated within an adaptive range $R$, allowing mutation operations to be guided by virtual individuals, $\mathbf{v}_i=\mathbf{x}_i+F(\mathbf{vx}_{i,1}-\mathbf{vx}_{i,2})$. To balance exploration and exploitation, $R$ starts large and decreases adaptively with an ARA strategy, which shrinks $R$ when improvement stalls for consecutive generations. Additionally, a lifetime mechanism gives each individual a limited lifespan to renew diversity upon reaching high fitness. In contrast, a Gaussian-based elite learning mechanism (ELM) refines archived elite solutions, enhancing accuracy on identified peaks. Ref.~\cite{wang2023improved} proposed a two-stage mutation-based DE (MMODE$\_$TSM$\_$MMED) for MMOPs, introducing a modified maximum extension distance (MMED) to boost diversity and convergence. The \textit{DE/rand-to-MMEDBest/2} mutation strategy and an MMED-based environmental selection method are designed to effectively identify multiple equivalent Pareto sets.

Proximity ranking-based multimodal DE (PRMDE)~\cite{zhang2023proximity} is an efficient framework for locating multiple optima in MMOPs. In PRMDE, a non-linear weight function ranks individuals by proximity, influencing selection probabilities\footnote{The probabilities are calculated as $P_j=w_j/\sum_{i=1}^{N_p-1}w_i$, where $w_j=\frac{\exp(-(rank(j)-1)^2/(2(\sigma(N_p-1))^2))}{\sqrt{2\pi}\sigma (N_p-1)}$, $rank(j)$ is the ranking of the $j$th individual sorted by proximity to the target individual, and $\sigma$ (a non-linear decreasing function of function evaluations) controls the decay rate of weight $w$ with rank.}, which encourages selecting Euclidean neighbors for mutation through roulette wheel selection. Specifically, in the classical mutation schemes of DE, the $\mathbf{x}_{r1}$ term is replaced with $\mathbf{x}_{r1}^*$ (the best one among all the randomly selected parent individuals). As evolution progresses, the selection probabilities of nearby individuals increase, enhancing local exploration. Additionally, better individuals have a chance to undergo Gaussian-based local search, provided $rand(0,1)<PL_i$, where $PL_i$ is the local search probability for the $i$th individual. In SA-DQN-DE~\cite{liao2024differential}, three mutation strategies are used in action as $action = \{ \textit{DE/rand/1}, \textit{DE/rand-to-best/1}, \textit{DE/improved strategy} \}$. A ring topology-based niching DE with self-adaptive parameters is proposed~\cite{jiang2022self} for MMOPs, where the $pbest$ term in the \textit{DE/current-to-pbest/1} strategy is replaced by a $pnbest$ defined by the ring topology. Additionally, an oppositional learning-based restart mechanism is incorporated to escape local optima.

\subsection{Self-adaptive Parameters}

Wang et al.~\cite{wang2023fitness} proposed an adaptive DE (FDLS-ADE) for MMOPs that incorporates a fitness- and distance-based local search (FDLS). In FDLS-ADE, a Gaussian distribution is used for the basic local search operation, $\mathbf{x}_i' = Gaussian(\mathbf{x}_i, \sigma)$, where $\sigma$ depends on the problem's dimensionality and the number of function evaluations (FEs). A fitness-based local search employs a probability function to focus on promising individuals to avoid unnecessary evaluations on suboptimal or local optima. A distance-based local search is also introduced, guiding exploration by defining a probability function that identifies promising regions. These local search strategies are integrated with DE using adaptive parameters. A detailed explanation of fitness-distance local search is provided in Section 6 of the \textit{Supplementary file}.

PRMDE~\cite{zhang2023proximity} enhanced adaptability through a niche-level parameter tuning scheme, where better base individuals are assigned smaller $u_{F,b}$ to generate mutation vectors closer to them for exploitation, while worse base individuals have larger $\mu_{F,b}$ to promote exploration. Similarly, better target individuals receive smaller $CR$, producing offspring close to their parents, whereas worse individuals are assigned larger $CR$, enhancing diversity. Random sampling of $F$ and $CR$ from Gaussian distributions further improves population diversity~\cite{tang2014differential}. In the mutation operation, the scaling factor $F$ is randomly generated using a Gaussian distribution with mean $\mu_{F,b}$ (Eq.~\eqref{eq:uF}) and standard deviation 0.1~\cite{tang2014differential}. For the crossover operation, the crossover probability $CR$ is sampled from a Gaussian distribution with mean $\mu_{CR, i}$ (Eq.~\eqref{eq:uCR}) and standard deviation 0.1~\cite{tang2014differential}.
\begin{align}
\mu_{F,b} &= \frac{f_{\text{max}} - f_b + \eta}{f_{\text{max}} - f_{\text{min}} + \eta},~F_b = \mathcal{N}(\mu_{F,b}, 0.1)  \label{eq:uF} \\
\mu_{CR,i} &= \frac{f_{\text{max}} - f_i + \eta}{f_{\text{max}} - f_{\text{min}} + \eta},~
CR_i = \mathcal{N}(\mu_{CR,i}, 0.1),  \label{eq:uCR} 
\end{align}
here, $b$ represents the index of the base individual in the mutation operator, while $i$ is the index of the target individual (parent $\mathbf{x}_i$). $f_{\text{max}}$ and $f_{\text{min}}$ are the maximum and minimum fitness in the current population, respectively, and $\eta = 1.0 \times 10^{-40}$ prevents division by zero.

Building upon the need for niche-level adaptability, Liang et al.~\cite{liang2024niche} proposed NAAPS, leveraging an IMM (detailed in Algorithm 6 of the \textit{Supplementary file}). Unlike population-wide parameter updates as in JADE~\cite{zhang2009jade}, NAAPS records historical niche behaviors, dynamically matching niche centers to optimize scaling factors and crossover probabilities. This niche-specific adaptation enhances the ability to handle diverse peaks in MMOPs. The algorithm incorporates Cauchy and Gaussian distributions to tune $F_j$ and $CR_j$, respectively, with the mutation strategy \textit{DE/current-to-best/1} guiding updates (Algorithm~\ref{alg: NAAPS}). \begin{algorithm}%[H]
\caption{NAAPS Algorithm~\cite{zhang2023proximity}}
\label{alg: NAAPS}
\begin{algorithmic}[1]
\STATE \textbf{Input:} Historical records, $\texttt{nicheCenter}$, and parameters
\STATE \textbf{Output:} Updated population
\STATE Execute IMM to retrieve niche-specific historical data
\FOR{each niche}
    \STATE Dynamically assign scaling factors and crossover probabilities
    \STATE Perform mutation, evaluate individuals, and update historical records
\ENDFOR
\end{algorithmic}
\end{algorithm}
Dominico et al.~\cite{dominico2020self} extended DE for MMOPs through two variations of the niching DE (NCDE) algorithm. The first, NCjDE, integrates the self-adaptive parameter mechanism from jDE with neighborhood mutation and crowding strategies, ensuring better convergence. The second, NCjDE-HJ enhances local search accuracy by incorporating the Hooke-Jeeves direct search\footnote{The Hooke-Jeeves (HJ) direct search algorithm minimizes a function in multiple dimensions through exploratory and heuristic pattern moves~\cite{hooke1961direct}. In the exploratory phase, each dimension $j$ of the current point $\mathbf{x}_i$ is perturbed by $\pm \Delta_i$. If both $x_{i,j} + \Delta_i$ and $x_{i,j} - \Delta_i$ fail, $\Delta_i$ is halved, and new directions are explored until the stopping criterion is met. After exploration, a heuristic pattern move combines the previous best point and the current point to generate a new point: $\mathbf{x}_{i+1} = \mathbf{x}_i + (\mathbf{x}_i - \mathbf{x}_{i-1})$. The process repeats until the function evaluation limit is reached.
}, complementing the optimization process. These methodologies underscore the versatility of DE in tackling MMOPs, with adaptive, niche-specific, and hybrid strategies pushing the boundaries of optimization. Each approach effectively addresses challenges of local optima, stagnation, and parameter sensitivity, contributing valuable insights to the field of evolutionary computation.

\section{Hybrid Approaches in DE}\label{sec: hybrid DE}
In this section, a detailed overview of the hybridization of DE algorithms with other evolutionary approaches, including memetic algorithms, is presented. The discussion highlights how such hybrids leverage the strengths of multiple techniques to enhance performance and address complex optimization challenges effectively. Specific focus is given to the role of hybridization in improving diversity, convergence speed, and solution quality in MMOPs.
 
\subsection{Hybrid with Memetic Algorithms}
The term memetic algorithms, introduced by Moscato~\cite{moscato1989evolution}, describes the combination of population-based search methods with local refinement techniques. MAs have gained prominence in evolutionary optimization~\cite{ong2010memetic} but are occasionally conflated with other meme-inspired paradigms~\cite{ong2007special,chen2011multi}. They enhance population-based methods, such as EAs, by integrating problem-specific refinement techniques. Local refinement, often called \textit{local search}, \textit{lifetime learning}, or \textit{individual learning}, leverages domain knowledge to iteratively improve solutions~\cite{moscato1989evolution}. DE has been significantly enhanced for MMOPs through innovative memetic, niching, and adaptive strategies, showcasing how adaptive techniques and local search mechanisms improve its efficiency and adaptability. Sheng et al.~\cite{sheng2022differential} proposed the adaptive neighborhood mutation-based memetic DE (ANLDE) algorithm, which integrates adaptive neighborhood mutation (ANM) and local improvement strategies. ANM facilitates diverse exploration in the early stages of evolution and transitions to intensive exploitation later by dynamically selecting neighborhood sizes ($M_i$) based on iteration data and fitness values. Within each neighborhood, the algorithm alternates between \textit{DE/rand/1} and \textit{DE/best/1} mutation strategies. Promising individuals are distinguished using a promising set ($\mathbf{P}$) derived from fitness and Euclidean distance metrics, i.e., $\mathbf{P}_S=\{i|i\in subA\cup BS\}$, where $subA$ consists of individuals with fitness differences ($fd$) less than the average $fd$, and $BS$ represents the best solution in the population, and unpromising set ($UPS=\{i|i\in subB\}$) includes individuals from $subB$, where $subB$ encompasses individuals with Euclidean distances ($ed$) less than the average $ed$.

ANLDE further refines solutions with Gaussian-based local search, which focuses on offspring with superior replacement potential. Neri and Todd~\cite{neri2022study} explored six memetic strategies to enhance the performance of five DE frameworks: SDE, CDE, $dADE/nrand/1$, CCDE, and CCjDE, integrated with the Broyden-Fletcher-Goldfarb-Shanno (BFGS) algorithm (Detailed is given in Section 7 of the \textit{Supplementary file}). The objective was to enhance the amalgamation of local search and niching techniques within these frameworks. The first strategy adds a BFGS search phase before the main DE phase, allocated up to 5$\%$ of MaxFE per individual. In the second approach, BFGS is applied after the primary DE phase. The DE algorithm consumes 95$\%$ of MaxFE, with BFGS used for the remaining evaluations. The third strategy combines an archive with BFGS, employing the BFGS search on archived individuals instead of the entire population. The fourth strategy, similar to the previous one, skips individuals considered optimal, presuming the best individual has reached an optimum. This avoids individuals within the acceptance range of the best. In the fifth strategy, an archive is combined with BFGS, and a Simulated Annealing acceptance mechanism is introduced in the DE selection mechanism. 

This operates alongside the BFGS search at the end. The final strategy resembles the previous one but includes a selective BFGS search on the archive, focusing on specific individuals. These strategies enhance the synergy of local search and niching techniques within the DE frameworks. Sheng et al. designed NSAMA, a Niching Competition-based Memetic DE with Supporting Archive and Adaptive Local Search Operation, aimed at addressing MMOPs~\cite{sheng2021adaptive}. NSAMA's workflow begins with the initialization of $(\mathbf{P})$, DE parameters, and an archive $(\mathbf{A})$. These components are then merged to form a joint population $(\mathbf{PA})$. Each generation sees the division of the joint population into niches using the speciation cluster niching (SCN)~\footnote{SCN~\cite{qu2012differential} is a technique for creating niches in the search space by iteratively selecting the best-performing individual as a seed, assembling a niche around it with $M-1$ of its closest neighbors, and removing these $M$ individuals from the population in each iteration.} method. Niches evolve through a niching competition strategy that highlights the exploitation of high-potential niches and exploration of low-potential ones. Niche potential ($PT_i$) is computed as $PT_i=f_{i,ave}\cdot (f_{i,best}-f_{i,ave})$, where $f_{i,best}$ and $f_{i,ave}$ denote the best fitness and average fitness within the niche. The normalized potential values ($pr_i$) decide whether recombination should occur within niches or between them, balancing subspace exploitation and exploration. During evolution, NSAMA employs a supporting archive strategy where individuals from archive $A$ within each niche take part in the mutation process, serving as support and replacements. This approach implicitly manages both the reading and writing of the archive. If an archive individual is replaced, the newcomer becomes part of the archive, ensuring that potential optima identified by the niches are maintained. Furthermore, an adaptive Cauchy-based local search strategy is applied, sampling a trial solution ($\mathbf{v}$) around the best solution ($\mathbf{x}_{best}$) using a Gaussian distribution with a small standard deviation. If $\mathbf{v}$ demonstrates superior fitness to $\mathbf{x}_{best}$, $\mathbf{x}_{best}$ is replaced with $\mathbf{v}$. Additionally, DE parameters are updated according to the SHADE-style parameter adaptation rule.

Wang et al.~\cite{wang2022memetic} introduced the MNCMA algorithm, an advanced memetic DE approach that integrates multi-niche sampling and neighborhood crossover strategies. In this method, the population is initially divided into multiple niches using SCN. At each generation, a subpopulation is dynamically sampled from these niches based on a probabilistic selection mechanism dependent on evolving probabilities ($p_j$), where the initial probability ($p_{j, ini}$) is determined similarly to AED-DDE, and the final probability ($p_{j, fin}$) complements it. Individual $j$ is included in the subpopulation if $p_j$ exceeds a random threshold. MNCMA combines this multi-niche sampling strategy with a neighborhood crossover approach. During the crossover operation, instead of selecting the target individual, one of its neighborhood individuals participates in the crossover process. The choice of neighborhood individuals is based on the radius ($r_i$) of the target individual, considering factors like fitness and Euclidean distance. A neighborhood individual within the calculated radius is randomly chosen to participate in the crossover operation, leading to the generation of a trial vector. Moreover, MNCMA incorporates an adaptive local search (ALS) mechanism designed to fine-tune promising solutions within the sampled subpopulation. Additionally, an adaptive elimination operation (AEO) is employed to dynamically remove less promising individuals from the population throughout the evolutionary process.

These methodologies collectively underscore the advancements in DE for MMOPs, showcasing innovative combinations of memetic frameworks, niching strategies, and adaptive mechanisms. By integrating local search, neighborhood-based operations, and adaptive sampling, these approaches effectively navigate the challenges of multimodal landscapes, achieving a balance between exploration and exploitation.

\subsection{Hybrid with PSO}
DE is frequently hybridized with PSO, a simple optimizer inspired by the social dynamics of fish schools or bird flocks. This trend may stem from their shared use of different operations to perturb solutions~\cite{das2016recent}. The QPSODE algorithm, introduced by Fahad et al.~\cite{fahad2023optimizing}, hybridizes quantum PSO (QPSO) with DE to tackle the challenges of slow convergence and suboptimal exploration often encountered in MMOPs. This hybrid approach leverages the strengths of QPSO's quantum-inspired exploration capabilities and DE's robust exploitation mechanisms. Key innovations in QPSODE include: (i) A dynamically adjusted control parameter enhances the algorithm's ability to transition seamlessly from exploration in early iterations to exploitation in later stages. (ii) The population is strategically divided into subgroups, facilitating focused exploration of diverse regions in the search space. (iii) Intelligent and Gaussian mutation strategies are employed alongside traditional crossover, ensuring a balance between diversity and convergence speed. (iv) This Boltzmann-based probabilistic selection mechanism prevents premature convergence by maintaining a healthy diversity in the population. By combining these techniques, QPSODE achieves a synergistic balance between the exploration capabilities of QPSO and the local search efficiency of DE. This hybrid approach has demonstrated superior performance on multimodal optimization benchmarks, showcasing its potential to address complex real-world problems requiring robust global and local search dynamics.

A combined summary of mutation-based and hybrid DE variants is presented in Table~\ref{tab: mutation hybrid DE}, highlighting their key innovations, results, and applications in solving MMOPs. Future research directions are included to provide insights into potential improvements and expansions.
\begin{table}[]
    \centering
    \caption{Summary of mutation-based and hybrid DE.}
    \label{tab: mutation hybrid DE}
    \resizebox{1\linewidth}{!}{
    \begin{tabular}{c|l|c|p{6.5cm}|p{4cm}|p{4.5cm}|p{5.5cm}}\hline
    S. No. & Variants & Year & Description & Applications & Results & Future Work \\\hline
    1 & ADEA~\cite{agrawal2022solving} & 2022 & Combines adaptive mutation, elite archive, and adjustable local search range. & MMOPs, CEC 2013 benchmarks. & Outperforms existing algorithms in accuracy and diversity. & Extend to high-dimensional and real-world problems. \\\hline
    2 & DIDE~\cite{chen2019distributed} & 2019 & Uses DIMP framework, lifetime mechanism, and ELM for enhanced diversity and accuracy. & MMOPs, Nash equilibria in markets. & Achieves competitive or superior performance in benchmark tests. & Apply to real-world MMOPs like machine design optimization. \\\hline
    3 & PRMDE~\cite{zhang2023proximity} & 2023 & Focuses on exploration-exploitation balance with proximity-based selection, adaptive parameters, and local search. & MMOPs, CEC 2013 benchmarks. & Highly competitive optimization performance. & Enhance with new mutation strategies and real-world engineering applications. \\\hline
    4 & FDLS-ADE~\cite{wang2023fitness} & 2023 & Uses fitness and distance information to improve local search and prevent redundancy. & MMOPs, CEC 2015 benchmark, real-world nonlinear systems. & Outperforms CEC 2015 winner and other algorithms. & Investigate local search strategies and apply them to real-world tasks like neural network training. \\\hline
    5 & ANLDE~\cite{sheng2022differential} & 2022 & Introduces ANM for early exploration and adaptive Gaussian-based local search for refinement. & MMOPs, twenty benchmark functions. & Superior to sixteen related algorithms. & Apply to image segmentation, nonlinear control, and parameter estimation. \\\hline
    6 & MNCMA~\cite{wang2022memetic} & 2022 & Uses dynamic niche sampling and neighborhood crossover to balance exploration and exploitation. & CEC 2015 benchmarks. & Shows superior performance with proposed strategies. & Extend to multi-objective and dynamic problems, and apply to real-world optimization tasks. \\\hline
    7 & QPSODE~\cite{fahad2023optimizing} & 2023 & Integrates QPSO, swarm partitioning, adaptive control, and Gaussian mutation. & Constrained, non-convex, SMES electromagnetic problems. & Outperforms other swarm optimizers in accuracy and speed. & Improve scalability and efficiency, apply to real-world engineering problems. \\\hline
    \end{tabular}}
\end{table}

\section{Machine Learning-based DE}\label{sec: machine DE}
With decades of research in evolutionary optimization at a pivotal point, advancements in AI, particularly in deep learning and reinforcement learning, present new opportunities. This section explores the integration of machine learning into DE for solving MMOPs, focusing on the use of reinforcement learning, deep learning, and surrogate models.

\subsection{Reinforcement Learning}
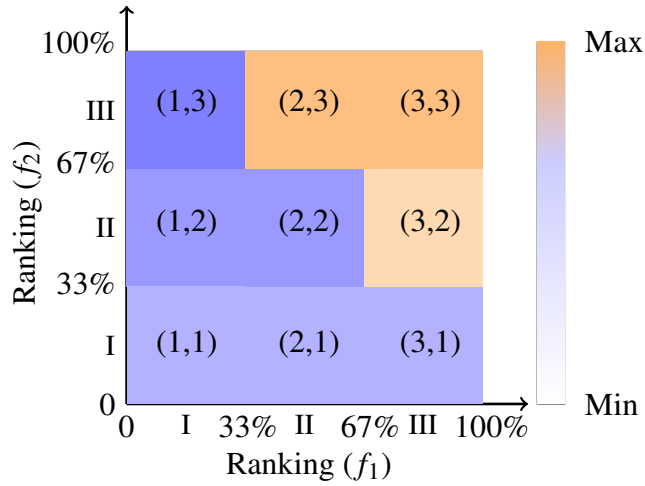
\begin{figure}
    \centering
\begin{tikzpicture}[scale=1.2]

% Grid and regions
% \draw[thick] (0,0) rectangle (4,4); % Outer box
% \draw[thick] (0,1.33) -- (4,1.33); % Horizontal lines
% \draw[thick] (0,2.67) -- (4,2.67);
% \draw[thick] (1.33,0) -- (1.33,4); % Vertical lines
% \draw[thick] (2.67,0) -- (2.67,4);
\draw[step=1.3, gray, thin] (0,0) grid (3.9,3.9);

% Labels for axes
\draw[->,line width=1pt] (0,0) -- (4.4,0)node[below] at (2,-0.4) {Ranking ($f_1$)};
\draw[->,line width=1pt] (0,0) -- (0,4.4) node[rotate=90, above] at (-0.8,2) {Ranking ($f_2$)};

% Percent labels on axes
\node[below] at (0,0) {0};
\node[below] at (1.33,0) {33\%};
\node[below] at (2.67,0) {67\%};
\node[below] at (4,0) {100\%};

\node[left] at (0,0) {0};
\node[left] at (0,1.33) {33\%};
\node[left] at (0,2.67) {67\%};
\node[left] at (0,4) {100\%};

\node[below] at (0.65,0) {\small I};
\node[below] at (1.95,0) {\small II};
\node[below] at (3.25,0) {\small III};

\node[left] at (0,0.65) {\small I};
\node[left] at (0,1.95) {\small II};
\node[left] at (0,3.25) {\small III};

% Shaded regions
\fill[blue!50] (0,2.6) rectangle (1.3,3.9); % Light blue region
\fill[blue!40] (0,1.3) rectangle (1.3,2.6); % Light blue region
\fill[blue!30] (0,0) rectangle (3.9,1.3); % Light blue region
\fill[blue!40] (2.6,1.3) rectangle (1.3,2.6); % Light blue region
% \fill[blue!20] (2.6,2.6) rectangle (1.3,3.9); % Light blue region

% \fill[blue!20] (0,0) rectangle (2,2); % Light blue region
\fill[orange!60] (2.6,2.6) rectangle (3.9,3.9); % Orange region
\fill[orange!50] (1.3,2.6) rectangle (3.9,3.9); % Orange region
\fill[orange!30] (2.6,1.3) rectangle (3.9,2.6); % Orange region
% Hierarchical states
\node at (0.66,0.66) {(1,1)};
\node at (0.66,2) {(1,2)};
\node at (0.66,3.33) {(1,3)};

\node at (2,0.66) {(2,1)};
\node at (2,2) {(2,2)};
\node at (2,3.33) {(2,3)};

\node at (3.33,0.66) {(3,1)};
\node at (3.33,2) {(3,2)};
\node at (3.33,3.33) {(3,3)};
% \node[anchor=center, fill=white, text=black, font=\bfseries] at (3.33,3.33) {(3,3)};

% Legend (color bar)
\draw[very thin] (4.5,0) rectangle (4.8,4); % Color bar outline
\shade[top color=blue!20, bottom color=white] (4.5,0) rectangle (4.8,2.67);
\shade[top color=orange!60, bottom color=blue!20] (4.5,2.67) rectangle (4.8,4);
\node[right] at (4.9,0) {Min};
\node[right] at (4.9,4) {Max};

% Title

\end{tikzpicture}
    \caption{The illustration of hierarchical states in the fitness-ranking space~\cite{li2019differential}.}
    \label{fig: fitness-ranking}
\end{figure}

Reinforcement learning (RL) has emerged as a powerful tool for enhancing DE algorithms by introducing adaptive mechanisms for strategy selection and population management. A basic procedure of the RL mechanism is illustrated in Fig.~\ref{fig: RL process} and detail is given in Section 8 of the \textit{Supplementary file}. Below, we discuss the following innovative RL-based DE frameworks designed for MMOPs.
\begin{figure}
		\begin{center}
			\begin{tikzpicture}[auto, node distance=2cm, >=latex']%[node distance=1.5cm,auto,,rotate=90,transform shape]
			\tikzset{->-/.style={decoration={
						markings,
						mark=at position #1 with {\arrow{>}}},postaction={decorate}}}
			%							\draw [help lines] (0,0) grid (9,6);
			\node[startstop,fill=lightgray](agent)at(3.5,5){\bf Agent};
			\node[startstop,fill=lightgray](environment)at(5.5,2){\bf Enviornment};
			\node[startstop,fill=lightgray](policy)at(1.5,2){\bf Policy};
			\draw [arrow,line width=2pt] (agent) -- (8,5)--(8,2)--(environment);
			\draw [line width=2pt] (4,2.2) -- (3,2.2);
			\draw [line width=2pt] (4,1.8) -- (3,1.8);
			\draw [arrow,line width=2pt] (0,2.2)--(-.5,2.2)--(-0.5,4.8)--(2,4.8);
			\draw [arrow,line width=2pt] (0,1.8) -- (-1,1.8)--(-1,5.2)--(2,5.2);
			\node[rotate=90]at(-1.3,3.5){state transfer $S_t$};
			\node[rotate=90]at(-0.2,3.5){reward $R_f$};
			\node[rotate=-90]at(8.3,3.5){action $A_t$};
			\end{tikzpicture}
		\end{center}
		\caption{\it An illustration of RL process}\label{fig: RL process}
	\end{figure}
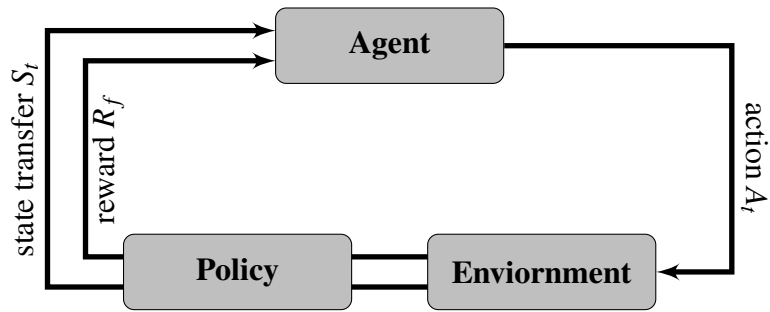
 Li et al.~\cite{li2019differential} introduced a novel DE algorithm called DE-RLFR, based on RL with Fitness Ranking. DE-RLFR operates within the Q-learning framework, treating each individual in the population as an agent. The fitness ranking values of each agent are used to encode hierarchical state variables, and three typical DE mutation operations are offered as optional actions for the agent. Through analyzing the population's distribution characteristics across objective space, decision space, and fitness-ranking space (illustrated in Fig.~\ref{fig: fitness-ranking}), a reward function for state-action pairs is designed. This function guides the population towards the PF asymptotically. Informed by its reinforcement learning experience represented by the corresponding $Q-$table values, each agent can adaptively select a mutation strategy to generate offspring individuals.

Aggarwal et al.~\cite{agrawal2021improved} introduced the multi-armed bandit-based DE (MABDE) algorithm, a novel approach for MMOPs. In MABDE, they employed the NBC method to sort individuals based on their fitness values. They applied the \textit{softmax} function with a parameter $temp$ to calculate individual probabilities, which determined the sorting order. Unlike traditional sorting based on descending edge lengths, MABDE sampled edges using cutting probabilities, requiring careful handling of the $temp$ parameter. When $temp$ is near $\infty$, MABDE employs a random strategy, cutting all edges with equal probability, promoting exploration. Conversely, when $temp$ is small, the algorithm behaves similarly to its original version, emphasizing exploitation based on fitness values and distance between superior individuals. This strategy ensures mutant vectors generated during mutation remain within the function domain. Stable mutation is applied iteratively until a valid mutant vector is obtained. MABDE also introduced a novel strategy for generating new individuals inspired by multi-armed bandits (MAB). This modification ensures that the generated individuals stay within function bounds and increases the chances of locating optimal solutions in promising areas. Additionally, an archive technique is used to store stagnated individuals, preventing wastage of limited $FE$ and enabling reinitialization for enhanced algorithm exploitation capabilities. A DE incorporating strategy adaptation and deep reinforcement learning, termed SA-DQN-DE, is proposed~\cite{liao2024differential} to enhance mutation strategy selection and optimize the search process. SA-DQN-DE employs three mutation strategies: \textit{DE/rand/1, DE/rand-to-best/1}, and an \textit{improved DE strategy}~\cite{liao2023history}. A strategy adaptation mechanism dynamically adjusts the selection probabilities of these strategies based on their rates (Table~\ref{tab: success failure}) during evolution. If the candidate pool contains $M$ strategies, the probability of selecting the $m$th strategy, $p_m$, is calculated as:

\begin{equation}
    p_m = \frac{S_{m,~G}}{\sum_{m=1}^{M} S_{m,~G}}, \quad S_{m,~G} = \frac{\sum_{g=G-L_p}^{G-1} Succ_{m,~g}}{\sum_{g=G-L_p}^{G-1} Succ_{m,~g} + \sum_{g=G-L_p}^{G-1} Fail_{m,~g}} + \epsilon,
\end{equation}where $\epsilon$ is a small constant to avoid zero success rates, and $L_p$ is the learning rate. A historical individual preservation method~\cite{liao2023history} is introduced to archive discarded individuals during evolution, based on an archive rate defined as:

\begin{equation}
    Arc_{rate}(i) = \left(1 - \frac{rank(i)}{N}\right) \times \left(1 + \exp(-(\beta - T(i)))\right)^{-1},
\end{equation}where $rank(i)$ is the fitness ranking, and $T(i)$ represents the number of stagnations without fitness updates. High-ranking individuals with low $T(i)$ are prioritized for archiving, while those with low fitness have minimal chances of being archived.

The algorithm integrates Deep Q-Learning (DQL) to refine promising solutions through local search. In DQL, the state is defined as $state = \{\mathbf{x}_1, \mathbf{x}_2, \ldots, \mathbf{x}_{num}\}$ (where $num$ is the size of the experience pool), the action space includes the mutation strategies $action = \{ \textit{DE/rand/1}, \textit{DE/rand-to-best/1}, \textit{DE/improved strategy} \}$, and rewards are assigned if the offspring outperform their parents.

\begin{table}[ht]
\centering
\caption{Success and failure memory~\cite{liao2024differential}.}\label{tab: success failure}
\renewcommand{\arraystretch}{1.5}
\resizebox{0.8\linewidth}{!}{\begin{tabular}{|c|c|c|c|c|c|c|}
\hline
\multicolumn{4}{|c|}{\textbf{Success memory}} & \multicolumn{3}{c|}{\textbf{Failure memory}} \\ \hline
\textbf{Index} & \textbf{Strategy-1}        & \textbf{Strategy-2}        & \textbf{Strategy-3}        & \textbf{Strategy-1}         & \textbf{Strategy-2}         & \textbf{Strategy-3}         \\ \hline
1              & $Succ_{1,G-L_p}$     & $Succ_{2,G-L_p}$     & $Succ_{3,G-L_p}$     & $Fail_{1,G-L_p}$      & $Fail_{2,G-L_p}$      & $Fail_{3,G-L_p}$      \\ \hline
2              & $Succ_{1,G-L_p+1}$   & $Succ_{2,G-L_p+1}$   & $Succ_{3,G-L_p+1}$   & $Fail_{1,G-L_p+1}$    & $Fail_{2,G-L_p+1}$    & $Fail_{3,G-L_p+1}$    \\ \hline
\vdots         & \vdots                     & \vdots                     & \vdots                     & \vdots                      & \vdots                      & \vdots                      \\ \hline
$L_p$           & $Succ_{1,G-1}$      & $Succ_{2,G-1}$      & $Succ_{3,G-1}$      & $Fail_{1,G-1}$       & $Fail_{2,G-1}$       & $Fail_{3,G-1}$       \\ \hline
\end{tabular}}
\end{table}

\subsection{Surrogate Models}
 A decomposition-based DE (D/REM) \cite{gao2021solving}, incorporating a radial basis function (RBF) model, is proposed for efficiently solving expensive MMOPs. The algorithm consists of two primary phases: promising subregions detection (PSD) and the local search phase (LSP). In PSD, inspired by previous studies \cite{gao2013cluster, yang2016multimodal, li2014history}, a population update strategy is combined with mean-shift clustering~\cite{cheng1995mean} to identify promising search regions. A local surrogate model $\mathbf{f}_i(\mathbf{x})$ is constructed using the $K = \min(\max((D+1)\cdot(D+2)/2, 50), |O|)$ nearest individuals from the archive set $O$, and the trial vector maximizing $\mathbf{f}_i(\mathbf{x})$ is added to $O$. The population is updated by retaining the better solution between $\mathbf{x}_i$ and its best corresponding vector. Mean-shift clustering then groups individuals into clusters by shifting them toward areas of higher density, detecting promising subregions. In LSP, a local RBF surrogate model (detailed in Section 9 of the \textit{Supplementary file}) is constructed for each identified subregion, effectively tracking global optima.

A summary of ML-based DE variants is presented in Table~\ref{tab:ML_DE}, showcasing their innovative integration of machine learning techniques, such as reinforcement learning, surrogate models, and adaptive strategies, to address MMOPs. The table highlights their key contributions, notable results, and diverse applications, demonstrating the effectiveness of ML-based approaches in enhancing convergence, diversity, and solution quality. Future research directions are also outlined to provide insights into potential advancements and broader applicability in real-world scenarios.
\section{Multi-level DE}\label{sec: multi-level DE}
Wang et al.~\cite{wang2019multilevel} introduced a multilevel sampling strategy to dynamically partition the population into fitness-based levels, enabling niching-based evolution. Subsequently, a subpopulation is adaptively sampled from individuals at various levels to undergo niching-based evolution, enabling the identification of multiple optima in the search space. During each generation of evolution, individuals in the population are initially sorted in descending order based on their fitness. The sorted population is then equally divided into $m$ levels, denoted as $L_j~(i = 0, 1\cdots, m-1)$, where individuals with higher fitness are assigned to higher levels with smaller indices. A crossover-based local search scheme refines niche seed solutions, while parameter adaptation leverages the SHADE-style mechanism. These techniques balance global exploration and local exploitation effectively. Hong et al.~\cite{hong2020multi} presented the multi-angle hierarchical DE (MaHDE) algorithm designed to address MMOPs by considering both solution quality and evolution stage. Within MaHDE, they introduced a fitness hierarchical mutation (FHM) strategy, categorizing niche individuals into low and high levels based on their fitness in comparison to the niche's mean fitness value ($f_{ave}$). If $f(\mathbf{x}_i) \leq f_{ave}$, $\mathbf{x}_i\in S$ is classified as a low-level individual; otherwise, it is designated as a high-level individual. In this algorithm, each level of individuals is updated using distinct guiding strategies. Low-level individuals draw knowledge from the best niche individual ($\mathbf{x}_{nbest}$), aiding them in swiftly approaching nearby global optima. Meanwhile, high-level individuals, also promising within the niche, rely on the self-guided mutation to sustain their superior status. Furthermore, the authors introduced two distinct strategies: a directed global search (DGS) for low-level individuals and an elite local search (ELS) for high-level individuals. 

\begin{figure}
	\centering
	\tikzset{
		block/.style={draw, fill=white, rectangle, 
			minimum height=3em, minimum width=6em},
		every path/.style={very thick},
	}
\resizebox{!}{!}{	\begin{tikzpicture}[auto, node distance=2cm, >=latex']%[node distance=1.5cm,auto,,rotate=90,transform shape]
	\tikzset{->-/.style={decoration={
				markings,
				mark=at position #1 with {\arrow{>}}},postaction={decorate}}}
%			\draw [help lines] (0,0) grid (14,14);
	\node[arrow,fill=green!50](niche)at(7,13){\bf Niche $\mathbf{x}_i$};
	\node[arrow,fill=lightgray!50](fit1)at (7,12){\bf Calculate fitness $f_1$};
	\node[arrow,fill=gray](fhm)at (7,10.8){\bf Apply FHM};
	\node[arrow,fill=lightgray](low1)at(3.5,10){\bf Low-level};
	\draw[arrow](fhm)--(3.5,10.8)--(low1);
	\node at (4.5,11.1){$f_1(\mathbf{x}_i)\leq f_{1_{ave}}$};
	\node[arrow,align=center,fill=lightgray](others1)at(3.5,8.8){$\mathbf{v}_i=\mathbf{x}_{nbest}+F\cdot$\\$(\mathbf{x}_{n\_r_1}-\mathbf{x}_{n\_r_2})$};
	\node[arrow,fill=lightgray](high1)at(10.5,10){\bf High-level};
	\draw[arrow](fhm)--(10.5,10.8)--(high1);
	\node at (9.5,11.1){$f_1(\mathbf{x}_i)> f_{1_{ave}}$};
	\node[arrow,align=center,fill=lightgray](others2)at(10.5,8.8){$\mathbf{v}_i=\mathbf{x}_{i}+F\cdot$\\$(\mathbf{x}_{g\_r_1}-\mathbf{x}_{g\_r_2})$};
	\draw[dashed] (1.8,8)--(1.8,11.35)--(12.1,11.35)--(12.1,8)--(1.8,8);
	\node[arrow,fill=lightgray!50](fit2)at(7,7){\bf Calculate fitness $f_2$};
	\node[arrow,fill=lightgray](low2)at(4.5,6){\bf Low-level};
	\draw[arrow](fit2)--(4.5,7)--(low2);
	\node[arrow,fill=gray](dgs)at(4.5,5){\bf DGS};
	\node[arrow,fill=lightgray](high2)at(9.5,6){\bf High-level};
	\draw[arrow](fit2)--(9.5,7)--(high2);
	\node[arrow,fill=gray](els)at(9.5,5){\bf ELS};
	\draw (dgs)--(4.5,4.3)--(9.5,4.3)--(els);
	\node[round,fill=red!50](terminate)at(7,3.3){Terminate?};
	\draw[dashed] (3.1,4.1)--(3.1,7.7)--(10.8,7.7)--(10.8,4.1)--(3.1,4.1);
	\draw[line width=1pt,arrow] (niche)--(fit1);
	\draw[line width=1pt,arrow] (fit1)--(7,11.35);
	\draw[line width=1pt,arrow] (7,8)--(7,7.7);
	\draw[line width=1pt,arrow] (low1)--(others1);
	\draw[line width=1pt,arrow] (high1)--(others2);
	\draw[line width=1pt,arrow] (low2)--(dgs);
	\draw[line width=1pt,arrow] (high2)--(els);
	\draw[line width=1pt,arrow] (7,4.3)--(terminate);
	\draw[line width=1.5pt,arrow] (terminate)--(1.5,3.3)--(1.5,9.5)--(1.8,9.5);
	\node at (3.5,3.5){\bf No};
	\node[rotate=-90,arrow,fill=blue!50] at (12.5,9.5){\bf Stage 1};
	\node[rotate=-90,arrow,fill=blue!50] at (11.2,6){\bf Stage 2};
	\end{tikzpicture}}
	\caption{\it The framework of MaHDE to solve MMOPs, where $\mathbf{x}_{n\_r_1}\neq\mathbf{x}_{n\_r_2}\in S$ (randomly selected in $S$), and $\mathbf{x}_{g\_r_1}\neq\mathbf{x}_{g\_r_2}\neq i \notin S$ (randomly selected in population excluding $S$).}\label{fig: framework of MaHDE}
\end{figure}
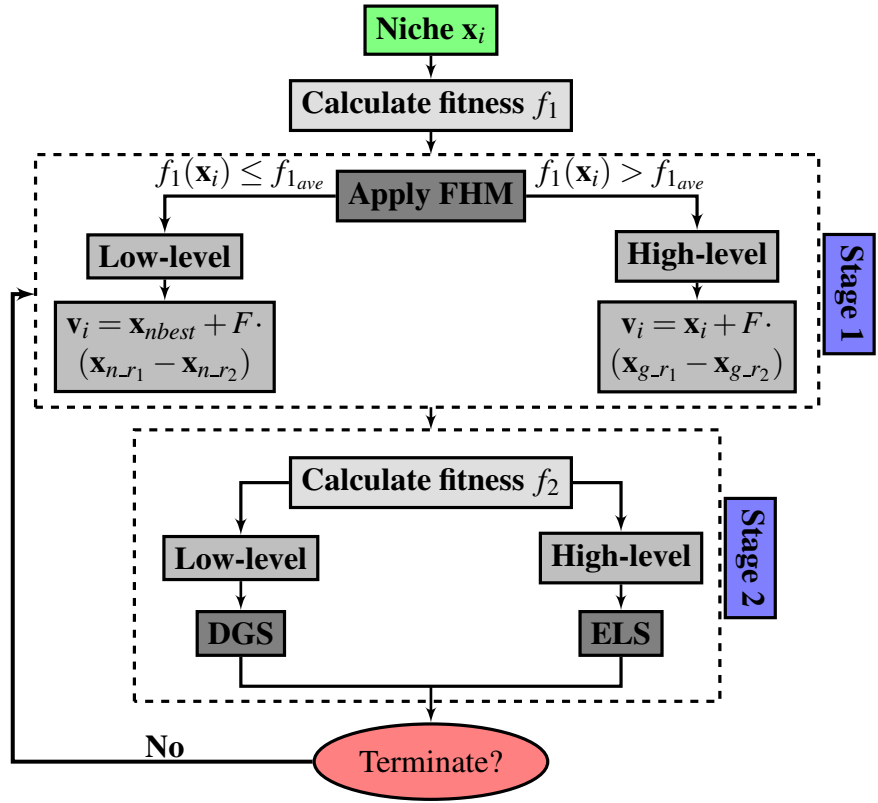

These strategies are applied after recalculating the fitness values of niche individuals and classifying them into low and high levels using the previously described procedure. Implementing the DGS strategy for low-level individuals during the later stages of evolution enhances population diversity and allows these individuals to revisit global peaks. Conversely, the ELS strategy refines the precision of high-level individuals in the latter phases of evolution. A schematic procedure of MaHDE is illustrated in Fig.~\ref{fig: framework of MaHDE}. 	
The first layer in ref.~\cite{liu2021double} divides the population into subpopulations through species-based clustering, where each subpopulation focuses on exploring local peaks. The second layer combines species seeds from all subpopulations into a new group that performs a global DE-based search, aiming to detect peaks overlooked in the first layer. A layered approach inspired by wireless sensor networks is proposed in ref.~\cite{huang2024wireless}. In the first layer, each individual \(\mathbf{x}_i\) forms relationships with others based on overlapping monitoring areas. These relationships are categorized into direct intersections (first level, \(L_1\)), indirect intersections (subsequent levels, \(L_2, L_3, \dots\)), and unrelated individuals (\(L_{\text{no}}\)). A multi-level reset strategy is integrated to handle stagnation, where different levels are reset adaptively based on the state of the population. This layered structure ensures comprehensive exploration and exploitation across multiple scales, facilitating the effective identification of multiple optima in MMOPs.

\begin{table}[]
    \centering
    \caption{Summary of ML-based DE.}
    \label{tab:ML_DE}
    \resizebox{1\linewidth}{!}{
    \begin{tabular}{c|l|c|p{6.5cm}|p{3.5cm}|p{4.5cm}|p{6cm}}\hline
    S. No. & Variants & Year & Description & Applications & Results & Future Work \\\hline
    1 & DE-RLFR~\cite{li2019differential} & 2019 & Uses Q-learning with hierarchical state variables and DE mutation operations to find multiple PSs. & MM-MOOPs requiring PF convergence. & Outperforms traditional methods in convergence and diversity. & Explore reinforcement learning in other evolutionary algorithms for broader optimization problems. \\\hline
    2 & MABDE~\cite{agrawal2021improved} & 2021 & Introduces population clustering and MAB strategy for mutation generation. & Tested on 20 CEC 2013 benchmark functions. & Outperforms 15 state-of-the-art algorithms in solution accuracy. & Improve application to high-dimensional and real-world problems. \\\hline
    3 & SA-DQN-DE~\cite{liao2024differential} & 2024 & Adapts mutation strategies using DRL and feedback, preserving historical individuals for efficient search. & Applied to MMOPs tested on CEC 2013. & Competitive performance in solving MMOPs. & Integrate more mutation strategies and apply them to real-world problems like job scheduling. \\\hline
    4 & D/REM~\cite{gao2021solving} & 2021 & Combines population update and mean-shift clustering in the PSD phase, using RBF surrogate models in the LSP phase to solve EMMOPs. & Designed for expensive MMOPs with costly objective functions. & Outperforms existing methods in optimization performance. & Improve surrogate model quality in high-dimensional spaces. \\\hline
    5 & MaHDE~\cite{hong2020multi} & 2020 & Balances exploration and exploitation with FHM, DGS, and ELS strategies for diverse and accurate solutions. & Applied to MMOPs tested on CEC 2013. & Outperforms other algorithms in solution quality and convergence. & Extend to real-world applications like project scheduling and economic game problems. \\\hline
    \end{tabular}}
\end{table}

\section{Multiple strategies-based DE}\label{sec: multi objective DE}
% \subsection{Constrained Multi-objective DE}
In this section, we discuss different strategies based on DE for multimodal multi-objective optimization problems (MM-MOOPs). MM-MOOP involves multiple conflicting objectives and constraints. These constraints divide the search domain into feasible and infeasible regions, making constrained MOPs more challenging than standard MM-MOOPs. A constrained MM-MOOP can be mathematically defined as~\cite{zhang2022two}:

\[
\min F(\mathbf{x}) = \left(f_1(\mathbf{x}), \dots, f_m(\mathbf{x})\right)^T 
\]
\[
\text{s.t. } 
\begin{cases} 
g_j(\mathbf{x}) \leq 0, & j = 1, \dots, p, \\
h_j(\mathbf{x}) = 0, & j = p+1, \dots, p+q, 
\end{cases}
\]
where $\mathbf{x} \in \Omega \subseteq \mathbb{R}^d$ represents the decision variables in a $d$-dimensional space, $F(\mathbf{x}) : \Omega \to \mathbb{R}^m$ represents $m$ objective functions, $g_j(\mathbf{x})$ are $p$ inequality constraints, and $h_j(\mathbf{x})$ are $q$ equality constraints.

The constraint violation for $\mathbf{x}$ under the $j$th constraint is calculated as:
\[
c_j(\mathbf{x}) = 
\begin{cases} 
\max(0, g_j(\mathbf{x})), & j = 1, \dots, p, \\
\max(0,|h_j(\mathbf{x})-\delta|), & j = p+1, \dots, p+q,
\end{cases}
\]
where $\delta$ is a tolerance value for equality constraints. The total constraint violation is defined as:
\[
CV(\mathbf{x}) = \sum_{j=1}^{p+q} c_j(\mathbf{x}).
\]

\begin{figure}
\begin{tikzpicture}[scale=1.5]

% Axes for decision space
\draw[->,line width=2pt] (-0.5,-1) -- (4.5,-1) node[anchor=north] {\(x_1\)};
\draw[->,line width=2pt] (-0.5,-1) -- (-0.5,3) node[anchor=east] {\(x_2\)};

% Axes for objective space
\draw[->,line width=2pt] (5.5,-1) -- (9.5,-1) node[anchor=north] {\(f_1\)};
\draw[->,line width=2pt] (5.5,-1) -- (5.5,3) node[anchor=east] {\(f_2\)};

% Feasible areas in decision space
\fill[gray!50,rotate=-30] (0,2.5) ellipse (1.5cm and 0.7cm); % CPS1
\fill[gray!50,rotate=10] (1,-0.45) ellipse (1.5cm and 0.7cm); % CPS2

% Feasible areas in objective space
\fill[gray!50] (7.5,1) ellipse (1.5cm and 0.8cm); % CPF

% Labels for feasible areas
\node[anchor=west] at (2.6,2.1) {\textbf{CPS\(_1\)}};
\node[anchor=west] at (2.6,-0.6) {\textbf{CPS\(_2\)}};
\node[anchor=west] at (9.2,1.3) {\textbf{CPF}};

% Points and connections for CPS1
\node[circle,fill=blue,inner sep=4pt] (a1) at (0.6,2.2) {};
\node[rectangle,fill=orange,inner sep=5pt] (a2) at (1.4,1.7) {};
\node[regular polygon,regular polygon sides=3,fill=green,inner sep=3pt] (a3) at (0.9,2.7) {};
 \draw[thick,red,dashed,line width=1.5pt] (a2) .. controls (0,2.2) and (0.8,2.7) .. (a3);

% \draw[dashed] (a1) -- (a3);
% \draw[dotted,thick] (a1) -- (a2);

% Points and connections for CPS2
\node[circle,fill=blue,inner sep=4pt] (b1) at (0.7,-0.3) {};
\node[rectangle,fill=orange,inner sep=5pt] (b2) at (1.5,-0.8) {};
\node[regular polygon,regular polygon sides=3,fill=green,inner sep=3pt] (b3) at (1.2,0.2) {};
\draw[thick,red,dashed,line width=1.5pt] (b2) .. controls (0,-0.3) and (1,0.2) .. (b3);

% \draw[dashed] (b1) -- (b3);
% \draw[dotted,thick] (b1) -- (b2);

% Connections to objective space
\draw[->,thick,line width=2pt,purple] (1.6,1.7) -- (6.8,0.8); % From CPS1 to CPF
\draw[->,thick,line width=2pt,purple] (1.7,-0.6) -- (6.8,0.6); % From CPS2 to CPF

% Points in objective space (CPF)
\node[circle,fill=blue,inner sep=4pt] (c1) at (7.5,1.6) {};
\node[rectangle,fill=orange,inner sep=5pt] (c2) at (7,0.8) {};
\node[regular polygon,regular polygon sides=3,fill=green,inner sep=3pt] (c3) at (8,1.1) {};
\draw[thick,red,dashed,line width=1.5pt] (c2) .. controls (7.5,2.37) and (8,1.1) .. (c3);

% \draw[dashed,red] (c1) -- (c3);
% \draw[dotted,thick,red] (c1) -- (c2);

% Labels for points
\node[anchor=west] at (0.1,2.3) {\(a_1\)};
\node[anchor=west] at (1.4,1.5) {\(a_2\)};
\node[anchor=west] at (0.6,3) {\(a_3\)};

\node[anchor=west] at (0.3,-0.5) {\(b_1\)};
\node[anchor=west] at (1.7,-0.8) {\(b_2\)};
\node[anchor=west] at (0.8,0.4) {\(b_3\)};

\node[anchor=west] at (7.6,1.6) {\(c_1\)};
\node[anchor=west] at (7,0.6) {\(c_2\)};
\node[anchor=west] at (8.1,1.0) {\(c_3\)};

% Legend
% \begin{scope}[shift={(8.5,2)}]
%     \node[rectangle,fill=orange,inner sep=2pt] at (-2.5,0.5) {};
%     \node[anchor=west] at (-2.3,0.5) {Square};
%     \node[circle,fill=blue,inner sep=2pt] at (-1,0.5) {};
%     \node[anchor=west] at (-0.8,0.5) {Circle};
%     \node[regular polygon,regular polygon sides=3,fill=green,inner sep=2.5pt] at (0.5,0.5) {};
%     \node[anchor=west] at (0.7,0.5) {Triangle};
% \end{scope}

\end{tikzpicture}
\caption{Example of constrained MM-MOOPs}\label{fig: CMM-MOOPs}
\end{figure}
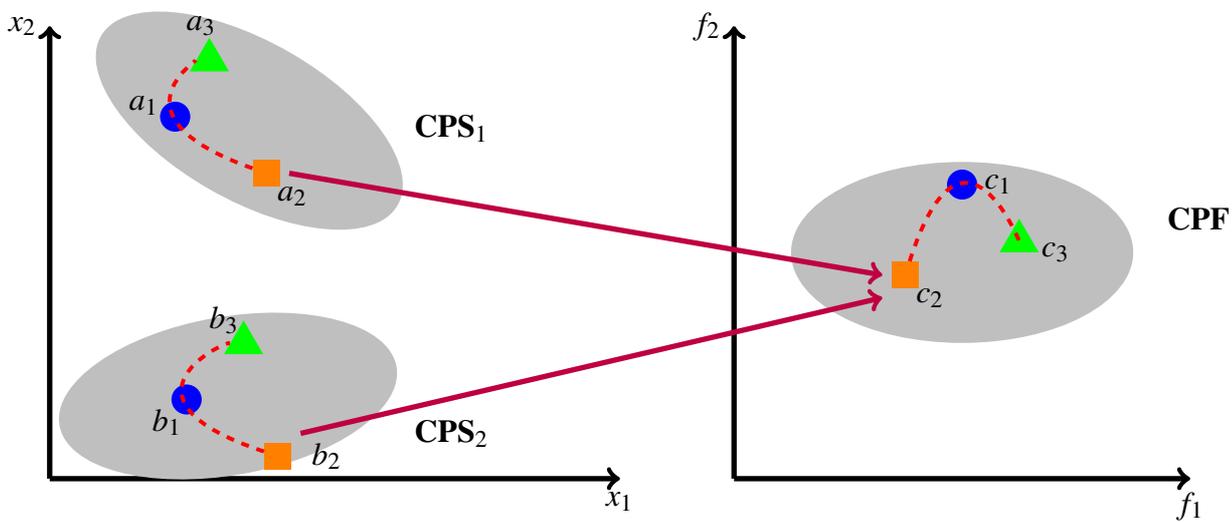

An individual $\mathbf{x}$ is feasible if $CV(\mathbf{x}) = 0$; otherwise, it is infeasible. The feasible domain is where all feasible solutions lie. Pareto-optimal solutions in this domain form the constrained Pareto optimal set (CPS), and their mapping in the objective space constitutes the constrained PF (CPF)~\cite{qiao2023self,liang2022survey}. Fig.~\ref{fig: CMM-MOOPs} illustrates a constrained MM-MOOP instance, where the decision space contains two equivalent CPS regions (CPS$_1$ and CPS$_2$), both mapping to the same CPF in the objective space. For example, point $a_2$ on CPS$_1$ and $b_2$ on CPS$_2$ correspond to the same point $c_2$ on the CPF. 
To tackle these challenging problems, Liang et al.~\cite{liang2022multiobjective} proposed a speciation-based DE algorithm. This method creates niches to preserve a diverse set of feasible Pareto-optimal solutions and employs an improved environmental selection criterion to enhance solution diversity. The algorithm effectively identifies feasible solutions while maintaining a well-distributed PF. Additionally, this work introduced a set of constrained MM-MOOP test functions, providing four types of test cases and seventeen example problems to evaluate algorithm performance.

\begin{figure}
    \centering
 \begin{tikzpicture}
    % Draw the main axes
    \draw[->, thick] (0, 0) -- (6, 0) node[right] {$x_1$};
    \draw[->, thick] (0, 0) -- (0, 6) node[above] {$x_2$};

    % Draw subpopulation ellipses
    \draw[thick, dashed] (1.5, 1) ellipse (1.2cm and 0.7cm) node[below=0pt] {$SubP_1$};
    \draw[thick, dashed] (4.5, 1) ellipse (1.2cm and 0.7cm) node[below=0pt] {$SubP_n$};
    \draw[thick, dashed] (3.8, 3.5) ellipse (1.2cm and 0.7cm) node[below=0.2pt] {$SubP_3$};
    \draw[thick, dashed] (1.5, 4.5) ellipse (1.2cm and 0.7cm) node[below=0pt] {$SubP_2$};

    % Draw the zoomed-in box
    \draw[thick, dashed] (4, 4) rectangle (9, 9);
    % \node at (6.5, 8.5) {SubP1};

    % Draw zoom-in arrows
    \draw[arrow, thick, blue,line width=4pt] (1.5, 1.7) -- (4, 4);

    % Draw the inner circle
    \draw[thick] (6.5, 6.5) circle (2.5cm);

    % Draw the mutant vector and reverse vector
    \filldraw[black] (6.5, 8.4) circle (4pt) node[above right] {$P_1$};
    \filldraw[cyan] (8, 6) circle (4pt) node[below right] {$P_2$};
    \filldraw[orange] (7.5, 6.8) circle (4pt) node[right] {$\mathbf{x}_{r1}$};
    \draw[red, thick, ->,line width=2pt] (7.45, 6.9)--(6.6, 8.35) node[midway, right] {\small The mutant vector $\mathbf{v_{P_1}}$};
    \draw[thick, ->,dashed,line width=2pt] (7.6, 6.6)--(7.95,6.2);
    % Draw supporting points
    \filldraw[blue] (5.5, 7) circle (3pt) node[left] {$\mathbf{x}_{r3}$};
    \filldraw[blue] (6.5, 7) circle (3pt) node[right] {$\mathbf{x}_{r2}$};
    % \filldraw[blue] (5.5, 6) circle (3pt) node[left] {Xr3};
    \filldraw[blue] (7, 5.5) circle (3pt) node[right] {$\mathbf{x}_{r4}$};
    \filldraw[blue] (6.5, 6) circle (3pt) node[right] {$\mathbf{x}_{r5}$};

    % Arrows from supporting points to the center
    \draw[->, thick, blue] (4, 4)--(5.45, 6.95);
    \draw[->, thick, blue] (4, 4)--(6.45, 6.95);
    % \draw[->, thick, blue] (4, 4)--(5.45, 5.95);
    \draw[->, thick, blue] (4, 4)--(6.95, 5.45);
    \draw[->, thick, blue] (4, 4)--(6.45, 5.95);

    % Add labels
    \node[below right] at (8, 6.8) {\small The reverse vector $\mathbf{v}_{P_2}$};
    \node[below] at (7.5, 4) {\small The difference and inverse vector};
\end{tikzpicture}
    \caption{Details of the mutant and opposition-based vector~\cite{gu2024multimodal}.}
    \label{fig: mutant and opposition-based}
\end{figure}
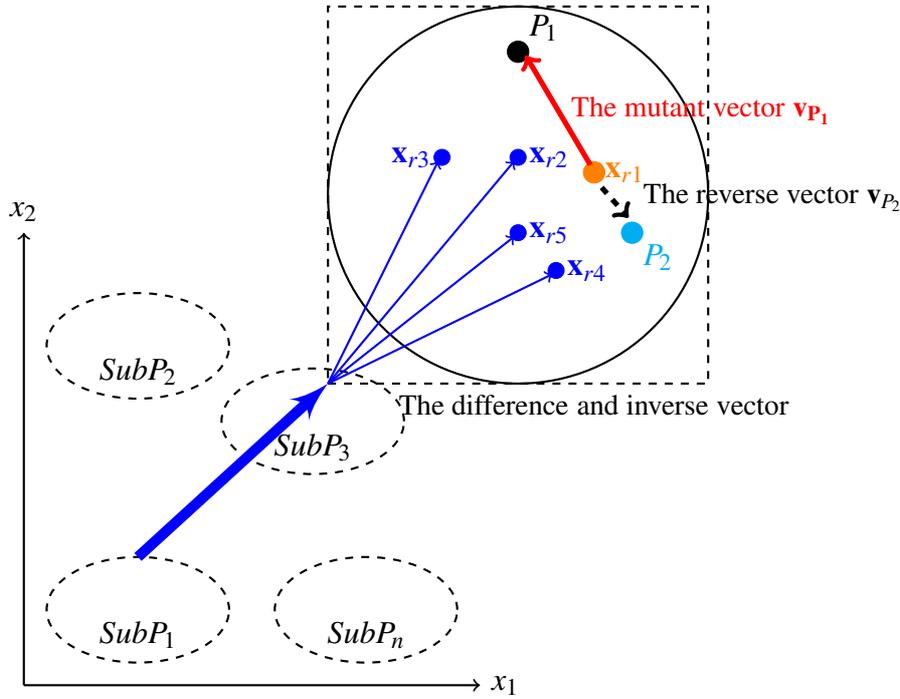

Building upon this, Li et al.~\cite{li2024dynamic} proposed a dynamic speciation-based DE (DSRDE) with a ring topology for solving constrained MM-MOOPs. In DSRDE, a dynamic speciation strategy partitions the population $Pop$ into $N_s$ species, $Pop = \{s_1, s_2, \ldots, s_{N_s}\}$, to explore diverse and equivalent CPS regions. The number of species $N_s$ dynamically decreases during the optimization process to focus on better-converging CPSs. The number of species is calculated as: 
\begin{equation} N_s = \lceil N_p / n_m \rceil, \quad n_m = \lceil\left( \left(\frac{3 - n_{m_{\text{max}}}}{\text{MaxFE}} \right) \cdot FE + n_{m_{\text{max}}} \right)\rceil, \end{equation}
where $n_{m_{\text{max}}}$ is the maximum number of species. The population $\mathbf{P}$ is sorted in ascending order of fitness. Each $\mathbf{x}_i$ is selected as the seed for species $s_i$, and its $(N_s - 1)$ nearest neighbors form the set $nr$. Together, $\mathbf{x}_i$ and $nr$ define species $s_i$, which is then removed from $\mathbf{P}$. A ring topology-based mutation strategy, \textit{DE/current-to-rand/1}, is employed to enhance diversity and feasibility:
\begin{equation}\label{eq: topology mutation} \mathbf{v}_i = \mathbf{x}_i + F(\mathbf{x}_i - \mathbf{x}_{r1}) + F(\mathbf{x}_{r2} - \mathbf{x}_{r3}), \end{equation}
where $\mathbf{x}_i$ is an individual in species $s_i$, and $\mathbf{x}_{r1}$, $\mathbf{x}_{r2}$, and $\mathbf{x}_{r3}$ are distinct individuals from neighboring species $\{s_{i-1}, s_i, s_{i+1}\}$. DSRDE utilizes constrained dominance principles~\cite{gu2022constrained} to identify diverse CPSs effectively. Furthermore, Gu et al.~\cite{gu2024multimodal} proposed a local optimal neighborhood crowding distance DE (LOMMODE$\_$NCD) algorithm. This approach combines an adaptive partitioning strategy during initialization to identify local optima rapidly, opposition-based learning with differential mutation (illustrated in Fig.~\ref{fig: mutant and opposition-based}), and a neighborhood crowding distance method. The latter employs weighted Euclidean distances to balance computational efficiency and crowding degree estimation. Together, these strategies enhance convergence and diversity in both decision and objective spaces.

% \subsection{Unconstrained Multi-objective DE}
To tackle MM-MOOPs, several novel DEs have been developed to address challenges such as maintaining population diversity, balancing convergence, and ensuring an efficient exploration of the decision space. One such approach is the MMODE algorithm~\cite{liang2019multimodal}, which is specifically designed for MM-MOOPs with a Pareto multimodal structure. In these problems, the Pareto set consists of multiple disjoint subsets, all mapping to the same PF. MMODE introduces a boundary-handling technique alongside a classical DE mutation scheme to improve mutation performance. The offspring exceeding the search boundaries are given a second chance to mutate:
\begin{equation}
\mathbf{v}_i = \mathbf{x}_{r1} - \phi (\mathbf{x}_{r2} - \mathbf{x}_{r3}) - \phi (\mathbf{x}_{r4} - \mathbf{x}_{r5}),
\end{equation}
which reduces boundary crowding. Additionally, crowding distance metrics in both decision and objective spaces act as a preselection mechanism to maintain diversity. This approach is further detailed in a two-phase selection process\footnote{The scheme's first phase generates a new population \( \mathbf{P}' \) from an initial population \( \mathbf{P} \) of size \( N_p \). Individuals in \( \mathbf{P} \) are sorted by nondomination ranking, using crowding distance in the objective space to form subsets \( R_1, R_2, \ldots, R_t \). The top-ranked subset \( R_1 \) in \( P' \) contains the best-nondominated solutions, followed by other subsets ordered by decreasing crowding distance. In the second phase, a population \( \mathbf{Q} \) of size \( \lfloor N_p/2 \rfloor \) is selected. If \( R_1 \) has at least \( \lfloor N_p/2 \rfloor \) solutions, the first \( \lfloor N_p/2 \rfloor \) solutions from \( R_1 \) form \( \mathbf{Q} \), and the remaining elements are disregarded. If \( R_1 \) is smaller than \( \lfloor N_p/2 \rfloor \), all of \( R_1 \) is included in \( \mathbf{Q} \), and additional members are selected from subsequent sets \( R_2, R_3, \ldots \), until \( \mathbf{Q} \) reaches \( \lfloor N_p/2 \rfloor \). To finalize \( \mathbf{Q} \), a truncation process limits its size to \( \lfloor N_p/2 \rfloor \) if necessary. If the total elements in \( R_1 \) to \( R_t \) exceed \( \lfloor N_p/2 \rfloor \), solutions in \( R_t \) are sorted by decision-space crowding distance, and the highest-ranked solutions are added until \( \mathbf{Q} \) reaches the desired size.}, where individuals are sorted by their nondomination ranking and then selected based on crowding distance. The resulting population helps preserve diversity while ensuring a balanced representation of solutions.

Further enhancing the capability of DE algorithms for MM-MOOPs, Yu et al.~\cite{yu2018tri} proposed a tri-objective DE (triDEMO) method. By transforming the MM-MOOP into a tri-objective optimization problem (TOP), triDEMO incorporates the original objective function, individual distance information, and shared fitness derived from niching techniques. This transformation effectively handles the inherent conflict between the first two objectives, while the third objective enhances population diversity. The mathematical foundation shows that the Pareto-optimal front of the TOP includes all global optima of the original MM-MOOP, allowing standard DE-based multi-objective techniques to be applied with greater effectiveness.
% \textbf{triDEMO:}  Leveraging the strengths of evolutionary multiobjective optimization to maintain population diversity, Yu et al.~\cite{yu2018tri} devised a tri-objective DE (triDEMO) approach for solving MM-MOOPs. In their method, the MMOP is first transformed into a tri-objective optimization problem (TOP), encompassing three optimization objectives: (i) The MMOP's original objective function; (ii) Individual distance information, measured using a set of reference points; and (iii) Shared fitness, derived from the niching technique. The first two objectives inherently conflict, allowing for the effective use of evolutionary multiobjective optimization. The third objective, driven by the niching technique, significantly enhances population diversity and is robust against niching parameter variations. Mathematical proofs establish that the Pareto-optimal front of the TOP encompasses all global optima of the original MMOP. Subsequently, DE-based multiobjective optimization techniques are applied to tackle the converted TOP. Additionally, they introduce a modified solution comparison criterion and an adaptive ranking strategy for DE to enhance solution accuracy.
A distributed DE approach, DI-MODE~\cite{wang2024distributed}, introduces a virtual population mechanism to further enhance diversity and find additional non-dominated solutions~\cite{chen2019distributed}. The algorithm mimics natural lifespans by archiving individuals as the search area narrows and reinitializing them to maintain population diversity. A probability-based selection strategy balances exploration and exploitation, thus improving the ability to find diverse and well-distributed solutions.

% \textbf{DI-MODE:} A distributed individuals and lifetime mechanism-based DE (DI-MODE) is proposed~\cite{wang2024distributed} for MM-MOOPs. It employs a virtual population approach with an adaptive adjustment strategy to enhance diversity and locate additional non-dominated solutions~\cite{chen2019distributed}. Inspired by natural lifespans, individuals are archived as the search area narrows and reinitialized to maintain diversity. A probability-based selection strategy balances exploration and exploitation during environmental selection, improving the algorithm's ability to find diverse, well-distributed solutions in decision and objective spaces.
For higher-dimensional MM-MOOPs, the HDMMODE algorithm~\cite{liang2024multiobjective} incorporates three distinct mutation strategies: \textit{DE/current-to-best/1/bin}, \textit{DE/current-to-better/1/bin}, and \textit{DE/current-to-rand/1}. Here, \textit{best} is a randomly selected non-dominated solution from the convergence archive and population. In contrast, \textit{better} refers to a solution with a lower Pareto dominance rank and higher diversity value than $\mathbf{x}_i$. The population is divided into two subpopulations: one focuses on improving solution quality through a convergence of external archives and preference-based strategies. At the same time, the other emphasizes exploration using penalty strategies and an enhanced environmental selection method. This improved method effectively estimates solution density in higher-dimensional spaces, enhancing search efficiency and maintaining diversity. Additionally, a new set of higher-dimensional MM-MOOPs is introduced to evaluate the algorithm.

In a similar vein, MMODE$\_$AP~\cite{qu2024improved} integrates an adaptive mutation strategy to address challenges in solving the CEC-2020 MM-MOOPs. By balancing exploration and exploitation, MMODE$\_$AP ensures that solutions are well-distributed across both the decision space and objective space. The algorithm also introduces a crowding degree definition in both spaces, using non-dominated sorting with specialized crowding distances to improve environmental selection and maintain local Pareto sets (PS). Additionally, local PF membership and a predefined parameter are employed to maintain local PSs and solutions near the global PS. Liang et al.~\cite{liang2021clustering} introduced a DE algorithm featuring clustering techniques and a novel elite selection mechanism, called MMODE$\_$CSCD. This method uses a clustering-based special crowding distance (CSCD) method to compute a comprehensive crowding degree in both decision and objective spaces, addressing limitations of the standard special crowding distance (SCD). To illustrate the essence of CSCD, consider a solution set comprising nine solutions divided into three classes such as the first class contains (6,7,8,9) solutions, the second class contains (1,2,3,4) solutions, and the third one has only (5) solutions. The CSCD calculation, denoted as CSCD$_8$, involves determining CD$_{8,x}$ and CD$_{8,f}$. For instance, CD$_{8,x}$ is computed as a function of $x$ values and distances among solutions. CSCD$_8$ is evaluated based on a threshold, CD$_{avg,x}$, and CD$_{avg,f}$, ensuring a well-distributed population in both spaces. The algorithm also introduces a distance-based elite selection mechanism (DBESM) to identify learning exemplars for individuals. New individuals are generated around these exemplars, aiming to achieve a balanced population distribution in decision and objective spaces.

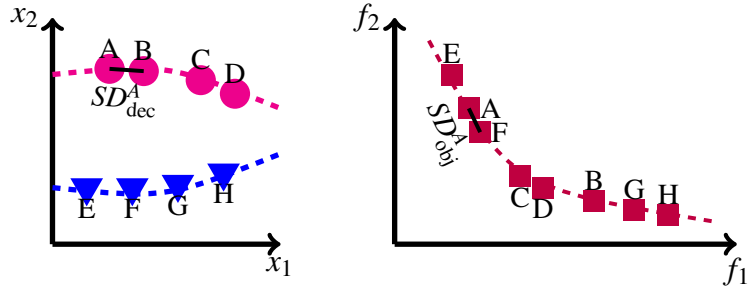
\begin{figure}
\centering
 \begin{tikzpicture}[scale=1.5]

% Decision space
% \node at (-0.3, 2) {(a) Decision Space};
\draw[->, thick,line width=2pt] (0,0) -- (2,0) node[below] {$x_1$};
\draw[->, thick,line width=2pt] (0,0) -- (0,2) node[left] {$x_2$};

% Feasible areas (dotted curves)
\draw[dashed, magenta, thick,line width=2pt] (0,1.5) .. controls (1,1.6) .. (2,1.2); % PS1
\draw[dashed, blue, thick,line width=2pt] (0,0.5) .. controls (1,0.4) .. (2,0.8); % PS2

% % Nodes for PS1 (circles)
 \node[circle, fill=magenta, minimum size=2pt] (A_dec) at (0.5,1.55) {};
 \node[circle, fill=magenta, minimum size=2pt] (B_dec) at (0.8,1.53) {};
 \node[circle, fill=magenta, minimum size=2pt] (C_dec) at (1.3,1.45) {};
 \node[circle, fill=magenta, minimum size=2pt] (D_dec) at (1.6,1.33) {};

% % Nodes for PS2 (triangles)
\node[regular polygon, regular polygon sides=3, fill=blue, inner sep=2.5pt, rotate=180] (E_dec) at (0.3,0.5) {};
\node[regular polygon, regular polygon sides=3, fill=blue, inner sep=2.5pt, rotate=180] (F_dec) at (0.7,0.5) {};
\node[regular polygon, regular polygon sides=3, fill=blue, inner sep=2.5pt, rotate=180] (G_dec) at (1.1,0.53) {};
\node[regular polygon, regular polygon sides=3, fill=blue, inner sep=2.5pt, rotate=180] (H_dec) at (1.5,0.63) {};

% % Label and arrow for shortest distance (SD_dec^A)
 \draw[thick,line width=1.5pt] (0.5,1.55) -- (0.8,1.53) node[midway, below, sloped] {\small $SD_{\text{dec}}^A$};

% % Objective space
% \node at (3.5, 2) {(b) Objective Space};
\draw[->, thick,line width=2pt] (3,0) -- (6,0) node[below] {$f_1$};
\draw[->, thick,line width=2pt] (3,0) -- (3,2) node[left] {$f_2$};

% % Feasible areas (dotted curve)
 \draw[dashed, purple, thick,line width=1.5pt] (3.3,1.8) .. controls (4,0.5) .. (5.8,0.2);

% % Nodes for PF (squares)
\node[rectangle, fill=purple,inner sep=4pt] (E_obj) at (3.5,1.5) {};
\node[rectangle, fill=purple,inner sep=4pt] (A_obj) at (3.65,1.2) {};
\node[rectangle, fill=purple,inner sep=4pt] (F_obj) at (3.75,0.99) {};
\node[rectangle, fill=purple,inner sep=4pt] (C_obj) at (4.1,0.6) {};
\node[rectangle, fill=purple,inner sep=4pt] (D_obj) at (4.3,0.5) {};
\node[rectangle, fill=purple,inner sep=4pt] (B_obj) at (4.75,0.38) {};
\node[rectangle, fill=purple,inner sep=4pt] (G_obj) at (5.1,0.3) {};
\node[rectangle, fill=purple,inner sep=4pt] (H_obj) at (5.4,0.26) {};

% % Label and arrow for shortest distance (SD_obj^A)
\draw[thick,line width=1.5pt] (3.65,1.2) -- (3.75,0.99) node[midway, below, sloped]at(3.75,1.5){\small $SD_{\text{obj}}^A$};

% % Mapping arrows
% \draw[->, thick, dashed] (A_dec) -- (A_obj) node[midway, above, sloped] {\small Mapping};
% \draw[->, thick, dashed] (B_dec) -- (B_obj);
% \draw[->, thick, dashed] (C_dec) -- (C_obj);
% \draw[->, thick, dashed] (D_dec) -- (D_obj);

% \draw[->, thick, dashed] (E_dec) -- (E_obj);
% \draw[->, thick, dashed] (F_dec) -- (F_obj);
% \draw[->, thick, dashed] (G_dec) -- (G_obj);
% \draw[->, thick, dashed] (H_dec) -- (H_obj);

% % Labels for nodes
\node[above] at (A_dec) {\small A};
\node[above] at (B_dec) {\small B};
\node[above] at (C_dec) {\small C};
\node[above] at (D_dec) {\small D};

\node[below] at (E_dec) {\small E};
\node[below] at (F_dec) {\small F};
\node[below] at (G_dec) {\small G};
\node[below] at (H_dec) {\small H};

\node[right] at (A_obj) {\small A};
\node[above] at (B_obj) {\small B};
\node[below] at (C_obj) {\small C};
\node[below] at (D_obj) {\small D};

\node[above] at (E_obj) {\small E};
\node[right] at (F_obj) {\small F};
\node[above] at (G_obj) {\small G};
\node[above] at (H_obj) {\small H};

\end{tikzpicture}
    \caption{Example of shortest distance~\cite{jia2024novel}.}\label{fig: SD}
\end{figure}

To address diversity across decision and objective spaces, Yue et al.~\cite{yue2021differential} developed an improved CD-based DE, MMODE$\_$ICD. By considering diversity-aware difference vectors and weighted Euclidean distances to neighboring solutions, this method prioritizes the selection of diverse solutions, even among lower-ranked individuals. It prevents overcrowding at the top-ranked solutions, thus supporting diversity across both spaces.
% \textbf{MMODE$\_$ICD:} Yue et al.~\cite{yue2021differential} proposed an improved CD-based DE (MMODE$\_$ICD) to address MM-MOOPs. In MMODE$\_$ICD, a diversity-aware difference vector accounts for diversity across decision and objective spaces. The enhanced CD is calculated by considering all selected individuals rather than restricting by Pareto rank (the difference between CD and ICD is illustrated in \textbf{Fig. S6}.). It uses a weighted Euclidean distance to neighbors in the decision space to prioritize proximity-based selection. Overcrowded top-rank individuals are not always selected, allowing lower-rank potential solutions to evolve and supporting diversity across both spaces.
% Another notable DE approach is MMODE\_SDNR~\cite{jia2024novel}, which introduces the Shortest Distance (SD) criterion for sorting individuals. The SD value measures the Euclidean distance between an individual and its nearest neighbors in both decision and objective spaces. Using fast non-dominated sorting and SD, this approach enhances the diversity of the population by selecting learning exemplars based on SD values and incorporating them into a modified mutation strategy. Additionally, the nearest neighbor repulsion (nNR) strategy ensures that similar individuals are removed from the population, thus maintaining diversity.
Another notable DE approach is MMODE\_SDNR~\cite{jia2024novel}, which introduces the shortest distance (SD) criterion for sorting individuals. SD measures the shortest Euclidean distance between an individual and its nearest neighbor in both decision and objective spaces. An example of SD is illustrated in Fig.~\ref{fig: SD}\footnote{As shown in Fig.~\ref{fig: SD}, individual $A$ has two nearest neighbors: $B$ in the decision space and $F$ in the objective space. For individual $C$, its closest neighbor $D$ is the same in both spaces. Taking $A$ as an example, the SD value can be $
     SD(A) = SD_{\text{dec}}^A + SD_{\text{obj}}^A,$ where $SD_{\text{dec}}^A = \text{norm}(A - B),~
     SD_{\text{obj}}^A = \text{norm}(A - F),$ where $SD(A)$ is the combined SD value for individual $A$. The $A$ nearest neighbor $B$ in the decision and objective spaces are $\text{Neigh}_{\text{dec}}^A$ and $\text{Neigh}_{\text{obj}}^A$, respectively.} For a general individual $\mathbf{x}_i$, the SD is defined as:
\begin{equation}
SD^{\mathbf{x}_i} = SD_{\text{dec}}^{\mathbf{x}_i} + SD_{\text{obj}}^{\mathbf{x}_i}, \quad
SD_{\text{dec}}^{\mathbf{x}_i} = \text{norm}(\mathbf{x}_i - \text{Neigh}_{\text{dec}}^{\mathbf{x}_i}), \, 
SD_{\text{obj}}^{\mathbf{x}_i} = \text{norm}(\mathbf{x}_i - \text{Neigh}_{\text{obj}}^{\mathbf{x}_i}).
\end{equation}After fast non-dominated SD sorting, the population is ranked by non-dominated levels, SD values, and neighbors. For an individual $\mathbf{x}_i^k$ (the $i$th solution in the $k$th level), learning exemplars are selected from the $a$th Pareto level, denoted as $F(a)$, where $a = \{1, 2, \dots, k-1\}$ for $k > 1$. Exemplars with larger SD values in $F(a)$ are prioritized, and the top $EXE$ individuals are chosen as candidate exemplars to guide evolution. The modified mutation strategy is used:

\begin{equation}
  \mathbf{v}_i^k = \mathbf{x}_i^k + F_F \cdot (\mathbf{x}_{e1} - \text{Neigh}_{\text{obj}}^{e1}) + F_F \cdot (\mathbf{x}_{e2} - \mathbf{x}_{e3}),  
\end{equation}
where $\mathbf{x}_{e1}$, $\mathbf{x}_{e2}$, and $\mathbf{x}_{e3}$ are randomly selected candidate exemplars, with $\mathbf{x}_{e1}$ being the exemplar closest to $\mathbf{x}_i^k$. Additionally, a novel environmental selection strategy, nearest neighbor repulsion (nNR), is introduced. This strategy removes similar individuals to maintain diversity in the decision space while retaining solutions with similar or identical PFs in the objective space.

The MMODE$\_$SPDN algorithm~\cite{peng2024multimodal} incorporates a dynamic neighbor strategy combined with clustering-based special crowding distance (CSCD) to balance convergence and diversity. With a serial structure, the algorithm uses multiple temporary archives\footnote{Four temporary archives ($AA$, $AB$, $AC$, and $AD$) guide the search process. Archive $AA$ employs a dynamic neighbor-based mutation strategy that selects neighbors based on their Euclidean distance in the decision space. The individual with the largest crowding distance becomes the primary parent $\mathbf{x}_{r1}$ and three auxiliary individuals, $\mathbf{x}_{r2}$, $\mathbf{x}_{r3}$, and $\mathbf{x}_{r4}$, are used for mutation. A polynomial mutation is applied for improved convergence.} to guide the search process, employing dynamic neighbor-based mutation and grid-based strategies. These archives work in tandem to ensure efficient exploration and exploitation, enabling the algorithm to discover diverse solutions while maintaining a focus on convergence. Finally, a ranking method refines the last archive, and the four archives are combined with the population for environmental selection to form the next generation. Zhang et al.~\cite{zhang2020decomposition} introduced a modified version of MOEA/D, referred to as MOEA/D-DE, incorporating a niching strategy. This modification addresses a limitation in MOEA/D, which typically fails to identify more than one optimal solution in the decision space corresponding to the same optimal solution in the PF. MOEA/D-DE has been adapted to balance convergence and diversity in the objective space, while the niching strategy preserves multiple solutions in the decision space. Additionally, it employs a redundant deletion strategy to eliminate redundancy and optimize computational resource usage.

In summary, these innovative DE-based approaches for MM-MOOPs emphasize the critical role of maintaining diversity, balancing exploration and exploitation, and tailoring strategies to address specific challenges, such as higher-dimensional spaces, multimodal PFs, and complex real-world scenarios. Table~\ref{tab: MO_DE} provides a comprehensive discussion of multi-objective (MO) based DE variants, highlighting their key contributions, applications, results, and potential future directions for advancing research in this domain.%Each algorithm brings unique techniques to tackle the challenges inherent in solving MM-MOOPs, ensuring better convergence, diversity, and solution quality.

\begin{table}[]
    \centering
    \caption{Summary of MO-based DE.}
    \label{tab: MO_DE}
    \resizebox{1\linewidth}{!}{
    \begin{tabular}{c|l|c|p{6cm}|p{3.5cm}|p{5cm}|p{5cm}}\hline
    S. No. & Variants & Year & Description & Applications & Results & Future Work \\\hline
    1 & DSRDE~\cite{li2024dynamic} & 2024 & Uses dynamic speciation and ring topology to balance diversity, convergence, and feasibility in constrained MM-MOOPs. & MM-MOOPs, including real-world constrained problems. & Outperforms several state-of-the-art algorithms in constrained MM-MOOPs. & Focus on balancing objective and decision space performance. Test on more practical constrained MM-MOOPs. \\\hline
    2 & MMODE~\cite{liang2019multimodal} & 2019 & Combines neighborhood information and data interpolation, with dual-space crowding distance for improved population distribution. & MM-MOOPs requiring evenly distributed PSs. & Outperforms 8 algorithms in generating high-quality solutions. & Improve parental solution usage and clustering for better convergence. \\\hline
    3 & triDEMO~\cite{yu2018tri} & 2018 & Transforms MMOP into tri-objective optimization, enhancing diversity and accuracy with evolutionary multiobjective techniques. & Multi-objective optimization, balancing convergence and diversity. & Competitive performance across 44 benchmarks. & Address challenges in high-dimensional spaces and explore real-world applications. \\\hline
    4 & MMODE$\_$CSCD~\cite{liang2021clustering} & 2021 & Combines clustering with elite selection using CSCD for well-distributed population in decision and objective spaces. & CEC 2019 MM-MOOPs. & Outperforms other algorithms in maintaining diversity and solving MM-MOOPs. & Test with different clustering algorithms and apply them to practical problems. \\\hline
    5 & MMODE$\_$SPDN~\cite{peng2024multimodal} & 2024 & Uses a series-parallel archive structure and dynamic neighbor strategy for diversity and reduced complexity. & Real-world MM-MOOPs requiring multiple optimal solutions. & Outperforms other state-of-the-art algorithms on various problems. & Explore local PS identification and extend to discrete MM-MOOPs. \\\hline
    \end{tabular}}
\end{table}

\section{Real-world Applications}\label{sec: applications DE}
DE has shown considerable success in solving various optimization challenges in engineering design, healthcare, and feature selection tasks. In engineering design, DE has been applied to problems such as protein structure prediction, orbit determination, and fault diagnosis, demonstrating its versatility across domains. In healthcare, DE has been used for brain voxel classification in MRI images, optimizing multi-factor analysis for accurate brain mapping. Meanwhile, in feature selection, DE has proven effective for high-dimensional data, identifying optimal feature subsets for classification tasks while maintaining computational efficiency.

\subsection{Optimization in Engineering Design}
 In protein structure prediction, a memetic algorithm combining DE and local search via protein fragment replacements is applied to find the native protein structure with minimum energy in a multimodal, deceptive landscape~\cite{varela2022niching}. Niching methods (crowding, fitness sharing, and speciation, the detailed is discussed in Algorithms 8 and 9 of the \textit{Supplementary file}) are integrated to yield a diverse set of optimized, structurally distinct protein conformations in different local minima. This approach facilitates efficient energy landscape exploration, providing multiple viable protein structures.

 The enhanced DE~\cite{xie2022multimodal} is applied to the angle-only initial orbit determination (IOD) problem, particularly under scenarios with limited observational arcs. This approach leveraged a coarse-to-fine convergence detector and a two-layer niching technique to balance exploration and exploitation, creating promising solution clusters. A manifold-assisted DE (MA-MMODE) is proposed~\cite{yang2024manifold} to address fault feature selection by modeling it as an MM-MOOP. Fault diagnosis in rotating mechanical systems is essential for ensuring the reliability of industrial equipment. In practical scenarios, various combinations of fault features can achieve similar diagnostic accuracy; however, their computational costs and measurement complexities differ. MA-MMODE enables decision-makers to identify multiple optimal feature subsets, facilitating the selection of those with lower costs while maintaining high diagnostic performance. The MHDE algorithm~\cite{bu2024multi} is applied to the node deployment problem in wireless sensor networks (WSNs). WSNs are distributed networks of sensor nodes that monitor and detect environmental conditions. By deploying these nodes in target areas, WSNs can collect, process, and transmit data efficiently, providing timely feedback to users. MMODE/SC~\cite{ji2023multimodal} is applied to solve two practical problems such as the map-based problem \cite{ishibuchi2011many} and the optimal two-impulse orbit transfer problem \cite{minisci2009orbit}.
 
 The mathematical formulation of these problems is formulated in Section 10 of the \textit{Supplementary file}. The QPSODE algorithm~\cite{fahad2023optimizing} has demonstrated improved performance on constrained, non-convex benchmark problems and superconducting magnetic energy storage (SMES) electromagnetic optimization tasks. MELSHADE-cnEpSin~\cite{zhang2024differential} employed an adaptive selection mechanism (ASM) for crossover rate adjustment, a Sigmoid-based nonlinear population reduction strategy to enhance distribution, and a restart strategy to avoid suboptimal convergence. It was effectively applied to UAV trajectory planning in complex mountainous terrain and tested on point cloud registration using a rapid global registration dataset, demonstrating its capability in solving real-world optimization challenges.

\subsection{Healthcare and Biomedical Problems}
Li et al.~\cite{li2017brain} introduced the VMG-NDE algorithm to enhance brain voxel classification in MRI images by addressing various factors simultaneously. The algorithm comprises four components: (i) Utilizing the variational mixture of the Gaussians (VMG) model to characterize voxel value variations due to partial volume effects (PVE). (ii) Training multiple local VMG models on small data volumes extracted from the image to mitigate intensity non-uniformity (INU) effects. (iii)Employing niche DE (NDE) to infer each local VMG model, preventing convergence to local optima. (iv) Constructing probabilistic brain atlases for each study to incorporate anatomical priors into the classification process. Following the training of local VMG models, brain voxel classification is accomplished through a linear combination of predictions generated by these models. The algorithm's performance has been evaluated against variational expectation-maximization and GA-based segmentation approaches, as well as segmentation routines in popular software packages like statistical parametric mapping (SPM), expectation-maximization segmentation (EMS), and FSL, using both synthetic and clinical T1-weighted brain MRI data.
\subsection{Data-driven and Feature Selection Tasks}
 Both synthetic and real datasets were utilized for evaluation in ref.~\cite{sheng2020differential}. The synthetic datasets were designed with highly overlapping clusters and varying clustering complexities. Seven real-life datasets: Connectionist, MFCCs, Shuttle, Isolet1, Isolet2, Letter, and HAR (human activity recognition using smartphones) were sourced from the UCI Machine Learning Repository. Additionally, the study included two image datasets (Flowers17 and MNIST) and two bioinformatics microarray datasets (Yeast2945 and Cancer728). Similar to FBK-DE~\cite{lin2019differential} and MMODE$\_$ICD~\cite{yue2021differential}, Agrawal et al.~\cite{agrawal2023feature,agrawal2023differential} proposed two DE variants, FSSDE and MMMODE, for feature subset selection. Both incorporate a balanced NBC (similar to NBC-Minsize~\cite{lin2019differential}) and an archive strategy for storing potential optimal solutions. However, FSSDE uniquely employs probability-based initialization to distribute selected features across the search space, and MMMODE employs an adaptive generation strategy. Comprehensive experiments on diverse datasets demonstrated the algorithm's ability to identify multiple feature subsets effectively. A new evaluation metric is also introduced and compared results with existing methods.

 A multimodal bare-bone niching DE (MBNDE)~\cite{hu2021multimodal} is proposed for feature selection (FS) in classification tasks. Three niching strategies are applied: clustering, crowding with speciation, and index-based niching. A modified 3-nearest neighbor (3-NN) classifier evaluates the classification performance of the selected feature combinations. Wang et al.~\cite{wang2022differential} introduced a multi-objective DE based on a niching strategy (NMDE) to minimize both the number of features and the classification error rate. A novel mutation operator in NMDE combines local niche information with global population data to identify high-quality feature subsets. NMDE reframes the environmental selection process as a search for ``$\psi$-quasi equal\footnote{Two feature subsets $S_1$ and $S_2$ from a set $S$ are considered $\psi$-quasi equal if they have the same feature ratio $f_2(S_1) = f_2(S_2)$ and a classification error difference, for given $\psi$ is a small constant $(0 \leq \psi < 1)$, of $|f_1(S_1) - f_1(S_2)| \leq \psi $. If $\psi = 0$, both subsets have identical $f_1$ and $f_2$ values.}" feature subsets, which share similar accuracy levels, by relaxing strict Pareto dominance criteria. Here, two subsets are considered $\psi$-quasi equal if they share the same feature ratio and have a classification error difference within a small constant $\psi$ $(0 \leq \psi < 1)$, preserving multiple subsets with similar accuracies and maintaining solution diversity. Additionally, a subset-repairing mechanism refines feature subsets by addressing three cases of $\psi$-quasi equal subsets, effectively removing redundant features and enhancing classification performance. BNDE~\cite{gong2017learning} has been applied to solve a set of twelve neural network ensemble (NNE) datasets. In this implementation, each individual in BNDE represents a floating-point vector encoding the neural network parameters, where each dimension corresponds to a connection weight. The individuals are randomly initialized and subsequently evolved using the standard procedures of BNDE. To address the challenge of feature selection in classification, a multiobjective differential evolution (MOCDE) approach incorporating a clustering technique has been developed \cite{wang2021multiobjective}. The goal of MOCDE is to identify multiple nondominated feature subsets that offer different trade-offs between the number of selected features and classification accuracy. Additionally, it aims to discover multiple feature subsets with identical objective values but different feature compositions, enhancing diversity and interpretability in feature selection.

MOCDE integrates several key components to improve efficiency and solution quality:
(i) A novel initialization strategy based on feature relevance is introduced to provide a strong starting point for the evolutionary process. The maximal information coefficient (MIC) is used to measure the relevance between each feature and the class label. The population is initialized by selecting top-ranking features while also incorporating features probabilistically based on their MIC values.
(ii) The k-means clustering method is employed to divide the population into multiple subpopulations, fostering diverse search behavior.
(iii) A dynamic SCD mechanism is implemented in each subarchive to maintain solution diversity by considering distances in both the objective and search spaces. If a subarchive surpasses a predefined size, solutions with lower SCD values are removed to ensure balanced exploration and exploitation.
(iv) A refined hypervolume contribution indicator is introduced to measure the convergence of solutions, effectively handling multiple feature subsets that achieve the same classification performance but with different feature combinations.

The effectiveness of MOCDE is evaluated using 14 datasets of varying difficulty from the UCI Machine Learning Repository. It is benchmarked against seven state-of-the-art methods, including multiobjective algorithms (NSGA-II, SPEA2, MOEA/D), a sparse multiobjective feature selection algorithm (SparseEA), and multimodal multiobjective algorithms (Omni-optimizer, DN-NSGAII, MO\_Ring\_PSO). The comparison is conducted using key performance indicators such as hypervolume (IH) and inverted generational distance (IIGD) on both training and testing datasets. The results demonstrate MOCDE's ability to generate diverse and high-quality feature subsets, making it a competitive approach for multiobjective feature selection.

\begin{table}[]
    \centering
    \caption{A list of DE multimodal applications.}
    \label{tab: DE applications}
    \resizebox{1\linewidth}{!}{
    \begin{tabular}{l|ll}
    \hline
    Variant & Application \\\hline
    Varela's DE~\cite{varela2022niching} & Protein structure prediction, identifying multiple viable protein conformations in a multimodal landscape. \\\hline
    Enhanced DE~\cite{xie2022multimodal} & Angle-only initial orbit determination (IOD) problem under limited observational arcs. \\\hline
    MA-MMODE~\cite{yang2024manifold} & Fault feature selection for fault diagnosis in rotating mechanical systems. \\\hline
    MHDE~\cite{bu2024multi} & Node deployment in wireless sensor networks (WSNs). \\\hline
    MMODE/SC~\cite{ji2023multimodal} & Map-based problem and optimal two-impulse orbit transfer problem. \\\hline
    QPSODE~\cite{fahad2023optimizing} & Constrained, non-convex benchmark problems and SMES electromagnetic optimization tasks. \\\hline
    MELSHADE-cnEpSin~\cite{zhang2024differential} & UAV trajectory planning in complex terrain and point cloud registration. \\\hline
    VMG-NDE~\cite{li2017brain} & Brain voxel classification in MRI images. \\\hline
    FBK-DE~\cite{lin2019differential} & Feature selection in classification tasks. \\\hline
    MMODE$\_$ICD~\cite{yue2021differential} & Feature selection in classification tasks. \\\hline
    FSSDE~\cite{agrawal2023feature} & Feature subset selection using a balanced NBC and archive strategy. \\\hline
    MMMODE~\cite{agrawal2023differential} & Feature subset selection with an adaptive generation strategy. \\\hline
    MBNDE~\cite{hu2021multimodal} & Feature selection for classification tasks. \\\hline
    NMDE~\cite{wang2022differential} & Feature subset selection to minimize the classification error rate. \\\hline
    BNDE~\cite{gong2017learning} & Neural network ensemble (NNE) datasets, optimizing neural network parameters. \\\hline
    JADE-STACK \cite{ribeiro2023decoding} & Decoding electroencephalography signal
response\\\hline
DE-PNN \cite{nasim2022pnn} & Imbalanced electrocardiogram (ECG) classification for arrhythmia detection\\\hline
SSODE-GCNDM \cite{babu2024hybridization} & Distributed denial of service detection and mitigation in IOT environment\\\hline
CFD-ML \cite{ran2024development} & Prediction the viscosity of ionic liquids-water mixtures\\\hline
AutoML \cite{vincent2023improved} & Hyperparameter tuning in automated ML systems\\\hline
M-SDE-ELM \cite{xie2022birdsongs} & Birdsongs recognition\\\hline
DBPboost \cite{sun2024dbpboost} & Identification of DNA-binding proteins\\\hline
CSA-DE-LR \cite{dedeturk2024csa} & Cardiovascular disease (CVD) diagnosis \\\hline
QLDE \cite{wang2025customer} & Customer segmentation \\\hline
SADEABC \cite{wang2023extreme} & Whole blood composition analysis\\\hline
\end{tabular}}
\end{table}

\subsection{Machine learning applications}
{ Recent advancements in hybrid evolutionary and machine learning (ML) approaches have demonstrated significant improvements in diverse fields, including biomedical signal processing, cybersecurity, industrial optimization, and automated machine learning (AutoML). By integrating DE with deep learning, probabilistic models, and ensemble learning, these hybrid techniques enhance predictive accuracy, feature selection, and hyperparameter optimization across multiple domains.

A prominent example of this synergy is JADE-STACK, a novel framework for nonlinear system identification of electroencephalography (EEG) signals in response to wrist joint perturbations \cite{ribeiro2023decoding}. JADE-STACK employs stacked generalization (STACK) ensemble learning alongside the adaptive DE algorithm, JADE, to develop a robust predictive model for neural signal identification. The ensemble consists of extreme gradient boosting (XGBoost), Gaussian Process, least absolute shrinkage and selection operator (LASSO), multilayer perceptron neural network (MLP), and support vector regression (SVR), whose predictions are refined by a Cubist meta-learner with JADE-optimized hyperparameters, as shown in Fig. \ref{fig: STACK}. Evaluated on EEG data from ten healthy participants, JADE-STACK demonstrated superior performance in both one-step-ahead and three-step-ahead predictions, assessed using variance accounted for (VAF) and root-mean-squared error (RMSE).

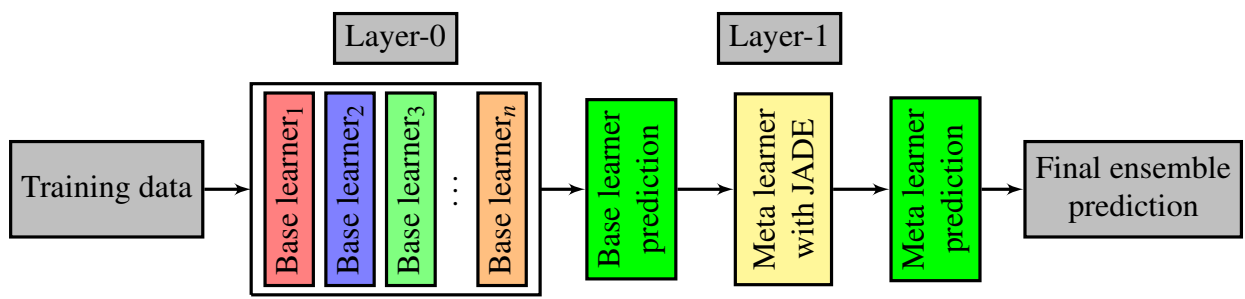
\begin{figure}
	\centering
	\tikzset{
		block/.style={draw, fill=white, rectangle, 
			minimum height=3em, minimum width=6em},
		every path/.style={very thick},
	}
\resizebox{!}{!}{	\begin{tikzpicture}[auto, node distance=2cm, >=latex']%[node distance=1.5cm,auto,,rotate=90,transform shape]
	\tikzset{->-/.style={decoration={
				markings,
				mark=at position #1 with {\arrow{>}}},postaction={decorate}}}
% \draw [help lines] (0,0) grid (14,14);
% Layers
    \node[arrow, fill=gray!50] at (4.4,10) (layer0) {Layer-0};
    \node (layer1) [arrow, fill=gray!50,right=3.4cm of layer0] {Layer-1};
    
    % Training Data
    \node (data) [block, fill=gray!50, minimum width=1.5cm]at (0.6,8) {Training data};

    % Base Learners
    \node (bl1) [arrow, fill=red!50, rotate=90]at (3,8) {Base learner$_1$};
    \node (bl2) [arrow, fill=blue!50, rotate=90]at (3.8,8) {Base learner$_2$};
     \node (bl3) [arrow, fill=green!50, rotate=90]at (4.6,8) {Base learner$_3$};
    \node (dots) [rotate=90]at (5.2,8) {$\cdots$};
    \node (bln) [arrow, fill=orange!50, rotate=90]at (5.8,8) {Base learner$_n$};
    \draw (2.5,6.6)--(6.3,6.6)--(6.3,9.4)--(2.5,9.4)--(2.5,6.6);

     % % Base Learner Prediction
    \node (basepred) [arrow, fill=green, align=center, rotate=90] at (7.5,8) {Base learner\\ prediction};

    % % Meta Learner
    \node (meta) [block, fill=yellow!50, align=center, rotate=90] at (9.5,8) {   Meta learner\\with JADE  };

    % % Meta Learner Prediction
    \node (metapred) [arrow, fill=green, align=center, rotate=90] at (11.5,8) {Meta learner\\ prediction};

    % % Final Prediction
     \node (finalpred) [block, fill=gray!50, align=center] at (14.1,8) {Final ensemble\\ prediction};

     % Arrows
    \draw [arrow] (data) -- (2.5,8);
    \draw [arrow] (6.3,8) -- (basepred);
    \draw [arrow] (basepred) -- (meta);
    \draw [arrow] (meta) -- (metapred);

    \draw [arrow] (metapred) -- (finalpred);
  \end{tikzpicture}}
    \caption{STACK ensemble learning architecture with JADE \cite{ribeiro2023decoding}.} \label{fig: STACK}
\end{figure}

Similarly, DE-based feature optimization has been effectively applied to electrocardiogram (ECG) classification for arrhythmia detection. The DE-PNN framework \cite{nasim2022pnn} enhances probabilistic neural network (PNN) classification by optimizing features extracted from the MIT-BIH Arrhythmia Database (\url{https://physionet.org/}), which contains eight heartbeat classes (one normal and seven arrhythmic types). The dataset, comprising 107,800 heartbeats, was split evenly into training and testing sets. Addressing the imbalance in arrhythmia heartbeat classes, DE-PNN reduces the original 253 features to 36, achieving an 85.77\% reduction in dimensionality while maintaining 99.33\% classification accuracy. Evaluated using multiple metrics, including F1-score, specificity, Matthews correlation coefficient (MCC), and area under the ROC curve (AUC), the approach demonstrates DE's effectiveness in biomedical feature selection.

Beyond healthcare, DE-driven hybrid models play a critical role in cybersecurity, particularly in distributed denial-of-service (DDoS) attack detection for Internet of Things (IoT) environments. The SSODE-GCNDM framework \cite{babu2024hybridization} integrates synergistic swarm optimization and DE with graph convolutional networks (GCNs) to identify network traffic anomalies and mitigate cyber threats. The model consists of Z-score normalization, SSODE-based feature selection, GCN-based attack detection, and the northern goshawk optimization (NGO)-based hyperparameter tuning. First, data normalization is performed using $Z-$score normalization to
standardize input data. This step enhances model stability and ensures that all features contribute equally, preventing biases caused by varying data scales. Second, feature selection is carried out using an SSODE approach, which efficiently identifies the most relevant features, reducing computational complexity while improving detection accuracy. In the third phase, attack detection and mitigation are handled using a GCN, which effectively identifies DDoS attacks by analyzing traffic patterns and structural dependencies within IoT networks. Finally, hyperparameter tuning is conducted using the NGO algorithm.  Inspired by the hunting behavior of northern goshawks, the NGO efficiently explores the hyperparameter space, preventing premature convergence. By integrating swarm intelligence, evolutionary optimization, and DL,
SSODE-GCNDM enhances IoT security, offering a robust and adaptive method for real-time
cyberattack detection and mitigation. SSODE-GCNDM achieved 99.62\% accuracy at 1000 epochs, outperforming other ML techniques such as Logistic Regression, KNN, Random Forest, and Deep Neural Networks, and demonstrating high precision, recall, and G-Means scores.

% \begin{figure}
%     \centering
%     \includegraphics[width=0.8\linewidth]{images/DDoS.png}
%     \caption{General structure of DDoS attack in IOT \cite{babu2024hybridization}.}\label{fig: DDoS}
% \end{figure}
The integration of evolutionary algorithms in industrial optimization is exemplified by a hybrid CFD-ML model for vacuum membrane distillation (VMD) separation processes \cite{ran2024development}. Computational fluid dynamics (CFD) simulations generate synthetic training data, which is used by support vector machines (SVM), elastic net regression (ENR), extremely randomized trees (ERT), and Bayesian ridge regression (BRR). To enhance prediction accuracy, DE optimizes hyperparameters, while Monte Carlo cross-validation (MCCV) ensures robust generalization. MCCV is applied for robust model validation, improving generalization
across different data splits. The dataset used in this study consists of over 13,000 instances, where each data point includes input variables ($r$ and $z$) and the output variable ($T$) obtained from CFD simulations of a hollow-fiber membrane contactor for VMD. Model performance is assessed using multiple evaluation metrics, including the $R^2$ score, mean squared error (MSE), mean absolute error (MAE), and mean absolute percentage error (MAPE). This hybrid approach significantly improved the prediction of temperature distribution within the hollow-fiber membrane contactor, optimizing VMD separation efficiency.

Further advancing AutoML techniques, Vincent and Jidesh \cite{vincent2023improved} explored Bayesian optimization (BO) with evolutionary algorithms like GA, DE, and CMA-ES. By maximizing the expected improvement (EI) acquisition function, BO-DE and BO-CMAES enhance hyperparameter tuning in automated ML systems, improving both efficiency and adaptability. This aligns with Xie et al.'s \cite{xie2022birdsongs} multi-strategy DE-enhanced ensemble learning for birdsong recognition, where an M-SDE-based extreme learning machine (ELM) optimizes input weights and hidden layer thresholds using a combination of \{\textit{DE/rand/2}, \textit{DE/best/2}, \textit{DE/current-to-best/1}\} mutation strategies. Evaluated on nine bird species, M-SDE-ELM and M-SDE-EnELM achieved 86.70\% and 89.05\% classification accuracy, surpassing PSO and GOA-optimized ELM models. DBPboost, a novel model is proposed \cite{sun2024dbpboost} for identifying DNA-binding proteins. The innovation lies in using eight feature extraction methods, an enhanced feature selection process, and an optimized DE algorithm to improve feature fusion. Experimental results demonstrate that DBPboost achieves 89.32\% accuracy and 89.01\% sensitivity on the UniSwiss dataset (\url{https://github.com/jun-csbio/TargetDBPplus}, \cite{hu2021targetdbp+}), outperforming most existing models.

Beyond classification tasks, hybrid DE-ML models have also been applied to cardiovascular disease (CVD) diagnosis and customer segmentation. The CSA-DE-LR framework \cite{dedeturk2024csa} combines clonal selection algorithm (CSA) and DE with logistic regression (LR), replacing gradient descent with DE-based optimization to avoid local minima and enhance classification accuracy. Meanwhile, QLDE-based K-means clustering \cite{wang2025customer} integrates reinforcement learning with DE to improve digital marketing customer segmentation. This framework is used to effectively identify and understand the distinct characteristics and needs of different customer groups, thereby enabling more targeted marketing strategies. Using a Kaggle dataset (\url{https://www.kaggle.com/code/fabiendaniel/customer-segmentation/notebook}), the method transforms eight features into 11 RFM-based features, demonstrating effective clustering and feature selection for targeted marketing strategies. The dataset is
publicly available \url{https://zenodo.org/records/14614253}.

In the domain of biochemical analysis, SADEABC-optimized extreme learning machines (ELMs) have been applied to Fourier Transform Raman spectroscopy for whole blood composition analysis \cite{wang2023extreme}. This method enhances non-invasive blood analysis, offering rapid and accurate predictions of blood component concentrations. In parallel, DE-CNN architectures have been developed for medical image analysis, where DE-encoded CNN structures optimize convolutional layers, filter sizes, and activation functions. Tested on brain MRI, lung, and colon histopathology datasets, DE-CNN achieved competitive accuracies ranging from 78.73\% to 99.50\%, outperforming conventional CNN models.}  We summarized these diverse applications of DE-based approaches across various domains in Table~\ref{tab: DE applications}, highlighting their problem-specific innovations and the potential for addressing real-world challenges effectively.

\section{Experimental analysis}\label{Sec: experimental}
\begin{figure}
\begin{subfigure}[b]{1\linewidth}
\includegraphics[width=0.5\linewidth]{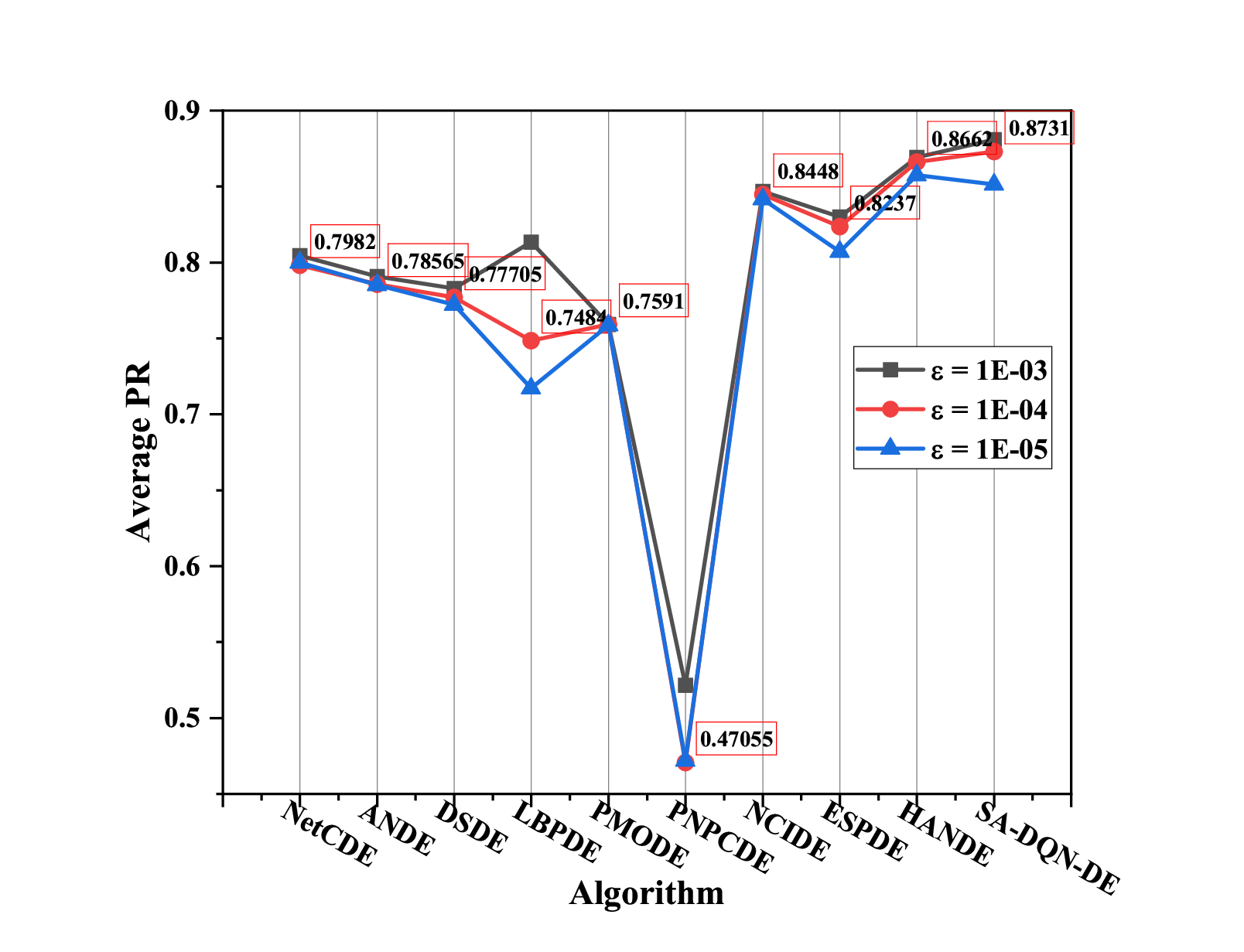}
\includegraphics[width=0.5\linewidth]{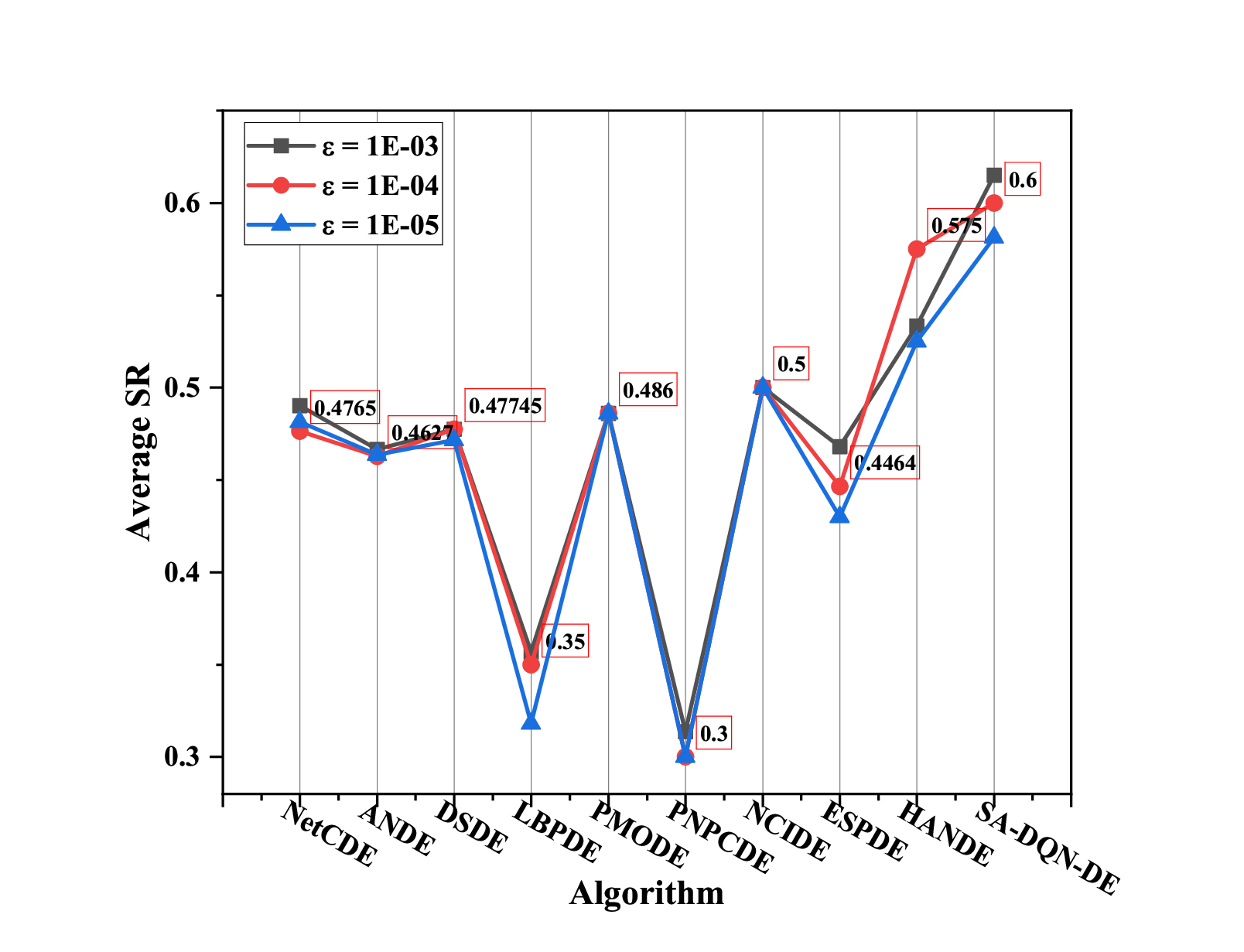}
\end{subfigure}
\caption{PR and SR values of different DE algorithms at different accuracy levels for CEC 2013 MMOP.}\label{fig: results the CEC 2013}
\end{figure}

{ In this section, an experimental analysis of various algorithms is conducted using two IEEE CEC benchmark test suites~\cite{li2013benchmark}: CEC 2013 and CEC 2015. The experimental results for algorithms such as NetCDE~\cite{chen2023network}, ANDE~\cite{wang2020automatic}, DSDE~\cite{wang2017dual}, LBPDE, PNPCDE\cite{biswas2014improved}, NCIDE~\cite{liang2024niche}, ESPDE~\cite{li2023history}, HANDE~\cite{liao2023history}, SA-DQN-DE~\cite{liao2024differential}, LMCEDA~\cite{yang2016multimodal}, LMSEDA~\cite{yang2016multimodal}, LBPADE~\cite{zhoa2020local}, FBK-DE~\cite{lin2019differential}, PMODE~\cite{wei2021penalty}, NCD-DE \cite{jiang2021optimizing}, NMMSO~\cite{li2013benchmark} (Winner of CEC 2015 competition), and WSNADE~\cite{huang2024wireless} are extracted from their respective papers. The comparative analysis is divided into two parts: first, the performance of these algorithms is evaluated on twenty CEC 2013 benchmark problems using metrics such as peak ratio (PR) and success rate (SR)\footnote{$PR=\frac{\sum_{r=1}^{Run}NOF_r}{N\times Run}$ and $SR=\frac{NSR}{Run}$, where $NOF$ and $NSR$ are the number of optima found and number of successful runs.} at three accuracy levels: $\epsilon = 1E{-03}$, $\epsilon = 1E{-04}$, and $\epsilon = 1E{-05}$. Second, their performance is assessed on twenty CEC 2015 benchmark problems using the same metrics and accuracy levels. The benchmark problems are categorized based on their nature to facilitate a structured evaluation. 

It is important to note that due to the unavailability of detailed results for each algorithm's runs, Wilcoxon's signed-rank test~\cite{derrac2011practical} could not be applied to evaluate the statistical significance of the performance differences. Consequently, the relative performance of the algorithms is assessed based on PR values, inspired by the methodology outlined in~\cite{wang2020automatic}. The performance comparison is represented using the symbols '$+$' (better), '$\approx$' (equal), and '$-$' (worse), indicating whether an algorithm outperforms, performs similarly to or underperforms compared to others.

Table~\ref{tab: results PR(SR) for CEC 2013} presents a comparative analysis of the performance of various DE variants on the CEC 2013 MMOPs at a high-precision accuracy level ($\varepsilon = 1E-05$). The metrics used for evaluation include the proportion of runs achieving a solution within the defined accuracy ($PR$) and the success rate ($SR$), with values presented as $PR (SR)$. Across the functions, HANDE, SA-DQN-DE, and NCIDE consistently perform, achieving optimal $PR$ and $SR$ values on most functions, reflecting their strong convergence and reliability. In contrast, variants like PNPCDE, PMODE, DSDE, LBPDE, and ANDE exhibit lower performance on specific functions, such as $F_7$, $F_8$, $F_9$, $F_{13}$, $F_{14}$, $F_{17}$, $F_{19}$, and $F_{20}$, indicating limitations in diversity maintenance or multimodal landscape navigation. The average results ($AVR$) highlight HANDE, NCIDE, and SA-DQN-DE as the top performers, with competitive values across all metrics. Tables S1, S2, and S3 also present similar results and analysis for accuracy levels $\varepsilon = 1E-03$ and $\varepsilon = 1E-04$. The statistical test is performed  SA-DQN-DE $vs.$ other algorithms because SA-DQN-DE performs better than the other algorithms. Notably,``$+/\approx/-$'' comparisons indicate that SA-DQN-DE achieves superior performance in most cases, emphasizing their robustness and effectiveness in tackling high-precision multimodal optimization tasks. These results underline the importance of algorithmic enhancements, such as adaptive mechanisms and diversity-preserving strategies, in addressing challenges in multimodal optimization.

Figure~\ref{fig: results the CEC 2013} illustrates the average $PR$ and $SR$ values of different algorithms across varying accuracy levels ($\varepsilon = 1E{-03}$, $1E{-04}$, and $1E{-05}$) on the CEC 2013 MMOPs. In the left graph, representing the average $PR$, most algorithms, including NCIDE, HANDE, and SA-DQN-DE, maintain consistently high performance across all accuracy levels, with only minor variations. Conversely, PMODE, LBPDE, and PNPCDE demonstrate significant drops in $PR$, indicating their inability to adapt effectively to the increasing precision requirements. The right graph, depicting $SR$, shows similar trends, where NCIDE, HANDE, and SA-DQN-DE achieve superior success rates. Notably, the $SR$ values for PMODE and PNPCDE plummet, reinforcing their struggles in maintaining reliability under challenging optimization scenarios. These results highlight the importance of robust mechanisms in DE variants, such as adaptive learning and diversity preservation, to ensure stable performance across varying problem complexities and accuracy demands.

The experimental results for the CEC 2015 MMOPs of various algorithms at different accuracy levels are presented in Tables S3, S4, and \ref{tab: results PR(SR) for CEC 2015}. Table~\ref{tab: results PR(SR) for CEC 2015} specifically showcases the performance of the algorithms on the CEC 2015 MMOPs at an accuracy level of $\epsilon = 1E{-05}$. From the table, it is evident that several algorithms, such as PNPCDE, LBPADE, and FBK-DE, consistently achieve PR values of 1.0 for the simpler functions (F1$-$F5), demonstrating their high accuracy and robustness in solving these benchmark problems. However, algorithms like NCD-DE and ANDE, while performing well on certain problems, struggle on multimodal and complex landscapes, as reflected by lower PR values for functions F4$-$F6 and F12$-$F20. Similarly, NMMSO exhibits strong performance on a few problems but faces challenges on the later functions. In contrast, WSNADE outperforms all other algorithms, showcasing superior performance across the majority of the benchmark problems.

When analyzing the $AVR$ values at the bottom of the table, WSNADE and ANDE achieve the highest overall PR values (0.82 and 0.79, respectively), followed closely by PNPCDE, LBPADE, and NCD-DE. Among these, WSNADE stands out due to its balance of high PR and SR values, indicating stable and consistent performance across all benchmark problems. Conversely, PNPCDE demonstrates slightly weaker average performance, with PR values below 0.5 for some complex functions, highlighting its challenges in handling more difficult landscapes. The ``$+/\approx/-$" statistics provide a comparative analysis of the algorithms, with WSNADE achieving a higher count of `$+$' outcomes compared to PNPCDE, LBPADE, LMCEDA, LMSEDA, and NMMSO, thereby reflecting its dominance in terms of solution quality. Meanwhile, algorithms such as FBK-DE, NCD-DE, PMODE, and ANDE, while competitive in certain cases, exhibit a greater number of `$\approx$' and `$-$' results.

 In the left graph of Fig.~\ref{Fig: results the CEC 2015}, representing the average $PR$, algorithms such as WSNADE and NMMSO demonstrate consistently high performance across all accuracy levels, with only minor variations. In contrast, algorithms like PNPCDE, LBPADE, LMCEDA, and LMSEDA experience noticeable drops in $PR$, indicating challenges in adapting effectively to increasing precision requirements. The right graph of Fig.~\ref{Fig: results the CEC 2015}, which depicts the average $SR$, shows a similar pattern. Algorithms such as PMODE, WSNADE, and NMMSO achieve superior success rates. These findings underscore the importance of incorporating robust mechanisms in DE variants, such as adaptive learning strategies and diversity preservation techniques, to maintain stable and reliable performance across varying problem complexities and accuracy demands.

}

\begin{table}[]
    \centering
        \caption{Experimental results of various DE variants on CEC 2013 MMOPs at $\epsilon=1E-05$ accuracy level.}\label{tab: results PR(SR) for CEC 2013}
\resizebox{1\linewidth}{!}{    \begin{tabular}{ccccccccccc}\hline
Func	&		NetCDE				&		ANDE				&		DSDE				&		LBPDE				&		PMODE				&		PNPCDE				&		NCIDE				&		ESPDE				&		HANDE				&		SA-DQN-DE				\\\hline
	&		PR	(	SR	)	&		PR	(	SR	)	&		PR	(	SR	)	&		PR	(	SR	)	&		PR	(	SR	)	&		PR	(	SR	)	&		PR	(	SR	)	&		PR	(	SR	)	&		PR	(	SR	)	&		PR	(	SR	)	\\\hline
F1	&	\cellcolor{gray}	1	(	1	)	&	\cellcolor{gray}	1	(	1	)	&	\cellcolor{gray}	1	(	1	)	&	\cellcolor{gray}	1	(	1	)	&	\cellcolor{gray}	1	(	1	)	&	\cellcolor{gray}	1	(	1	)	&	\cellcolor{gray}	1	(	1	)	&	\cellcolor{gray}	1	(	1	)	&	\cellcolor{gray}	1	(	1	)	&	\cellcolor{gray}	1	(	1	)	\\
F2	&	\cellcolor{gray}	1	(	1	)	&	\cellcolor{gray}	1	(	1	)	&	\cellcolor{gray}	1	(	1	)	&	\cellcolor{gray}	1	(	1	)	&	\cellcolor{gray}	1	(	1	)	&	\cellcolor{gray}	1	(	1	)	&	\cellcolor{gray}	1	(	1	)	&	\cellcolor{gray}	1	(	1	)	&	\cellcolor{gray}	1	(	1	)	&	\cellcolor{gray}	1	(	1	)	\\
F3	&	\cellcolor{gray}	1	(	1	)	&	\cellcolor{gray}	1	(	1	)	&	\cellcolor{gray}	1	(	1	)	&	\cellcolor{gray}	1	(	1	)	&	\cellcolor{gray}	1	(	1	)	&	\cellcolor{gray}	1	(	1	)	&	\cellcolor{gray}	1	(	1	)	&	\cellcolor{gray}	1	(	1	)	&	\cellcolor{gray}	1	(	1	)	&	\cellcolor{gray}	1	(	1	)	\\
F4	&	\cellcolor{gray}	1	(	1	)	&	\cellcolor{gray}	1	(	1	)	&	\cellcolor{gray}	1	(	1	)	&	\cellcolor{gray}	1	(	1	)	&	\cellcolor{gray}	1	(	1	)	&	\cellcolor{gray}	1	(	1	)	&	\cellcolor{gray}	1	(	1	)	&	\cellcolor{gray}	1	(	1	)	&	\cellcolor{gray}	1	(	1	)	&	\cellcolor{gray}	1	(	1	)	\\
F5	&	\cellcolor{gray}	1	(	1	)	&	\cellcolor{gray}	1	(	1	)	&	\cellcolor{gray}	1	(	1	)	&	\cellcolor{gray}	1	(	1	)	&	\cellcolor{gray}	1	(	1	)	&	\cellcolor{gray}	1	(	1	)	&	\cellcolor{gray}	1	(	1	)	&	\cellcolor{gray}	1	(	1	)	&	\cellcolor{gray}	1	(	1	)	&	\cellcolor{gray}	1	(	1	)	\\
F6	&	\cellcolor{gray}	1	(	1	)	&	\cellcolor{gray}	1	(	1	)	&	\cellcolor{gray}	1	(	1	)	&		0.96	(	0.36	)	&	\cellcolor{gray}	1	(	1	)	&		0.54	(	0	)	&	\cellcolor{gray}	1	(	1	)	&		1.00	(	0.96	)	&	\cellcolor{gray}	1	(	1	)	&	\cellcolor{gray}	1	(	1	)	\\
F7	&		0.97	(	0.33	)	&		0.94	(	0.20	)	&		0.88	(	0	)	&		0.84	(	0	)	&		0.67	(	0	)	&		0.87	(	0	)	&		0.84	(	0	)	&		0.96	(	0.36	)	&	\cellcolor{gray}	1	(	1	)	&		0.99	(	0.90	)	\\
F8	&	\cellcolor{gray}	1	(	0.90	)	&		0.95	(	0.08	)	&		0.63	(	0	)	&		0.47	(	0	)	&		0.62	(	0	)	&		0	(	0	)	&		0.87	(	0	)	&		0.86	(	0	)	&		0.98	(	0.17	)	&		0.93	(	0.67	)	\\
F9	&		0.51	(	0	)	&		0.51	(	0	)	&		0.34	(	0	)	&		0.43	(	0	)	&		0.32	(	0	)	&		0.47	(	0	)	&		0.55	(	0	)	&		0.73	(	0	)	&		0.81	(	0	)	&	\cellcolor{gray}	0.88	(	0	)	\\
F10	&	\cellcolor{gray}	1	(	1	)	&	\cellcolor{gray}	1	(	1	)	&	\cellcolor{gray}	1	(	1	)	&		1	(	1	)	&	\cellcolor{gray}	1	(	1	)	&		1	(	1	)	&	\cellcolor{gray}	1	(	1	)	&	\cellcolor{gray}	1	(	1	)	&	\cellcolor{gray}	1	(	1	)	&	\cellcolor{gray}	1	(	1	)	\\
F11	&		0.98	(	0.86	)	&	\cellcolor{gray}	1	(	1	)	&	\cellcolor{gray}	1	(	1	)	&		0.67	(	0	)	&	\cellcolor{gray}	1	(	1	)	&		0.67	(	0	)	&	\cellcolor{gray}	1	(	1	)	&	\cellcolor{gray}	1	(	1	)	&	\cellcolor{gray}	1	(	1	)	&	\cellcolor{gray}	1	(	1	)	\\
F12	&		0.91	(	0.53	)	&	\cellcolor{gray}	1	(	1	)	&		0.99	(	0.88	)	&		0.68	(	0	)	&	\cellcolor{gray}	1	(	1	)	&		0.00	(	0	)	&	\cellcolor{gray}	1	(	1	)	&		0.84	(	0.2	)	&		0.92	(	0.33	)	&		0.99	(	0.9	)	\\
F13	&		0.67	(	0	)	&		0.69	(	0	)	&		0.91	(	0.55	)	&		0.67	(	0	)	&		0.95	(	0.72	)	&		0.46	(	0	)	&	\cellcolor{gray}	1	(	1	)	&		0.77	(	0.08	)	&	\cellcolor{gray}	1	(	1	)	&	\cellcolor{gray}	1	(	1	)	\\
F14	&		0.67	(	0	)	&		0.67	(	0	)	&		0.67	(	0	)	&		0.67	(	0	)	&		0.8	(	0	)	&		0.26	(	0	)	&		0.76	(	0	)	&		0.71	(	0	)	&		0.75	(	0	)	&	\cellcolor{gray}	0.83	(	0.16	)	\\
F15	&		0.63	(	0	)	&		0.63	(	0	)	&		0.62	(	0	)	&		0.63	(	0	)	&	\cellcolor{gray}	0.75	(	0	)	&		0.02	(	0	)	&		0.71	(	0	)	&		0.73	(	0	)	&	\cellcolor{gray}	0.75	(	0	)	&		0.68	(	0	)	\\
F16	&	\cellcolor{gray}	0.67	(	0	)	&	\cellcolor{gray}	0.67	(	0	)	&	\cellcolor{gray}	0.67	(	0	)	&		0.57	(	0	)	&	\cellcolor{gray}	0.67	(	0	)	&		0	(	0	)	&	\cellcolor{gray}	0.67	(	0	)	&	\cellcolor{gray}	0.67	(	0	)	&	\cellcolor{gray}	0.67	(	0	)	&	\cellcolor{gray}	0.67	(	0	)	\\
F17	&		0.48	(	0	)	&		0.40	(	0	)	&		0.38	(	0	)	&		0.42	(	0	)	&		0.41	(	0	)	&		0	(	0	)	&		0.68	(	0	)	&	\cellcolor{gray}	0.68	(	0	)	&		0.67	(	0	)	&		0.63	(	0	)	\\
F18	&	\cellcolor{gray}	0.67	(	0	)	&		0.65	(	0	)	&	\cellcolor{gray}	0.67	(	0	)	&	\cellcolor{gray}	0.67	(	0	)	&		0.5	(	0	)	&		0.15	(	0	)	&	\cellcolor{gray}	0.67	(	0	)	&		0.66	(	0	)	&	\cellcolor{gray}	0.67	(	0	)	&	\cellcolor{gray}	0.67	(	0	)	\\
F19	&		0.46	(	0	)	&		0.36	(	0	)	&		0.40	(	0	)	&		0.42	(	0	)	&		0.25	(	0	)	&		0	(	0	)	&	\cellcolor{gray}	0.61	(	0	)	&		0.44	(	0	)	&		0.5	(	0	)	&		0.42	(	0	)	\\
F20	&		0.38	(	0	)	&		0.25	(	0	)	&		0.31	(	0	)	&		0.25	(	0	)	&		0.24	(	0	)	&		0	(	0	)	&	\cellcolor{gray}	0.49	(	0	)	&		0.10	(	0	)	&		0.44	(	0	)	&		0.35	(	0	)	\\\hline
$AVR$	&		0.80	(	0.48	)	&		0.79	(	0.46	)	&		0.77	(	0.47	)	&		0.72	(	0.32	)	&		0.76	(	0.49	)	&		0.47	(	0.30	)	&		0.84	(	0.50	)	&		0.81	(	0.43	)	&	\cellcolor{gray}	0.86	(	0.53	)	&		0.85	(	0.58	)	\\\hline
$+/\approx/-$	&		$8/9/3$				&		$9/9/2$				&		$9/11/0$				&		$13/7/0$				&		$9/9/2$				&		$14/6/0$				&		$4/11/5$				&		$9/8/3$				&		$3/11/6$				&						\\\hline
\end{tabular}} 
\end{table}

\section{Open Issues}\label{sec: open questions}
In the preceding sections, we reviewed DE variants developed to address various MMOPs, including MM-MOOPs. We categorized state-of-the-art approaches into niching-based, clustering-based, mutation and parameter adaptation-based, hybrid, machine learning-based, and multi-objective DE techniques. Additionally, we explored multi-level and specialized niching methods, along with real-world applications in engineering, healthcare, and data-driven tasks. Despite these advancements, several open questions remain regarding DE modifications, application domains, and multimodal methods (MM). While some of these future directions were highlighted in previous reviews, they continue to be active areas for research.

\begin{table}[]
    \centering
        \caption{Experimental results of various DE variants on CEC 2015 MMOPs at $\epsilon=1E-05$ accuracy level.}\label{tab: results PR(SR) for CEC 2015}
\resizebox{1\linewidth}{!}{    \begin{tabular}{ccccccccccc}\hline		
Func	&		PNPCDE				&		LMCEDA				&		LMSEDA				&		LBPADE				&		FBK-DE				&		PMODE				&		NCD-DE				&		ANDE				&		NMMSO				&		WSNADE				\\\hline
	&		PR	(	SR	)	&		PR	(	SR	)	&		PR	(	SR	)	&		PR	(	SR	)	&		PR	(	SR	)	&		PR	(	SR	)	&		PR	(	SR	)	&		PR	(	SR	)	&		PR	(	SR	)	&		PR	(	SR	)	\\\hline
F1	&	\cellcolor{gray}	1	(	1	)	&	\cellcolor{gray}	1	(	1	)	&	\cellcolor{gray}	1	(	1	)	&	\cellcolor{gray}	1	(	1	)	&	\cellcolor{gray}	1	(	1	)	&	\cellcolor{gray}	1	(	1	)	&	\cellcolor{gray}	1	(	1	)	&	\cellcolor{gray}	1	(	1	)	&	\cellcolor{gray}	1	(	1	)	&	\cellcolor{gray}	1	(	1	)	\\
F2	&	\cellcolor{gray}	1	(	1	)	&	\cellcolor{gray}	1	(	1	)	&	\cellcolor{gray}	1	(	1	)	&	\cellcolor{gray}	1	(	1	)	&	\cellcolor{gray}	1	(	1	)	&	\cellcolor{gray}	1	(	1	)	&	\cellcolor{gray}	1	(	1	)	&	\cellcolor{gray}	1	(	1	)	&	\cellcolor{gray}	1	(	1	)	&	\cellcolor{gray}	1	(	1	)	\\
F3	&	\cellcolor{gray}	1	(	1	)	&	\cellcolor{gray}	1	(	1	)	&	\cellcolor{gray}	1	(	1	)	&	\cellcolor{gray}	1	(	1	)	&	\cellcolor{gray}	1	(	1	)	&	\cellcolor{gray}	1	(	1	)	&	\cellcolor{gray}	1	(	1	)	&	\cellcolor{gray}	1	(	1	)	&	\cellcolor{gray}	1	(	1	)	&	\cellcolor{gray}	1	(	1	)	\\
F4	&		0.32	(	0	)	&	\cellcolor{gray}	1	(	1	)	&	\cellcolor{gray}	1	(	1	)	&	\cellcolor{gray}	1	(	1	)	&		\cellcolor{gray} 1	(	1	)	&	\cellcolor{gray}	1	(	1	)	&	\cellcolor{gray}	1	(	1	)	&	\cellcolor{gray}	1	(	1	)	&	\cellcolor{gray}	1	(	1	)	&	\cellcolor{gray}	1	(	1	)	\\
F5	&	\cellcolor{gray}	1	(	1	)	&	\cellcolor{gray}	1	(	1	)	&	\cellcolor{gray}	1	(	1	)	&	\cellcolor{gray}	1	(	1	)	&	\cellcolor{gray}	1	(	1	)	&	\cellcolor{gray}	1	(	1	)	&	\cellcolor{gray}	1	(	1	)	&	\cellcolor{gray}	1	(	1	)	&	\cellcolor{gray}	1	(	1	)	&	\cellcolor{gray}	1	(	1	)	\\
F6	&		0.93	(	0.55	)	&		0.97	(	0.57	)	&		0.92	(	0.18	)	&		0.96	(	0.41	)	&		0.80	(	0.08	)	&		1.00	(	0.96	)	&	\cellcolor{gray}	1	(	1	)	&	\cellcolor{gray}	1	(	1	)	&		0	(	0	)	&	\cellcolor{gray}	1	(	1	)	\\
F7	&		0.84	(	0.02	)	&		0.70	(	0	)	&		0.68	(	0	)	&		0.84	(	0	)	&		0.67	(	0	)	&		0.65	(	0	)	&		0.83	(	0	)	&		0.94	(	0.20	)	&	\cellcolor{gray}	1	(	1	)	&		0.86	(	0	)	\\
F8	&		0.22	(	0	)	&		0.25	(	0	)	&		0.28	(	0	)	&		0.36	(	0	)	&		0.32	(	0	)	&		0.64	(	0	)	&		0.82	(	0	)	&		0.95	(	0.04	)	&		0.87	(	0	)	&	\cellcolor{gray}	0.96	(	0	)	\\
F9	&		0.54	(	0	)	&		0.25	(	0	)	&		0.22	(	0	)	&		0.43	(	0	)	&		0.29	(	0	)	&		0.32	(	0	)	&		0.41	(	0	)	&		0.51	(	0	)	&	\cellcolor{gray}	0.98	(	0.12	)	&		0.49	(	0	)	\\
F10	&	\cellcolor{gray}	1	(	1	)	&	\cellcolor{gray}	1	(	1	)	&	\cellcolor{gray}	1	(	1	)	&	\cellcolor{gray}	1	(	1	)	&	\cellcolor{gray}	1	(	1	)	&	\cellcolor{gray}	1	(	1	)	&	\cellcolor{gray}	1	(	1	)	&	\cellcolor{gray}	1	(	1	)	&	\cellcolor{gray}	1	(	1	)	&	\cellcolor{gray}	1	(	1	)	\\
F11	&		0.42	(	0	)	&		0.67	(	0	)	&		0.67	(	0	)	&		0.67	(	0	)	&		0.99	(	0.94	)	&	\cellcolor{gray}	1	(	1	)	&	\cellcolor{gray}	1	(	1	)	&	\cellcolor{gray}	1	(	1	)	&		0.99	(	0.94	)	&	\cellcolor{gray}	1	(	1	)	\\
F12	&		0.02	(	0	)	&		0.75	(	0	)	&		0.77	(	0.04	)	&		0.70	(	0	)	&		0.91	(	0.31	)	&	\cellcolor{gray}	1	(	1	)	&		0.91	(	0.45	)	&	\cellcolor{gray}	1	(	1	)	&		0.99	(	0.92	)	&		0.96	(	0.57	)	\\
F13	&		0.24	(	0	)	&		0.67	(	0	)	&		0.67	(	0	)	&		0.67	(	0	)	&	\cellcolor{gray}	0.99	(	0.94	)	&		0.91	(	0.61	)	&		0.86	(	0.27	)	&		0.69	(	0	)	&		0.98	(	0.90	)	&		0.96	(	0.73	)	\\
F14	&		0.25	(	0	)	&		0.67	(	0	)	&		0.67	(	0	)	&		0.67	(	0	)	&	\cellcolor{gray}	0.78	(	0.10	)	&		0.79	(	0.04	)	&		0.67	(	0	)	&		0.67	(	0	)	&		0.72	(	0	)	&		0.73	(	0	)	\\
F15	&		0.13	(	0	)	&		0.68	(	0	)	&		0.60	(	0	)	&		0.64	(	0	)	&		0.70	(	0	)	&		0.74	(	0	)	&		0.64	(	0	)	&		0.63	(	0	)	&		0.63	(	0	)	&	\cellcolor{gray}	0.74	(	0	)	\\
F16	&		0.15	(	0	)	&		0.67	(	0	)	&		0.67	(	0	)	&		0.67	(	0	)	&	\cellcolor{gray}	0.68	(	0	)	&		0.67	(	0	)	&		0.67	(	0	)	&		0.67	(	0	)	&		0.66	(	0	)	&		0.67	(	0	)	\\
F17	&		0.09	(	0	)	&		0.45	(	0	)	&		0.48	(	0	)	&		0.48	(	0	)	&		0.66	(	0	)	&		0.42	(	0	)	&		0.48	(	0	)	&		0.40	(	0	)	&		0.46	(	0	)	&	\cellcolor{gray}	0.66	(	0	)	\\
F18	&		0.17	(	0	)	&		0.65	(	0	)	&		0.66	(	0	)	&		0.66	(	0	)	&	\cellcolor{gray}	0.67	(	0	)	&		0.53	(	0	)	&		0.66	(	0	)	&		0.65	(	0	)	&		0.65	(	0	)	&	\cellcolor{gray}	0.67	(	0	)	\\
F19	&		0	(	0	)	&		0.45	(	0	)	&		0.49	(	0	)	&		0.32	(	0	)	&		0.45	(	0	)	&		0.28	(	0	)	&		0.49	(	0	)	&		0.36	(	0	)	&		0.44	(	0	)	&	\cellcolor{gray}	0.50	(	0	)	\\
F20	&		0.12	(	0	)	&		0.24	(	0	)	&		0.23	(	0	)	&		0.18	(	0	)	&	\cellcolor{gray}	0.33	(	0	)	&		0.25	(	0	)	&		0.25	(	0	)	&		0.25	(	0	)	&		0.17	(	0	)	&		0.27	(	0	)	\\\hline
$AVR$	&		0.47	(	0.28	)	&		0.70	(	0.33	)	&		0.70	(	0.31	)	&		0.71	(	0.32	)	&		0.76	(	0.42	)	&		0.76	(	0.48	)	&		0.78	(	0.44	)	&		0.79	(	0.46	)	&		0.78	(	0.49	)	&	\cellcolor{gray}	0.82	(	0.47	)	\\\hline
$+/\approx/-$	&		$14/5/1$				&		$13/7/0$				&		$13/7/0$				&		$13/7/0$				&		$9/7/4$				&		$9/7/4$				&		$9/8/3$				&		$8/9/3$				&		$10/6/4$				&						\\\hline
\end{tabular}} 
\end{table}

\begin{figure}
\begin{subfigure}[b]{1\linewidth}
\includegraphics[width=0.5\linewidth]{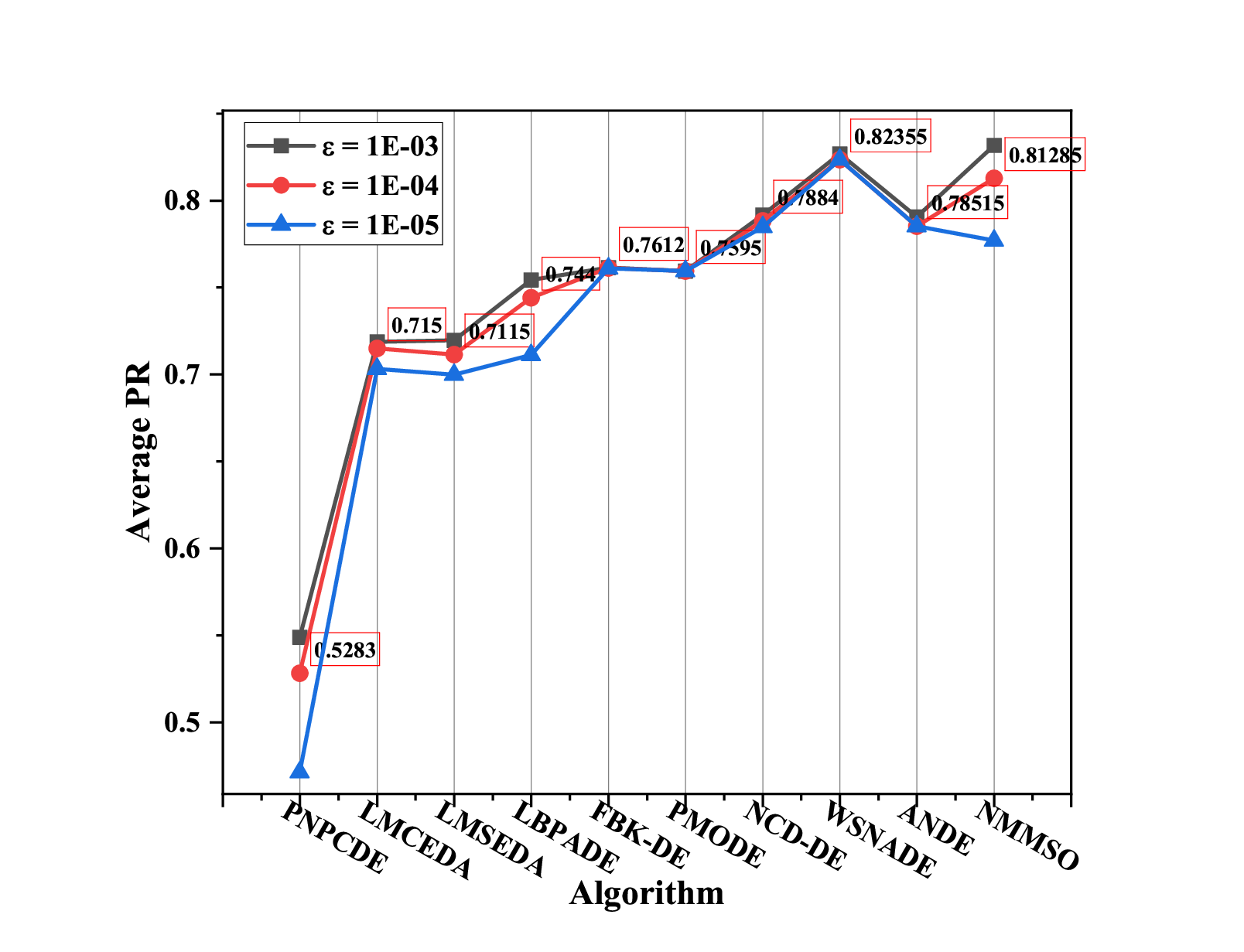}
\includegraphics[width=0.5\linewidth]{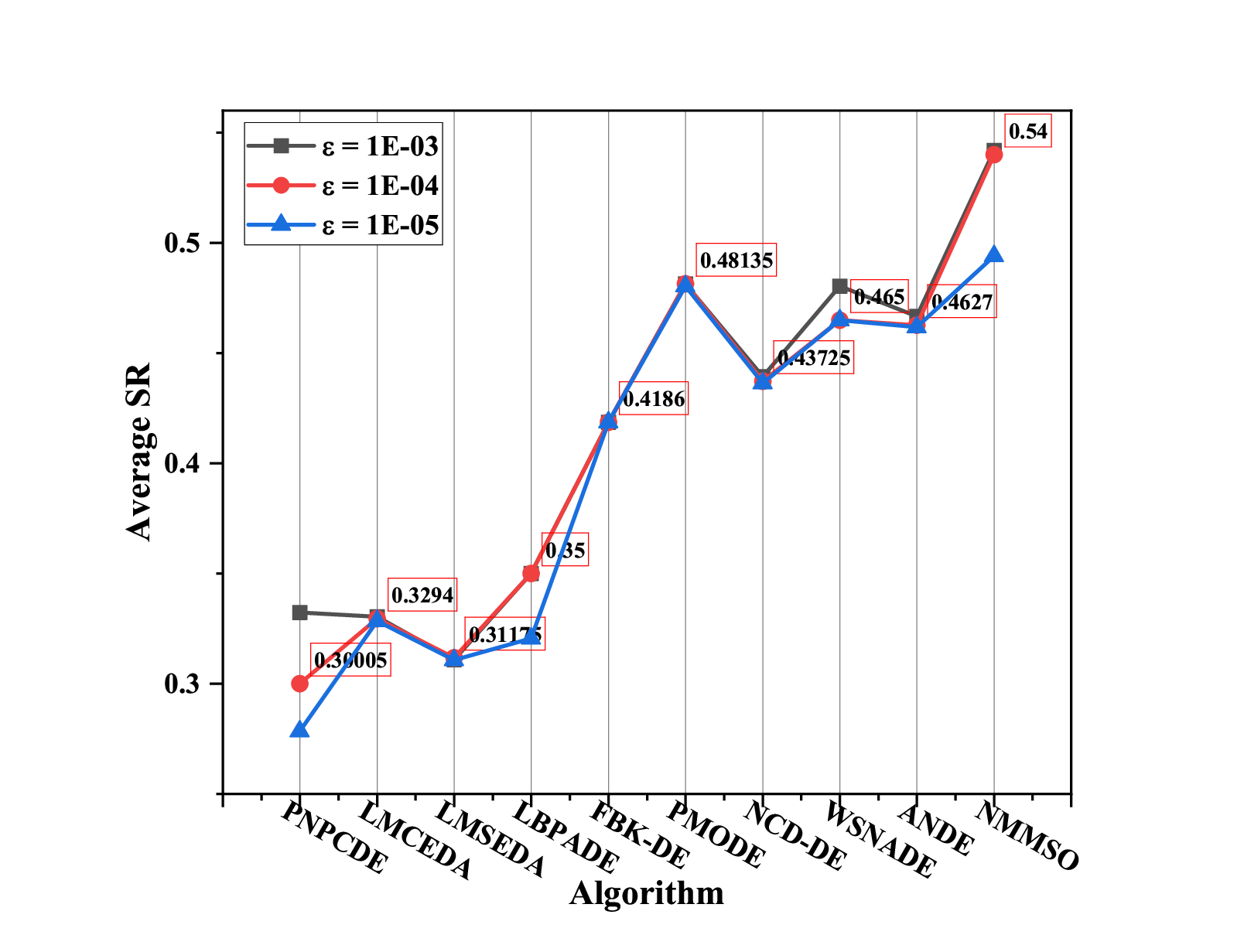}
\end{subfigure}
\caption{PR and SR values of different DE algorithms at different accuracy levels for CEC 2015 MMOP.}\label{Fig: results the CEC 2015}
\end{figure}
\subsection{DE Perspective}
In this subsection, we provide some open questions from the perspective of DE modification to solve MMOPs. In the context of DE-based algorithms, there are several open questions and directions for future research. First, as DE-based algorithms often involve multiple parameters, the introduction of niching methods typically increases the number of parameters, making it essential to explore the design of parameter-free DE algorithms to ensure consistent performance across a range of tasks~\cite{gong2017learning, wang2020automatic}. This could involve the development of strategies similar to those in other algorithms, such as PSO~\cite{li2009niching}. Additionally, there is a lack of theoretical understanding regarding the distributed convergence behavior of DE-based niching techniques~\cite{das2011real}. A key question is how to analyze the impact of niching parameters on convergence and establish guidelines for their optimal selection across different problem domains. Another challenge is the limited application of DE-based algorithms to low-dimensional MM spaces, and further research is needed to assess their effectiveness in high-dimensional MMOPs, exploring their potential limitations in such landscapes.

Furthermore, DE methods must be adapted to handle dynamic and noisy conditions, which will be crucial for their real-world applicability~\cite{li2016seeking}. Another important question is how niching techniques in DE can be used to address constraints in MMOPs, especially in combinatorial and real-world restricted scenarios. In addition, existing evaluation metrics for DE-based algorithms often rely on assumptions that may not hold in practice~\cite{li2016seeking}, thus refining these metrics is crucial to enhance performance assessment, particularly when the number and location of optima are unknown. Furthermore, there is a need to explore how DE can be adapted to identify a subset of solutions when the number of optima is unknown or excessively large. Hybrid DE approaches combining multi-level and multi-niche techniques offer another promising direction to improve the discovery of multiple solutions across various regions of the fitness landscape~\cite{huang2024wireless, wang2019multilevel}. Lastly, there are relatively few DE hybridization methods involving other EAs and ML-based approaches, and future research could explore how hybridizing DE with other techniques could enhance its effectiveness in solving high-dimensional, expensive MMOPs.

\subsection{Application Perspective}
In this subsection, we outline several future directions from the perspective of DE applications. These open questions highlight areas where further research is needed to advance DE-based methods and their applicability to real-world challenges. (i) How can we improve the formulation of real-world problems, especially in clustering and multi-modal optimization tasks, to better capture the complexities inherent in these applications? For instance, could modeling a typical clustering problem as an MMOP enhance both the quality and applicability of the solutions? (ii) How could niching methods be effectively applied to ensure a diverse set of learners in ensemble learning models~\cite{awad2018ensemble}, particularly in nonstationary environments? This may include exploring how niching can address uncertainties across various learning systems.
(iii) What approaches could be adopted to benchmark DE-based niching methods more effectively in real-world applications such as engineering design, healthcare, and feature selection tasks? These domains often present challenges like unknown optima, complex fitness landscapes, and evolving objectives. (iv) How might multi-objective DE techniques be adapted to better handle real-world scenarios, where competing objectives, constraints, and trade-offs are common? This is an area that remains a significant challenge for practical implementation.
(v) How can DE-based niching methods be improved to scale efficiently for large-scale real-world systems and datasets, where the complexity of the search space tends to grow substantially?

\subsection{MM Perspective}
In this subsection, we highlight potential future directions from the perspective of MM optimization. These open questions aim to address key challenges in enhancing DE-based algorithms for solving complex MMOPs and improving their adaptability to various problem landscapes. (i) How can we more effectively model and solve MMOPs using DE, particularly considering the complexity of fitness landscapes with multiple basins of attraction? Identifying and characterizing these basins remains a significant challenge in multimodal optimization.
(ii) How can we extend niching methods to address dynamic MMOPs, where the optima evolve over time and the optimization algorithm must continually adapt to track new solutions~\cite{liu2023learning}?
(iii) Given the need for further theoretical exploration, how can we deepen our understanding of the convergence properties of niching methods in MMOPs, particularly regarding their ability to locate and maintain multiple optima?
(iv) Niching-based algorithms have successfully solved MMOPs but are often sensitive to parameter settings and cannot dynamically adjust the number of niches~\cite{liang2024niche,luo2021survey}. Future research could focus on developing methods to adapt the number of niches during optimization, offering greater flexibility and robustness for different MMOPs.
(v) In NBC, the population is divided by sorting individuals by fitness and linking each to its nearest higher fitness neighbor, forming a spanning tree. However, this division is heavily influenced by the problem landscape, limiting NBC's applicability in diverse scenarios~\cite{lin2019differential,li2023minimum}. Future research should explore how landscape variations affect population division and how NBC can be adapted to improve performance across different MMOPs.
\section{Future directions}
Future research in DE for high-dimensional MMOPs will have focus on improving scalability, efficiency, and robustness. Key areas include developing adaptive mechanisms for parameter tuning, enhancing exploration-exploitation balance through dynamic niching and hybrid approaches, and leveraging dimensionality reduction techniques to handle complex search spaces. Novel mutation and crossover strategies, along with improved diversity maintenance and clustering methods, will be explored to better identify and preserve multiple optima. Additionally, integrating DE with machine learning, such as surrogate models and automated algorithm configuration, will enhance performance in real-world applications. For instance, in futuristic smart city planning, DE could optimize the placement of renewable energy sources, transportation networks, and waste management systems across a high-dimensional landscape, ensuring multiple sustainable solutions are identified. Theoretical studies on convergence and parameter sensitivity, along with extensions to dynamic, multi-objective, and noisy environments, will further advance DE's applicability to high-dimensional MMOPs. Here are some anticipated of future research trends in this area.

\subsection{Adaptation to dynamic and noisy environments}
In dynamic environments, such as optimizing resource allocation in a futuristic smart grid, DE algorithms must adapt to fluctuating energy demands, weather conditions, and renewable energy availability. This requires developing adaptive mechanisms for real-time parameter tuning and dynamic niching techniques to maintain population diversity and track shifting optima. In noisy environments, such as optimizing sensor placement for environmental monitoring in urban areas, DE must handle uncertain or imprecise fitness evaluations. Techniques like surrogate modeling, robust fitness evaluation methods, and noise-resistant mutation strategies will be critical. Additionally, integrating machine learning for predictive modeling and automated adaptation will enhance DE's ability to navigate complex, high-dimensional landscapes under uncertainty. These advancements will ensure DE remains effective in solving MMOPs in dynamic and noisy real-world scenarios.
\subsection{Better evaluation metrics}
Future evaluation metrics for DE will focus on capturing not only solution quality but also adaptability, robustness, and efficiency in increasingly complex and dynamic environments. As industries like smart cities, renewable energy, and autonomous systems may rely on DE for complex decision-making, more robust metrics will be required to prioritize adaptability and robustness. Machine learning-based performance evaluation metrics are the need of the hour. Some of the early models like IOH-Xplainer~\cite{van2024explainable} are some early works in this direction. 
\subsection{Theoretical analysis}
Despite the widespread success of Differential Evolution (DE) in optimization, a deeper theoretical understanding is still needed to enhance its reliability and efficiency. Future research should focus on rigorous convergence analysis, mathematical modeling of exploration-exploitation dynamics, and adaptive parameter control with provable guarantees. Additionally, insights into DE's behavior in high-dimensional, noisy, and constrained environments will be crucial for its scalability. The integration of DE with quantum computing, bio-inspired mechanisms, and hybrid optimization frameworks also demands strong theoretical foundations. Addressing these gaps will not only improve DE's applicability but also establish it as a more robust and interpretable optimization method.
\subsection{Surrogate-assisted DE}
The future of Surrogate-Assisted DE (SADE) lies in enhancing adaptability, scalability, and efficiency for solving complex optimization problems. Research is focusing on dynamic surrogate selection, hybrid models, and real-time learning to improve accuracy and computational speed. Advances in parallel computing, GPU acceleration, and cloud-based implementations will further expand its capabilities. Additionally, integrating SADE with deep learning and explainable AI will make it more interpretable and applicable to real-world challenges in engineering, machine learning, and robotics. As computational demands grow, SADE has the potential to become a key tool for optimizing high-dimensional and expensive problems.
\subsection{Quantum-inspired DE}
Quantum-Inspired Differential Evolution (QIDE) is an emerging optimization algorithm that integrates principles from quantum computing, such as superposition and entanglement, into the classical Differential Evolution framework. Currently, QIDE is in the experimental phase, showing promise in enhancing exploration and avoiding local optima in complex optimization problems, though its performance is problem-dependent and limited by classical simulation constraints. Future advancements in quantum hardware and hybrid quantum-classical algorithms could unlock its full potential, enabling applications in areas like logistics, drug discovery, and machine learning. As research progresses, QIDE may bridge the gap between classical and quantum optimization, offering powerful solutions for high-dimensional and multimodal problems.
\subsection{Explainable DE}
Explainability of DE refers to the enhancement of the traditional DE algorithm to make its operations, decisions, and outcomes more transparent and understandable to users. This involves incorporating mechanisms that allow users to trace how solutions are generated, why certain solutions are preferred over others, and how the algorithm converges to the final solution. This is one of the potential areas where possible research could enhance the reliability of DE in real-life situations. 

\subsection{DE and LLM}
The future of DE and large language models (LLMs) holds immense potential, driven by advancements in optimization and AI. DE, a robust evolutionary algorithm, could evolve through hybridization with machine learning, adaptive mechanisms, and quantum computing, enabling efficient solutions for complex, real-world problems. Meanwhile, LLMs are expected to become more efficient, multimodal, and personalized, with enhanced ethical frameworks and explainability. The synergy between DE and LLMs could revolutionize fields like scientific discovery, autonomous systems, and decision-making, as DE optimizes processes while LLMs provide reasoning and natural language understanding. Together, they promise to push the boundaries of AI and optimization, though challenges like computational costs and ethical considerations must be addressed to fully realize their potential.

\section{Conclusion}\label{sec: conclusion}
In this article, we have systematically examined various DE algorithms designed to address the challenges of MMOPs, including multimodal multi-objective optimization problems (MM-MOOPs). We have categorized the proposed approaches based on their distinct strategies, including speciation-based methods, clustering techniques, dynamic population adaptations, and hybrid models. Furthermore, we have discussed the integration of advanced strategies such as adaptive mutation, surrogate models, and reinforcement learning, which enhance the ability of DE algorithms to locate multiple optima while maintaining diversity. A key theme throughout the review is balancing exploration and exploitation in solving MM-MOOPs. Methods that use niching techniques, crowding distance metrics, and adaptive strategies show significant promise in maintaining population diversity and preventing premature convergence. Additionally, the ability of DE to handle high-dimensional search spaces and multiple conflicting objectives demonstrates its versatility in real-world applications across fields such as engineering design, healthcare, and data-driven tasks. Despite the advancements in DE for MM-MOOPs, several challenges remain, including the effective handling of large-scale problems, the scalability of algorithms in high-dimensional spaces, and the need for robust and adaptive parameter tuning mechanisms. % Future work in this area could explore the integration of machine learning-based techniques to further automate and refine the optimization process, as well as the development of hybrid algorithms that combine the strengths of DE with other evolutionary or nature-inspired methods. 

%In conclusion, DE-based approaches continue to be a powerful tool for solving multimodal multi-objective optimization problems, and ongoing innovations promise to enhance their efficiency, diversity-maintenance capabilities, and applicability to increasingly complex real-world problems.

 \section*{Acknowledgment}
\noindent This work was supported by the National Research Foundation of Korea (NRF) grant funded by the Korea government (MSIT) (No. NRF-RS-2021-NR060085). The authors are also thankful to Dr. B. R. Ambedkar National Institute of Technology Jalandhar, India, and the National University of Singapore for providing necessary support to this research.

	\section*{Compliance with ethical standards}
	\noindent \textbf{Conflict of interest:} All the authors declare that they have no conflict of interest.
	
	\noindent \textbf{Ethical approval:} This article does not contain any studies with human participants or animals performed by any of the authors.
	%	%\begin{landscape}
	%	\clearpage
	\bibliographystyle{unsrt}
	%	%%\bibliographystyle{model5-names}\biboptions{authoryear}
	\small
	\bibliography{niching_EAs}

\begin{thebibliography}{100}

\bibitem{bandaru2017data}
Sunith Bandaru, Amos~HC Ng, and Kalyanmoy Deb.
\newblock Data mining methods for knowledge discovery in multi-objective optimization: Part a-survey.
\newblock {\em Expert Systems with Applications}, 70:139--159, 2017.

\bibitem{del2019bio}
Javier Del~Ser, Eneko Osaba, Daniel Molina, Xin-She Yang, Sancho Salcedo-Sanz, David Camacho, Swagatam Das, Ponnuthurai~N Suganthan, Carlos A~Coello Coello, and Francisco Herrera.
\newblock Bio-inspired computation: Where we stand and what's next.
\newblock {\em Swarm and Evolutionary Computation}, 48:220--250, 2019.

\bibitem{zhan2022survey}
Zhi-Hui Zhan, Lin Shi, Kay~Chen Tan, and Jun Zhang.
\newblock A survey on evolutionary computation for complex continuous optimization.
\newblock {\em Artificial Intelligence Review}, 55(1):59--110, 2022.

\bibitem{das2010differential}
Swagatam Das and Ponnuthurai~Nagaratnam Suganthan.
\newblock Differential evolution: A survey of the state-of-the-art.
\newblock {\em IEEE Transactions on Evolutionary Computation}, 15(1):4--31, 2010.

\bibitem{nekouie2016new}
Nadia Nekouie and Mahdi Yaghoobi.
\newblock A new method in multimodal optimization based on firefly algorithm.
\newblock {\em Artificial Intelligence Review}, 46:267--287, 2016.

\bibitem{li2013benchmark}
Xiaodong Li, Andries Engelbrecht, and Michael~G Epitropakis.
\newblock Benchmark functions for cec’2013 special session and competition on niching methods for multimodal function optimization.
\newblock {\em RMIT University, Evolutionary Computation and Machine Learning Group, Australia, Tech. Rep}, 2013.

\bibitem{ghorbanali2023comprehensive}
Alireza Ghorbanali and Mohammad~Karim Sohrabi.
\newblock A comprehensive survey on deep learning-based approaches for multimodal sentiment analysis.
\newblock {\em Artificial Intelligence Review}, 56(Suppl 1):1479--1512, 2023.

\bibitem{li2016seeking}
Xiaodong Li, Michael~G Epitropakis, Kalyanmoy Deb, and Andries Engelbrecht.
\newblock Seeking multiple solutions: An updated survey on niching methods and their applications.
\newblock {\em IEEE Transactions on Evolutionary Computation}, 21(4):518--538, 2016.

\bibitem{oskouie2014multimodal}
Payam Oskouie, Sara Alipour, and Amir-Masoud Eftekhari-Moghadam.
\newblock Multimodal feature extraction and fusion for semantic mining of soccer video: a survey.
\newblock {\em Artificial Intelligence Review}, 42(2):173--210, 2014.

\bibitem{lu2022evolving}
Zhenyu Lu, Shaoyang Liang, Qiang Yang, and Bo~Du.
\newblock Evolving block-based convolutional neural network for hyperspectral image classification.
\newblock {\em IEEE Transactions on Geoscience and Remote Sensing}, 60:1--21, 2022.

\bibitem{yang2021probabilistic}
Qiang Yang, Wei-Neng Chen, and Jun Zhang.
\newblock Probabilistic multimodal optimization.
\newblock {\em Metaheuristics for Finding Multiple Solutions}, pages 191--228, 2021.

\bibitem{yoo2015novel}
Chung-Hee Yoo, Dong-Kuk Lim, and Hyun-Kyo Jung.
\newblock A novel multimodal optimization algorithm for the design of electromagnetic machines.
\newblock {\em IEEE Transactions on Magnetics}, 52(3):1--4, 2015.

\bibitem{yoo2017new}
Chung-Hee Yoo.
\newblock A new multi-modal optimization approach and its application to the design of electric machines.
\newblock {\em IEEE Transactions on Magnetics}, 54(3):1--4, 2017.

\bibitem{huang2019niching}
Ting Huang, Yue-Jiao Gong, Sam Kwong, Hua Wang, and Jun Zhang.
\newblock A niching memetic algorithm for multi-solution traveling salesman problem.
\newblock {\em IEEE Transactions on Evolutionary Computation}, 24(3):508--522, 2019.

\bibitem{wong2010protein}
Ka-Chun Wong, Kwong-Sak Leung, and Man-Hon Wong.
\newblock Protein structure prediction on a lattice model via multimodal optimization techniques.
\newblock In {\em Proceedings of the 12th annual conference on Genetic and evolutionary computation}, pages 155--162, 2010.

\bibitem{ji2021dual}
Xinfang Ji, Yong Zhang, Dunwei Gong, and Xiaoyan Sun.
\newblock Dual-surrogate-assisted cooperative particle swarm optimization for expensive multimodal problems.
\newblock {\em IEEE Transactions on Evolutionary Computation}, 25(4):794--808, 2021.

\bibitem{hong2017simulation}
Jeong~Hee Hong and Kwang~Ryel Ryu.
\newblock Simulation-based multimodal optimization of decoy system design using an archived noise-tolerant genetic algorithm.
\newblock {\em Engineering Applications of Artificial Intelligence}, 65:230--239, 2017.

\bibitem{liu2018multimodal}
Yiping Liu, Gary~G Yen, and Dunwei Gong.
\newblock A multimodal multiobjective evolutionary algorithm using two-archive and recombination strategies.
\newblock {\em IEEE Transactions on Evolutionary Computation}, 23(4):660--674, 2018.

\bibitem{jaszkiewicz2002performance}
Andrzej Jaszkiewicz.
\newblock On the performance of multiple-objective genetic local search on the 0/1 knapsack problem-a comparative experiment.
\newblock {\em IEEE Transactions on Evolutionary Computation}, 6(4):402--412, 2002.

\bibitem{rahbar2022architectural}
Morteza Rahbar, Mohammadjavad Mahdavinejad, Amir~HD Markazi, and Mohammadreza Bemanian.
\newblock Architectural layout design through deep learning and agent-based modeling: A hybrid approach.
\newblock {\em Journal of Building Engineering}, 47:103822, 2022.

\bibitem{han2017evolutionary}
Yuyan Han, Dunwei Gong, Yaochu Jin, and Quanke Pan.
\newblock Evolutionary multiobjective blocking lot-streaming flow shop scheduling with machine breakdowns.
\newblock {\em IEEE Transactions on Cybernetics}, 49(1):184--197, 2017.

\bibitem{wang2014mommop}
Yong Wang, Han-Xiong Li, Gary~G Yen, and Wu~Song.
\newblock Mommop: Multiobjective optimization for locating multiple optimal solutions of multimodal optimization problems.
\newblock {\em IEEE Transactions on Cybernetics}, 45(4):830--843, 2014.

\bibitem{akopov2019parallel}
Andranik~S Akopov, Levon~A Beklaryan, Manoj Thakur, and Bhisham~Dev Verma.
\newblock Parallel multi-agent real-coded genetic algorithm for large-scale black-box single-objective optimisation.
\newblock {\em Knowledge-Based Systems}, 174:103--122, 2019.

\bibitem{houssein2021major}
Essam~H Houssein, Ahmed~G Gad, Kashif Hussain, and Ponnuthurai~Nagaratnam Suganthan.
\newblock Major advances in particle swarm optimization: theory, analysis, and application.
\newblock {\em Swarm and Evolutionary Computation}, 63:100868, 2021.

\bibitem{zhang2020modified}
XuWei Zhang, Hao Liu, and LiangPing Tu.
\newblock A modified particle swarm optimization for multimodal multi-objective optimization.
\newblock {\em Engineering Applications of Artificial Intelligence}, 95:103905, 2020.

\bibitem{price2006differential}
K~Price.
\newblock {\em Differential Evolution: a Practical Approach to Global Optimization}.
\newblock Springer Science \& Business Media, 2006.

\bibitem{das2016recent}
Swagatam Das, Sankha~Subhra Mullick, and Ponnuthurai~N Suganthan.
\newblock Recent advances in differential evolution--an updated survey.
\newblock {\em Swarm and evolutionary computation}, 27:1--30, 2016.

\bibitem{zhang2011evolutionary}
Jun Zhang, Zhi-hui Zhan, Ying Lin, Ni~Chen, Yue-jiao Gong, Jing-hui Zhong, Henry~SH Chung, Yun Li, and Yu-hui Shi.
\newblock Evolutionary computation meets machine learning: A survey.
\newblock {\em IEEE Computational Intelligence Magazine}, 6(4):68--75, 2011.

\bibitem{talbi2009metaheuristics}
EG~Talbi.
\newblock Metaheuristics: From design to implementation.
\newblock {\em John Wiley \& Sons google schola}, 2:268--308, 2009.

\bibitem{qu2012differential}
B.~Y. Qu, P.~N. Suganthan, and J.~J. Liang.
\newblock Differential evolution with neighborhood mutation for multimodal optimization.
\newblock {\em IEEE Transactions on Evolutionary Computation}, 16(5):601--614, 2012.

\bibitem{li2009niching}
Xiaodong Li.
\newblock Niching without niching parameters: particle swarm optimization using a ring topology.
\newblock {\em IEEE Transactions on Evolutionary Computation}, 14(1):150--169, 2009.

\bibitem{yang2016multimodal}
Qiang Yang, Wei-Neng Chen, Yun Li, CL~Philip Chen, Xiang-Min Xu, and Jun Zhang.
\newblock Multimodal estimation of distribution algorithms.
\newblock {\em IEEE Transactions on Cybernetics}, 47(3):636--650, 2016.

\bibitem{wu2016differential}
Guohua Wu, Rammohan Mallipeddi, Ponnuthurai~Nagaratnam Suganthan, Rui Wang, and Huangke Chen.
\newblock Differential evolution with multi-population based ensemble of mutation strategies.
\newblock {\em Information Sciences}, 329:329--345, 2016.

\bibitem{de1975analysis}
Kenneth~Alan De~Jong.
\newblock {\em An analysis of the behavior of a class of genetic adaptive systems.}
\newblock University of Michigan, 1975.

\bibitem{holland1992adaptation}
John~H Holland.
\newblock {\em Adaptation in natural and artificial systems: an introductory analysis with applications to biology, control, and artificial intelligence}.
\newblock MIT press, 1992.

\bibitem{stoean2010multimodal}
Catalin Stoean, Mike Preuss, Ruxandra Stoean, and Dumitru Dumitrescu.
\newblock Multimodal optimization by means of a topological species conservation algorithm.
\newblock {\em IEEE Transactions on Evolutionary Computation}, 14(6):842--864, 2010.

\bibitem{gao2013cluster}
Weifeng Gao, Gary~G Yen, and Sanyang Liu.
\newblock A cluster-based differential evolution with self-adaptive strategy for multimodal optimization.
\newblock {\em IEEE Transactions on Cybernetics}, 44(8):1314--1327, 2013.

\bibitem{lacroix2016region}
Benjamin Lacroix, Daniel Molina, and Francisco Herrera.
\newblock Region-based memetic algorithm with archive for multimodal optimisation.
\newblock {\em Information Sciences}, 367:719--746, 2016.

\bibitem{ren2013scatter}
Zhigang Ren, Aimin Zhang, Changyun Wen, and Zuren Feng.
\newblock A scatter learning particle swarm optimization algorithm for multimodal problems.
\newblock {\em IEEE Transactions on Cybernetics}, 44(7):1127--1140, 2013.

\bibitem{pant2020differential}
Millie Pant, Hira Zaheer, Laura Garcia-Hernandez, Ajith Abraham, et~al.
\newblock Differential evolution: A review of more than two decades of research.
\newblock {\em Engineering Applications of Artificial Intelligence}, 90:103479, 2020.

\bibitem{biswas2014improved}
Subhodip Biswas, Souvik Kundu, and Swagatam Das.
\newblock An improved parent-centric mutation with normalized neighborhoods for inducing niching behavior in differential evolution.
\newblock {\em IEEE Transactions on Cybernetics}, 44(10):1726--1737, 2014.

\bibitem{biswas2014inducing}
Subhodip Biswas, Souvik Kundu, and Swagatam Das.
\newblock Inducing niching behavior in differential evolution through local information sharing.
\newblock {\em IEEE Transactions on Evolutionary Computation}, 19(2):246--263, 2014.

\bibitem{zhoa2020local}
Hong Zhao, Zhi-Hui Zhan, Ying Lin, Xiaofeng Chen, Xiao-Nan Luo, Jie Zhang, Sam Kwong, and Jun Zhang.
\newblock Local binary pattern-based adaptive differential evolution for multimodal optimization problems.
\newblock {\em IEEE Transactions on Cybernetics}, 50(7):3343--3357, 2020.

\bibitem{bandaru2013parameterless}
Sunith Bandaru and Kalyanmoy Deb.
\newblock A parameterless-niching-assisted bi-objective approach to multimodal optimization.
\newblock In {\em 2013 IEEE Congress on Evolutionary Computation}, pages 95--102. IEEE, 2013.

\bibitem{basak2012multimodal}
Aniruddha Basak, Swagatam Das, and Kay~Chen Tan.
\newblock Multimodal optimization using a biobjective differential evolution algorithm enhanced with mean distance-based selection.
\newblock {\em IEEE Transactions on Evolutionary Computation}, 17(5):666--685, 2012.

\bibitem{lin2019differential}
Xin Lin, Wenjian Luo, and Peilan Xu.
\newblock Differential evolution for multimodal optimization with species by nearest-better clustering.
\newblock {\em IEEE Transactions on Cybernetics}, 51(2):970--983, 2019.

\bibitem{wang2022adaptive}
Zi-Jia Wang, Yu-Ren Zhou, and Jun Zhang.
\newblock Adaptive estimation distribution distributed differential evolution for multimodal optimization problems.
\newblock {\em IEEE Transactions on Cybernetics}, 52(7):6059--6070, 2022.

\bibitem{wang2022multimodal}
Kai Wang, Wenyin Gong, Libao Deng, and Ling Wang.
\newblock Multimodal optimization via dynamically hybrid niching differential evolution.
\newblock {\em Knowledge-Based Systems}, 238:107972, 2022.

\bibitem{liang2014differential}
Jing~J Liang, Bo-Yang Qu, XB~Mao, Ben Niu, and DY~Wang.
\newblock Differential evolution based on fitness euclidean-distance ratio for multimodal optimization.
\newblock {\em Neurocomputing}, 137:252--260, 2014.

\bibitem{deb2012multimodal}
Kalyanmoy Deb and Amit Saha.
\newblock Multimodal optimization using a bi-objective evolutionary algorithm.
\newblock {\em Evolutionary Computation}, 20(1):27--62, 2012.

\bibitem{cheng2017evolutionary}
Ran Cheng, Miqing Li, Ke~Li, and Xin Yao.
\newblock Evolutionary multiobjective optimization-based multimodal optimization: Fitness landscape approximation and peak detection.
\newblock {\em IEEE Transactions on Evolutionary Computation}, 22(5):692--706, 2017.

\bibitem{das2011real}
Swagatam Das, Sayan Maity, Bo-Yang Qu, and Ponnuthurai~Nagaratnam Suganthan.
\newblock Real-parameter evolutionary multimodal optimization—a survey of the state-of-the-art.
\newblock {\em Swarm and Evolutionary Computation}, 1(2):71--88, 2011.

\bibitem{barrera2009review}
Julio Barrera and Carlos A~Coello Coello.
\newblock A review of particle swarm optimization methods used for multimodal optimization.
\newblock {\em Innovations in Swarm Intelligence}, pages 9--37, 2009.

\bibitem{tanabe2019review}
Ryoji Tanabe and Hisao Ishibuchi.
\newblock A review of evolutionary multimodal multiobjective optimization.
\newblock {\em IEEE Transactions on Evolutionary Computation}, 24(1):193--200, 2019.

\bibitem{storn1997differential}
Rainer Storn and Kenneth Price.
\newblock Differential evolution--a simple and efficient heuristic for global optimization over continuous spaces.
\newblock {\em Journal of Global Optimization}, 11:341--359, 1997.

\bibitem{awad2018improved}
Noor~H Awad, Mostafa~Z Ali, Rammohan Mallipeddi, and Ponnuthurai~N Suganthan.
\newblock An improved differential evolution algorithm using efficient adapted surrogate model for numerical optimization.
\newblock {\em Information Sciences}, 451:326--347, 2018.

\bibitem{tanabe2013success}
Ryoji Tanabe and Alex Fukunaga.
\newblock Success-history based parameter adaptation for differential evolution.
\newblock In {\em 2013 IEEE Congress on Evolutionary Computation}, pages 71--78. IEEE, 2013.

\bibitem{zhang2009jade}
Jingqiao Zhang and Arthur~C. Sanderson.
\newblock Jade: Adaptive differential evolution with optional external archive.
\newblock {\em IEEE Transactions on Evolutionary Computation}, 13(5):945--958, 2009.

\bibitem{tanabe2014improving}
Ryoji Tanabe and Alex~S Fukunaga.
\newblock Improving the search performance of shade using linear population size reduction.
\newblock In {\em 2014 IEEE Congress on Evolutionary Computation (CEC)}, pages 1658--1665. IEEE, 2014.

\bibitem{awad2016ensemble}
Noor~H Awad, Mostafa~Z Ali, Ponnuthurai~N Suganthan, and Robert~G Reynolds.
\newblock An ensemble sinusoidal parameter adaptation incorporated with l-shade for solving cec2014 benchmark problems.
\newblock In {\em 2016 IEEE Congress on Evolutionary Computation (CEC)}, pages 2958--2965. IEEE, 2016.

\bibitem{awad2017ensemble}
Noor~H Awad, Mostafa~Z Ali, and Ponnuthurai~N Suganthan.
\newblock Ensemble sinusoidal differential covariance matrix adaptation with euclidean neighborhood for solving cec2017 benchmark problems.
\newblock In {\em 2017 IEEE Congress on Evolutionary Computation (CEC)}, pages 372--379. IEEE, 2017.

\bibitem{li2002species}
Jian-Ping Li, Marton~E Balazs, Geoffrey~T Parks, and P~John Clarkson.
\newblock A species conserving genetic algorithm for multimodal function optimization.
\newblock {\em Evolutionary computation}, 10(3):207--234, 2002.

\bibitem{li2004adaptively}
Xiaodong Li.
\newblock Adaptively choosing neighbourhood bests using species in a particle swarm optimizer for multimodal function optimization.
\newblock In {\em Genetic and Evolutionary Computation--GECCO 2004: Genetic and Evolutionary Computation Conference, Seattle, WA, USA, June 26-30, 2004. Proceedings, Part I}, pages 105--116. Springer, 2004.

\bibitem{petrowski1996clearing}
Alain P{\'e}trowski.
\newblock A clearing procedure as a niching method for genetic algorithms.
\newblock In {\em Proceedings of IEEE International Conference on Evolutionary Computation}, pages 798--803. IEEE, 1996.

\bibitem{li2005efficient}
Xiaodong Li.
\newblock Efficient differential evolution using speciation for multimodal function optimization.
\newblock In {\em Proceedings of the 7th annual conference on Genetic and evolutionary computation}, pages 873--880, 2005.

\bibitem{gong2017learning}
Yue-Jiao Gong, Jun Zhang, and Yicong Zhou.
\newblock Learning multimodal parameters: A bare-bones niching differential evolution approach.
\newblock {\em IEEE Transactions on Neural Networks and Learning Systems}, 29(7):2944--2959, 2017.

\bibitem{wang2013gaussian}
Hui Wang, Shahryar Rahnamayan, Hui Sun, and Mahamed~GH Omran.
\newblock Gaussian bare-bones differential evolution.
\newblock {\em IEEE Transactions on Cybernetics}, 43(2):634--647, 2013.

\bibitem{liu2021double}
Qingxue Liu, Shengzhi Du, Barend~Jacobus van Wyk, and Yanxia Sun.
\newblock Double-layer-clustering differential evolution multimodal optimization by speciation and self-adaptive strategies.
\newblock {\em Information Sciences}, 545:465--486, 2021.

\bibitem{hui2016ensemble}
Sheldon Hui and Ponnuthurai~N. Suganthan.
\newblock Ensemble and arithmetic recombination-based speciation differential evolution for multimodal optimization.
\newblock {\em IEEE Transactions on Cybernetics}, 46(1):64--74, 2016.

\bibitem{preuss2010niching}
Mike Preuss.
\newblock Niching the cma-es via nearest-better clustering.
\newblock In {\em Proceedings of the 12th annual conference companion on Genetic and evolutionary computation}, pages 1711--1718, 2010.

\bibitem{ji2023multimodal}
Junzhong Ji, Tongxuan Wu, and Cuicui Yang.
\newblock Multimodal multiobjective differential evolutionary optimization with species conservation.
\newblock {\em IEEE Transactions on Systems, Man, and Cybernetics: Systems}, 2023.

\bibitem{chen2023network}
Xi-Yuan Chen, Hong Zhao, and Jing Liu.
\newblock A network community-based differential evolution for multimodal optimization problems.
\newblock {\em Information Sciences}, 645:119359, 2023.

\bibitem{huang2017niching}
Ting Huang, Zhi-Hui Zhan, Xing-dong Jia, Hua-qiang Yuan, Jing-qing Jiang, and Jun Zhang.
\newblock Niching community based differential evolution for multimodal optimization problems.
\newblock In {\em 2017 IEEE Symposium Series on Computational Intelligence (SSCI)}, pages 1--8, 2017.

\bibitem{yuan2022self}
Shihao Yuan, Hong Zhao, Jing Liu, and Binjie Song.
\newblock Self-organizing map based differential evolution with dynamic selection strategy for multimodal optimization problems.
\newblock {\em Mathematical Biosciences and Engineering}, 19(6):5968--5997, 2022.

\bibitem{huang2024wireless}
Yi-Biao Huang, Zi-Jia Wang, Yu-Hui Zhang, Yuan-Gen Wang, Sam Kwong, and Jun Zhang.
\newblock Wireless sensor networks-based adaptive differential evolution for multimodal optimization problems.
\newblock {\em Applied Soft Computing}, 158:111541, 2024.

\bibitem{liang2024niche}
Shao-Min Liang, Zi-Jia Wang, Yi-Biao Huang, Zhi-Hui Zhan, Sam Kwong, and Jun Zhang.
\newblock Niche center identification differential evolution for multimodal optimization problems.
\newblock {\em Information Sciences}, page 121009, 2024.

\bibitem{goldberg1987genetic}
David~E Goldberg, Jon Richardson, et~al.
\newblock Genetic algorithms with sharing for multimodal function optimization.
\newblock In {\em Genetic algorithms and their applications: Proceedings of the Second International Conference on Genetic Algorithms}, volume 4149, pages 414--425. Cambridge, MA, 1987.

\bibitem{della2007niches}
Antonio Della~Cioppa, Claudio De~Stefano, and Angelo Marcelli.
\newblock Where are the niches? dynamic fitness sharing.
\newblock {\em IEEE Transactions on Evolutionary Computation}, 11(4):453--465, 2007.

\bibitem{miller1996genetic}
Brad~L Miller and Michael~J Shaw.
\newblock Genetic algorithms with dynamic niche sharing for multimodal function optimization.
\newblock In {\em Proceedings of IEEE International Conference on Evolutionary Computation}, pages 786--791. IEEE, 1996.

\bibitem{li2007multimodal}
Xiaodong Li.
\newblock A multimodal particle swarm optimizer based on fitness euclidean-distance ratio.
\newblock In {\em Proceedings of the 9th annual conference on Genetic and evolutionary computation}, pages 78--85, 2007.

\bibitem{ojala2002multiresolution}
Timo Ojala, Matti Pietikainen, and Topi Maenpaa.
\newblock Multiresolution gray-scale and rotation invariant texture classification with local binary patterns.
\newblock {\em IEEE Transactions on pattern analysis and machine intelligence}, 24(7):971--987, 2002.

\bibitem{wei2021penalty}
Zhifang Wei, Weifeng Gao, Genghui Li, and Qingfu Zhang.
\newblock A penalty-based differential evolution for multimodal optimization.
\newblock {\em IEEE Transactions on Cybernetics}, 52(7):6024--6033, 2021.

\bibitem{pourjafari2012solving}
Ebrahim Pourjafari and Hamed Mojallali.
\newblock Solving nonlinear equations systems with a new approach based on invasive weed optimization algorithm and clustering.
\newblock {\em Swarm and Evolutionary Computation}, 4:33--43, 2012.

\bibitem{hirsch2009solving}
Michael~J Hirsch, Panos~M Pardalos, and Mauricio~GC Resende.
\newblock Solving systems of nonlinear equations with continuous grasp.
\newblock {\em Nonlinear Analysis: Real World Applications}, 10(4):2000--2006, 2009.

\bibitem{mahfoud1992crowding}
SW~Mahfoud.
\newblock Crowding and preselection revisited. parallel problem solving from nature ii, ma{\`e}nner, r. and manderick, b, 1992.

\bibitem{mahfoud1993simple}
Samir~W Mahfoud.
\newblock Simple analytical models of genetic algorithms for multimodal function optimization.
\newblock In {\em ICGA}, volume 643. Citeseer, 1993.

\bibitem{mahfoud1995niching}
Samir~W Mahfoud.
\newblock {\em Niching methods for genetic algorithms}.
\newblock University of Illinois at Urbana-Champaign, 1995.

\bibitem{thomsen2004multimodal}
Rene Thomsen.
\newblock Multimodal optimization using crowding-based differential evolution.
\newblock In {\em Proceedings of the 2004 Congress on Evolutionary Computation (IEEE Cat. No. 04TH8753)}, volume~2, pages 1382--1389. IEEE, 2004.

\bibitem{awad2018ensemble}
Noor~H Awad, Mostafa~Z Ali, and Ponnuthurai~N Suganthan.
\newblock Ensemble of parameters in a sinusoidal differential evolution with niching-based population reduction.
\newblock {\em Swarm and evolutionary computation}, 39:141--156, 2018.

\bibitem{huang2018hypercube}
Haihuang Huang, Liwei Jiang, Xue Yu, and Dongqing Xie.
\newblock Hypercube-based crowding differential evolution with neighborhood mutation for multimodal optimization.
\newblock {\em International Journal of Swarm Intelligence Research (IJSIR)}, 9(2):15--27, 2018.

\bibitem{zhao2023outlier}
Hong Zhao, Zhi-Hui Zhan, and Jing Liu.
\newblock Outlier aware differential evolution for multimodal optimization problems.
\newblock {\em Applied Soft Computing}, 140:110264, 2023.

\bibitem{ursem1999multinational}
Rasmus~K Ursem.
\newblock Multinational evolutionary algorithms.
\newblock In {\em Proceedings of the 1999 congress on evolutionary computation-CEC99 (Cat. No. 99TH8406)}, volume~3, pages 1633--1640. IEEE, 1999.

\bibitem{yao2006clustering}
Jie Yao, Nawwaf Kharma, and Yu~Qing Zhu.
\newblock On clustering in evolutionary computation.
\newblock In {\em 2006 IEEE International Conference on Evolutionary Computation}, pages 1752--1759. IEEE, 2006.

\bibitem{li2014history}
Lingxi Li and Ke~Tang.
\newblock History-based topological speciation for multimodal optimization.
\newblock {\em IEEE Transactions on Evolutionary Computation}, 19(1):136--150, 2014.

\bibitem{damanahi2016novel}
Parisa~Molavi Damanahi, Gelareh Veisi, and Seyyed~Javad Seyyed Mahdavi~Chabok.
\newblock A novel method for multi-modal optimization problems based on differential evolution algorithm.
\newblock In {\em 2015 International Congress on Technology, Communication and Knowledge (ICTCK)}, pages 352--358, 2015.

\bibitem{damanahi2016improved}
Parisa~Molavi Damanahi, Gelareh Veisi, and Seyyed~Javad Seyyed Mahdavi~Chabok.
\newblock Improved differential evolution algorithm based on chaotic theory and a novel hill-valley method for large-scale multimodal optimization problems.
\newblock In {\em 2015 International Congress on Technology, Communication and Knowledge (ICTCK)}, pages 268--275, 2015.

\bibitem{li2023history}
Yu~Li, Lingling Huang, Weifeng Gao, Zhifang Wei, Tianqi Huang, Jingwei Xu, and Maoguo Gong.
\newblock History information-based hill-valley technique for multimodal optimization problems.
\newblock {\em Information Sciences}, 631:15--30, 2023.

\bibitem{liao2023history}
Zuowen Liao, Xianyan Mi, Qishuo Pang, and Yu~Sun.
\newblock History archive assisted niching differential evolution with variable neighborhood for multimodal optimization.
\newblock {\em Swarm and Evolutionary Computation}, 76:101206, 2023.

\bibitem{li2021adaptive}
Jie Li, Yu~Li, Xiaoli Gao, and Weifeng Gao.
\newblock Differential evolution with adaptive subpopulation size for multimodal optimization problems.
\newblock In {\em 2021 China Automation Congress (CAC)}, pages 4904--4908, 2021.

\bibitem{sheng2020differential}
Weiguo Sheng, Xi~Wang, Zidong Wang, Qi~Li, Yujun Zheng, and Shengyong Chen.
\newblock A differential evolution algorithm with adaptive niching and k-means operation for data clustering.
\newblock {\em IEEE Transactions on Cybernetics}, 52(7):6181--6195, 2020.

\bibitem{duan2018adaptive}
Danting Duan, Yuejiao Gong, Ting Huang, and Jun Zhang.
\newblock Adaptive clustering-based differential evolution for multimodal optimization.
\newblock In {\em 2018 Eighth International Conference on Information Science and Technology (ICIST)}, pages 370--376, 2018.

\bibitem{huang2020concurrent}
Ting Huang, Dan-Ting Duan, Yue-Jiao Gong, Long Ye, Wing~WY Ng, and Jun Zhang.
\newblock Concurrent optimization of multiple base learners in neural network ensembles: An adaptive niching differential evolution approach.
\newblock {\em Neurocomputing}, 396:24--38, 2020.

\bibitem{epitropakis2013dynamic}
Michael~G. Epitropakis, Xiaodong Li, and Edmund~K. Burke.
\newblock A dynamic archive niching differential evolution algorithm for multimodal optimization.
\newblock In {\em 2013 IEEE Congress on Evolutionary Computation}, pages 79--86, 2013.

\bibitem{wang2020automatic}
Zi-Jia Wang, Zhi-Hui Zhan, Ying Lin, Wei-Jie Yu, Hua Wang, Sam Kwong, and Jun Zhang.
\newblock Automatic niching differential evolution with contour prediction approach for multimodal optimization problems.
\newblock {\em IEEE Transactions on Evolutionary Computation}, 24(1):114--128, 2020.

\bibitem{yang2018cluster}
Yuan-Hua Yang, Xian-Bin Xu, Shui-Bing He, Jin-Bo Wang, and Yu-Hua Wen.
\newblock Cluster-based niching differential evolution algorithm for optimizing the stable structures of metallic clusters.
\newblock {\em Computational Materials Science}, 149:416--423, 2018.

\bibitem{bovskovic2017clustering}
Borko Bo{\v{s}}kovi{\'c} and Janez Brest.
\newblock Clustering and differential evolution for multimodal optimization.
\newblock In {\em 2017 IEEE Congress on Evolutionary Computation (CEC)}, pages 698--705. IEEE, 2017.

\bibitem{ji2023surrogate}
Jing-Yu Ji, Zusheng Tan, Sanyou Zeng, Eric~WK See-To, and Man-Leung Wong.
\newblock A surrogate-assisted evolutionary algorithm for seeking multiple solutions of expensive multimodal optimization problems.
\newblock {\em IEEE Transactions on Emerging Topics in Computational Intelligence}, 2023.

\bibitem{wang2017dual}
Zi-Jia Wang, Zhi-Hui Zhan, Ying Lin, Wei-Jie Yu, Hua-Qiang Yuan, Tian-Long Gu, Sam Kwong, and Jun Zhang.
\newblock Dual-strategy differential evolution with affinity propagation clustering for multimodal optimization problems.
\newblock {\em IEEE Transactions on Evolutionary Computation}, 22(6):894--908, 2017.

\bibitem{wang2019distributed}
Zi-Jia Wang, Zhi-Hui Zhan, and Jun Zhang.
\newblock Distributed minimum spanning tree differential evolution for multimodal optimization problems.
\newblock {\em Soft Computing}, 23:13339--13349, 2019.

\bibitem{bu2024multi}
Xianglong Bu, Qingke Zhang, Hao Gao, and Huaxiang Zhang.
\newblock Multi-strategy differential evolution algorithm based on adaptive hash clustering and its application in wireless sensor networks.
\newblock {\em Expert Systems with Applications}, 246:123214, 2024.

\bibitem{gionis1999similarity}
Aristides Gionis, Piotr Indyk, Rajeev Motwani, et~al.
\newblock Similarity search in high dimensions via hashing.
\newblock In {\em Vldb}, volume~99, pages 518--529, 1999.

\bibitem{zhang2016toward}
Yu-Hui Zhang, Yue-Jiao Gong, Hua-Xiang Zhang, Tian-Long Gu, and Jun Zhang.
\newblock Toward fast niching evolutionary algorithms: A locality sensitive hashing-based approach.
\newblock {\em IEEE Transactions on Evolutionary Computation}, 21(3):347--362, 2016.

\bibitem{gray2011entropy}
Robert~M Gray.
\newblock {\em Entropy and information theory}.
\newblock Springer Science \& Business Media, 2011.

\bibitem{li2023minimum}
Xiangqian Li, Hong Zhao, and Jing Liu.
\newblock Minimum spanning tree niching-based differential evolution with knowledge-driven update strategy for multimodal optimization problems.
\newblock {\em Applied Soft Computing}, 145:110589, 2023.

\bibitem{agrawal2022solving}
Suchitra Agrawal and Aruna Tiwari.
\newblock Solving multimodal optimization problems using adaptive differential evolution with archive.
\newblock {\em Information Sciences}, 612:1024--1044, 2022.

\bibitem{chen2019distributed}
Zong-Gan Chen, Zhi-Hui Zhan, Hua Wang, and Jun Zhang.
\newblock Distributed individuals for multiple peaks: A novel differential evolution for multimodal optimization problems.
\newblock {\em IEEE Transactions on Evolutionary Computation}, 24(4):708--719, 2019.

\bibitem{wang2023improved}
Yong Wang, Zhen Liu, and Gai-Ge Wang.
\newblock Improved differential evolution using two-stage mutation strategy for multimodal multi-objective optimization.
\newblock {\em Swarm and Evolutionary Computation}, 78:101232, 2023.

\bibitem{zhang2023proximity}
Junna Zhang, Degang Chen, Qiang Yang, Yiqiao Wang, Dong Liu, Sang-Woon Jeon, and Jun Zhang.
\newblock Proximity ranking-based multimodal differential evolution.
\newblock {\em Swarm and Evolutionary Computation}, 78:101277, 2023.

\bibitem{liao2024differential}
Zuowen Liao, Qishuo Pang, and Qiong Gu.
\newblock Differential evolution based on strategy adaptation and deep reinforcement learning for multimodal optimization problems.
\newblock {\em Swarm and Evolutionary Computation}, 87:101568, 2024.

\bibitem{jiang2022self}
Ruizheng Jiang, Jundong Zhang, Yuanyuan Tang, Jinhong Feng, and Chuan Wang.
\newblock Self-adaptive de algorithm without niching parameters for multi-modal optimization problems.
\newblock {\em Applied Intelligence}, 52(11):12888--12923, 2022.

\bibitem{wang2023fitness}
Zi-Jia Wang, Zhi-Hui Zhan, Yun Li, Sam Kwong, Sang-Woon Jeon, and Jun Zhang.
\newblock Fitness and distance based local search with adaptive differential evolution for multimodal optimization problems.
\newblock {\em IEEE Transactions on Emerging Topics in Computational Intelligence}, 7(3):684--699, 2023.

\bibitem{tang2014differential}
Lixin Tang, Yun Dong, and Jiyin Liu.
\newblock Differential evolution with an individual-dependent mechanism.
\newblock {\em IEEE Transactions on Evolutionary Computation}, 19(4):560--574, 2014.

\bibitem{dominico2020self}
Gabriel Dominico, Mateus Boiani, and Rafael~Stubs Parpinelli.
\newblock A self-adaptive differential evolution with local search applied to multimodal optimization.
\newblock In {\em Intelligent Systems Design and Applications: 18th International Conference on Intelligent Systems Design and Applications (ISDA 2018) held in Vellore, India, December 6-8, 2018, Volume 1}, pages 1143--1153. Springer, 2020.

\bibitem{hooke1961direct}
Robert Hooke and Terry~A Jeeves.
\newblock ``direct search''solution of numerical and statistical problems.
\newblock {\em Journal of the ACM (JACM)}, 8(2):212--229, 1961.

\bibitem{moscato1989evolution}
Pablo Moscato et~al.
\newblock On evolution, search, optimization, genetic algorithms and martial arts: Towards memetic algorithms.
\newblock {\em Caltech concurrent computation program, C3P Report}, 826(1989):37, 1989.

\bibitem{ong2010memetic}
Yew-Soon Ong, Meng~Hiot Lim, and Xianshun Chen.
\newblock Memetic computation—past, present \& future [research frontier].
\newblock {\em IEEE Computational Intelligence Magazine}, 5(2):24--31, 2010.

\bibitem{ong2007special}
Yew-Soon Ong, Natalio Krasnogor, and Hisao Ishibuchi.
\newblock Special issue on memetic algorithms.
\newblock {\em IEEE Transactions on Systems, Man, and Cybernetics, Part B (Cybernetics)}, 37(1):2--5, 2007.

\bibitem{chen2011multi}
Xianshun Chen, Yew-Soon Ong, Meng-Hiot Lim, and Kay~Chen Tan.
\newblock A multi-facet survey on memetic computation.
\newblock {\em IEEE Transactions on Evolutionary Computation}, 15(5):591--607, 2011.

\bibitem{sheng2022differential}
Mengmeng Sheng, Shengyong Chen, Weibo Liu, Jiafa Mao, and Xiaohui Liu.
\newblock A differential evolution with adaptive neighborhood mutation and local search for multi-modal optimization.
\newblock {\em Neurocomputing}, 489:309--322, 2022.

\bibitem{neri2022study}
Ferrante Neri and Matthew Todd.
\newblock A study on six memetic strategies for multimodal optimisation by differential evolution.
\newblock In {\em 2022 IEEE Congress on Evolutionary Computation (CEC)}, pages 1--8, 2022.

\bibitem{sheng2021adaptive}
Weiguo Sheng, Xi~Wang, Zidong Wang, Qi~Li, and Yun Chen.
\newblock Adaptive memetic differential evolution with niching competition and supporting archive strategies for multimodal optimization.
\newblock {\em Information Sciences}, 573:316--331, 2021.

\bibitem{wang2022memetic}
Zuling Wang, Ze~Chen, Zidong Wang, Jing Wei, Xin Chen, Qi~Li, Yujun Zheng, and Weiguo Sheng.
\newblock Adaptive memetic differential evolution with multi-niche sampling and neighborhood crossover strategies for global optimization.
\newblock {\em Information Sciences}, 583:121--136, 2022.

\bibitem{fahad2023optimizing}
Shah Fahad, Shoaib~Ahmed Khan, Shiyou Yang, Shafi~Ullah Khan, Mustafa Tahir, and Muhammad Salman.
\newblock Optimizing multi-modal electromagnetic design problems using quantum particle swarm optimization with differential evolution.
\newblock {\em IEEE Access}, 2023.

\bibitem{li2019differential}
Zhihui Li, Li~Shi, Caitong Yue, Zhigang Shang, and Boyang Qu.
\newblock Differential evolution based on reinforcement learning with fitness ranking for solving multimodal multiobjective problems.
\newblock {\em Swarm and Evolutionary Computation}, 49:234--244, 2019.

\bibitem{agrawal2021improved}
Suchitra Agrawal, Aruna Tiwari, Prathamesh Naik, and Arjun Srivastava.
\newblock Improved differential evolution based on multi-armed bandit for multimodal optimization problems.
\newblock {\em Applied Intelligence}, pages 1--22, 2021.

\bibitem{gao2021solving}
Weifeng Gao, Zhifang Wei, Maoguo Gong, and Gary~G Yen.
\newblock Solving expensive multimodal optimization problem by a decomposition differential evolution algorithm.
\newblock {\em IEEE Transactions on Cybernetics}, 53(4):2236--2246, 2021.

\bibitem{cheng1995mean}
Yizong Cheng.
\newblock Mean shift, mode seeking, and clustering.
\newblock {\em IEEE Transactions on Pattern Analysis and Machine Intelligence}, 17(8):790--799, 1995.

\bibitem{wang2019multilevel}
Xi~Wang, Mengmeng Sheng, Kangfei Ye, Jian Lin, Jiafa Mao, Shengyong Chen, and Weiguo Sheng.
\newblock A multilevel sampling strategy based memetic differential evolution for multimodal optimization.
\newblock {\em Neurocomputing}, 334:79--88, 2019.

\bibitem{hong2020multi}
Zhao Hong, Zong-Gan Chen, Dong Liu, Zhi-Hui Zhan, and Jun Zhang.
\newblock A multi-angle hierarchical differential evolution approach for multimodal optimization problems.
\newblock {\em IEEE Access}, 8:178322--178335, 2020.

\bibitem{zhang2022two}
Kai Zhang, Zhiwei Xu, Gary~G Yen, and Ling Zhang.
\newblock Two-stage multiobjective evolution strategy for constrained multiobjective optimization.
\newblock {\em IEEE Transactions on Evolutionary Computation}, 28(1):17--31, 2022.

\bibitem{qiao2023self}
Kangjia Qiao, Jing Liang, Kunjie Yu, Minghui Wang, Boyang Qu, Caitong Yue, and Yinan Guo.
\newblock A self-adaptive evolutionary multi-task based constrained multi-objective evolutionary algorithm.
\newblock {\em IEEE Transactions on Emerging Topics in Computational Intelligence}, 7(4):1098--1112, 2023.

\bibitem{liang2022survey}
Jing Liang, Xuanxuan Ban, Kunjie Yu, Boyang Qu, Kangjia Qiao, Caitong Yue, Ke~Chen, and Kay~Chen Tan.
\newblock A survey on evolutionary constrained multiobjective optimization.
\newblock {\em IEEE Transactions on Evolutionary Computation}, 27(2):201--221, 2022.

\bibitem{liang2022multiobjective}
Jing Liang, Hongyu Lin, Caitong Yue, Kunjie Yu, Ying Guo, and Kangjia Qiao.
\newblock Multiobjective differential evolution with speciation for constrained multimodal multiobjective optimization.
\newblock {\em IEEE Transactions on Evolutionary Computation}, 27(4):1115--1129, 2022.

\bibitem{li2024dynamic}
Guoqing Li, Weiwei Zhang, Caitong Yue, Yirui Wang, Jun Tang, and Shangce Gao.
\newblock A dynamic-speciation-based differential evolution with ring topology for constrained multimodal multi-objective optimization.
\newblock {\em Information Sciences}, page 120879, 2024.

\bibitem{gu2022constrained}
Qinghua Gu, Jiaming Bai, Xuexian Li, Naixue Xiong, and Caiwu Lu.
\newblock A constrained multi-objective evolutionary algorithm based on decomposition with improved constrained dominance principle.
\newblock {\em Swarm and Evolutionary Computation}, 75:101162, 2022.

\bibitem{gu2024multimodal}
Qinghua Gu, Yifan Peng, Qian Wang, and Song Jiang.
\newblock Multimodal multi-objective optimization based on local optimal neighborhood crowding distance differential evolution algorithm.
\newblock {\em Neural Computing and Applications}, 36(1):461--481, 2024.

\bibitem{liang2019multimodal}
Jing Liang, Weiwei Xu, Caitong Yue, Kunjie Yu, Hui Song, Oscar~D Crisalle, and Boyang Qu.
\newblock Multimodal multiobjective optimization with differential evolution.
\newblock {\em Swarm and evolutionary computation}, 44:1028--1059, 2019.

\bibitem{yu2018tri}
Wei-Jie Yu, Jing-Yu Ji, Yue-Jiao Gong, Qiang Yang, and Jun Zhang.
\newblock A tri-objective differential evolution approach for multimodal optimization.
\newblock {\em Information Sciences}, 423:1--23, 2018.

\bibitem{wang2024distributed}
Wei Wang, Zhifang Wei, Tianqi Huang, Xiaoli Gao, and Weifeng Gao.
\newblock A distributed individuals based multimodal multi-objective optimization differential evolution algorithm.
\newblock {\em Memetic Computing}, pages 1--13, 2024.

\bibitem{liang2024multiobjective}
Jing Liang, Hongyu Lin, Caitong Yue, Ponnuthurai~Nagaratnam Suganthan, and Yaonan Wang.
\newblock Multiobjective differential evolution for higher-dimensional multimodal multiobjective optimization.
\newblock {\em IEEE/CAA Journal of Automatica Sinica}, 11(6):1458--1475, 2024.

\bibitem{qu2024improved}
Dan Qu, Hualin Xiao, Huafei Chen, and Hongyi Li.
\newblock An improved differential evolution algorithm for multi-modal multi-objective optimization.
\newblock {\em PeerJ Computer Science}, 10:e1839, 2024.

\bibitem{liang2021clustering}
Jing Liang, Kangjia Qiao, Caitong Yue, Kunjie Yu, Boyang Qu, Ruohao Xu, Zhimeng Li, and Yi~Hu.
\newblock A clustering-based differential evolution algorithm for solving multimodal multi-objective optimization problems.
\newblock {\em Swarm and Evolutionary Computation}, 60:100788, 2021.

\bibitem{yue2021differential}
Caitong Yue, Ponnuthurai~N Suganthan, Jing Liang, Boyang Qu, Kunjie Yu, Yongsheng Zhu, and Li~Yan.
\newblock Differential evolution using improved crowding distance for multimodal multiobjective optimization.
\newblock {\em Swarm and Evolutionary Computation}, 62:100849, 2021.

\bibitem{jia2024novel}
Yingjuan Jia, Liangdong Qu, and Xiaoqin Li.
\newblock A novel multimodal multi-objective differential evolution algorithm based on nearest neighbor-repulsion strategy.
\newblock {\em Information Sciences}, page 120832, 2024.

\bibitem{peng2024multimodal}
Hu~Peng, Wenwen Xia, Zhongtian Luo, Changshou Deng, Hui Wang, and Zhijian Wu.
\newblock A multimodal multi-objective differential evolution with series-parallel combination and dynamic neighbor strategy.
\newblock {\em Information Sciences}, page 120999, 2024.

\bibitem{zhang2020decomposition}
Weiwei Zhang, Ningjun Zhang, Hanwen Wan, Daoying Huang, Xiaoyu Wen, and Yinghui Meng.
\newblock Decomposition based differentiate evolution algorithm with niching strategy for multimodal multi-objective optimization.
\newblock In {\em Bio-inspired Computing: Theories and Applications: 14th International Conference, BIC-TA 2019, Zhengzhou, China, November 22--25, 2019, Revised Selected Papers, Part I 14}, pages 714--726. Springer, 2020.

\bibitem{varela2022niching}
Daniel Varela and Jos{\'e} Santos.
\newblock Niching methods integrated with a differential evolution memetic algorithm for protein structure prediction.
\newblock {\em Swarm and Evolutionary Computation}, 71:101062, 2022.

\bibitem{xie2022multimodal}
Hui Xie, Shengli Sun, Tianru Xue, Wenjun Xu, Huikai Liu, Linjian Lei, and Yue Zhang.
\newblock A multimodal differential evolution algorithm in initial orbit determination for a space-based too short arc.
\newblock {\em Remote Sensing}, 14(20):5140, 2022.

\bibitem{yang2024manifold}
Xiongyan Yang, Xianfeng Yuan, Lin Dong, Xiaoxue Mei, and Ke~Chen.
\newblock Manifold assistant multi-modal multi-objective differential evolution algorithm and its application in actual rolling bearing fault diagnosis.
\newblock {\em Engineering Applications of Artificial Intelligence}, 133:108040, 2024.

\bibitem{ishibuchi2011many}
Hisao Ishibuchi, Naoya Akedo, and Yusuke Nojima.
\newblock A many-objective test problem for visually examining diversity maintenance behavior in a decision space.
\newblock In {\em Proceedings of the 13th annual conference on Genetic and evolutionary computation}, pages 649--656, 2011.

\bibitem{minisci2009orbit}
Edmondo~A Minisci and Giulio Avanzini.
\newblock Orbit transfer manoeuvres as a test benchmark for comparison metrics of evolutionary algorithms.
\newblock In {\em 2009 IEEE Congress on Evolutionary Computation}, pages 350--357. IEEE, 2009.

\bibitem{zhang2024differential}
Guozhang Zhang, Shengwei Fu, Ke~Li, and Haisong Huang.
\newblock Differential evolution with multi-strategies for uav trajectory planning and point cloud registration.
\newblock {\em Applied Soft Computing}, page 112466, 2024.

\bibitem{li2017brain}
Zhe Li, Yong Xia, Zexuan Ji, and Yanning Zhang.
\newblock Brain voxel classification in magnetic resonance images using niche differential evolution based bayesian inference of variational mixture of gaussians.
\newblock {\em Neurocomputing}, 269:47--57, 2017.

\bibitem{agrawal2023feature}
Suchitra Agrawal, Aruna Tiwari, Bhaskar Yaduvanshi, and Prashant Rajak.
\newblock Feature subset selection using multimodal multiobjective differential evolution.
\newblock {\em Knowledge-Based Systems}, 265:110361, 2023.

\bibitem{agrawal2023differential}
Suchitra Agrawal, Aruna Tiwari, Bhaskar Yaduvanshi, and Prashant Rajak.
\newblock Differential evolution with nearest better clustering for multimodal multiobjective optimization.
\newblock {\em Applied Soft Computing}, 148:110852, 2023.

\bibitem{hu2021multimodal}
Xiao-Min Hu and Zi-Wen Guo.
\newblock Multimodal bare-bone niching differential evolution in feature selection.
\newblock In {\em 2021 IEEE International Conference on Systems, Man, and Cybernetics (SMC)}, pages 1553--1558. IEEE, 2021.

\bibitem{wang2022differential}
Peng Wang, Bing Xue, Jing Liang, and Mengjie Zhang.
\newblock Differential evolution-based feature selection: A niching-based multiobjective approach.
\newblock {\em IEEE Transactions on Evolutionary Computation}, 27(2):296--310, 2022.

\bibitem{wang2021multiobjective}
Peng Wang, Bing Xue, Jing Liang, and Mengjie Zhang.
\newblock Multiobjective differential evolution for feature selection in classification.
\newblock {\em IEEE Transactions on Cybernetics}, 53(7):4579--4593, 2021.

\bibitem{ribeiro2023decoding}
Matheus Henrique Dal~Molin Ribeiro, Ramon~Gomes da~Silva, Jos{\'e} Henrique~Kleinubing Larcher, Andre Mendes, Viviana~Cocco Mariani, and Leandro dos~Santos Coelho.
\newblock Decoding electroencephalography signal response by stacking ensemble learning and adaptive differential evolution.
\newblock {\em Sensors}, 23(16):7049, 2023.

\bibitem{nasim2022pnn}
Amnah Nasim and Yoon~Sang Kim.
\newblock De-pnn: Differential evolution-based feature optimization with probabilistic neural network for imbalanced arrhythmia classification.
\newblock {\em Sensors}, 22(12):4450, 2022.

\bibitem{babu2024hybridization}
Chukka~Ramesh Babu, M~Suneetha, Mohammed~Altaf Ahmed, Palamakula~Ramesh Babu, Mohamad~Khairi Ishak, Hend~Khalid Alkahtani, and Samih~M Mostafa.
\newblock Hybridization of synergistic swarm and differential evolution with graph convolutional network for distributed denial of service detection and mitigation in iot environment.
\newblock {\em Scientific Reports}, 14(1):30868, 2024.

\bibitem{ran2024development}
Longyi Ran, Zheng Wang, Bing Yang, Alireza Amiri-Margavi, and Najim Alshahrani.
\newblock Development of novel computational models based on artificial intelligence technique to predict the viscosity of ionic liquids-water mixtures.
\newblock {\em Case Studies in Thermal Engineering}, 54:104076, 2024.

\bibitem{vincent2023improved}
Amala~Mary Vincent and P~Jidesh.
\newblock An improved hyperparameter optimization framework for automl systems using evolutionary algorithms.
\newblock {\em Scientific Reports}, 13(1):4737, 2023.

\bibitem{xie2022birdsongs}
Shanshan Xie, Yan Zhang, Danjv Lv, Haifeng Xu, Jiang Liu, and Yue Yin.
\newblock Birdsongs recognition based on ensemble elm with multi-strategy differential evolution.
\newblock {\em Scientific Reports}, 12(1):9739, 2022.

\bibitem{sun2024dbpboost}
Ailun Sun, Hongfei Li, Guanghui Dong, Yuming Zhao, and Dandan Zhang.
\newblock Dbpboost: a method of classification of dna-binding proteins based on improved differential evolution algorithm and feature extraction.
\newblock {\em Methods}, 223:56--64, 2024.

\bibitem{dedeturk2024csa}
Beyhan~Adanur Dedeturk, Bilge~Kagan Dedeturk, and Burcu Bakir-Gungor.
\newblock Csa-de-lr: enhancing cardiovascular disease diagnosis with a novel hybrid machine learning approach.
\newblock {\em PeerJ Computer Science}, 10:e2197, 2024.

\bibitem{wang2025customer}
Guanqun Wang.
\newblock Customer segmentation in the digital marketing using a q-learning based differential evolution algorithm integrated with k-means clustering.
\newblock {\em PloS one}, 20(2):e0318519, 2025.

\bibitem{wang2023extreme}
Qiaoyun Wang, Shuai Song, Lei Li, Da~Wen, Peng Shan, Zhigang Li, and YongQing Fu.
\newblock An extreme learning machine optimized by differential evolution and artificial bee colony for predicting the concentration of whole blood with fourier transform raman spectroscopy.
\newblock {\em Spectrochimica Acta Part A: Molecular and Biomolecular Spectroscopy}, 292:122423, 2023.

\bibitem{hu2021targetdbp+}
Jun Hu, Liang Rao, Yi-Heng Zhu, Gui-Jun Zhang, and Dong-Jun Yu.
\newblock Targetdbp+: enhancing the performance of identifying dna-binding proteins via weighted convolutional features.
\newblock {\em Journal of Chemical Information and Modeling}, 61(1):505--515, 2021.

\bibitem{jiang2021optimizing}
Yi~Jiang, Zhi-Hui Zhan, Kay~Chen Tan, and Jun Zhang.
\newblock Optimizing niche center for multimodal optimization problems.
\newblock {\em IEEE Transactions on Cybernetics}, 53(4):2544--2557, 2021.

\bibitem{derrac2011practical}
Joaqu{\'\i}n Derrac, Salvador Garc{\'\i}a, Daniel Molina, and Francisco Herrera.
\newblock A practical tutorial on the use of nonparametric statistical tests as a methodology for comparing evolutionary and swarm intelligence algorithms.
\newblock {\em Swarm and Evolutionary Computation}, 1(1):3--18, 2011.

\bibitem{liu2023learning}
Xin Liu, Jianyong Sun, Qingfu Zhang, Zhenkun Wang, and Zongben Xu.
\newblock Learning to learn evolutionary algorithm: A learnable differential evolution.
\newblock {\em IEEE Transactions on Emerging Topics in Computational Intelligence}, 7(6):1605--1620, 2023.

\bibitem{luo2021survey}
Wenjian Luo, Xin Lin, Jiajia Zhang, and Mike Preuss.
\newblock A survey of nearest-better clustering in swarm and evolutionary computation.
\newblock In {\em 2021 IEEE Congress on Evolutionary Computation (CEC)}, pages 1961--1967. IEEE, 2021.

\bibitem{van2024explainable}
Niki van Stein, Diederick Vermetten, Anna~V Kononova, and Thomas B{\"a}ck.
\newblock Explainable benchmarking for iterative optimization heuristics.
\newblock {\em arXiv preprint arXiv:2401.17842}, 2024.

\end{thebibliography}
\end{document}